\documentclass[12pt, English]{report}
\usepackage{amsmath,amssymb,amsfonts}
\usepackage{graphics}
\usepackage[sfdefault]{roboto} 
\usepackage[T1]{fontenc}
\usepackage{epsfig}
\usepackage{psfrag}
\usepackage{setspace}
\usepackage{amsthm}
\usepackage{soul}
\newtheorem{definition}{Problem}
\newtheorem{theorem}{Theorem} 
\usepackage{cite}
\usepackage{url}
\usepackage{titlesec}
\usepackage{multirow}
\usepackage{algorithm,algpseudocode}
\usepackage[paper=portrait,pagesize]{typearea}
\usepackage{blindtext}
\usepackage{color, colortbl}
\usepackage[colorlinks=true,linkcolor=red,urlcolor=blue,citecolor=blue]{hyperref}
\usepackage[singlelinecheck=off]{caption}
\usepackage[colorinlistoftodos]{todonotes}
\usepackage[hang,flushmargin]{footmisc}
\usepackage{lipsum}
\usepackage{bm}
\usepackage{array}
\newcolumntype{P}[1]{>{\centering\arraybackslash}p{#1}}
\usepackage{booktabs}
\usepackage{float}
\usepackage{array}
\usepackage{lscape}
\usepackage{longtable}
\usepackage{threeparttable}
\usepackage{amssymb}
\usepackage{balance}  
\usepackage{url}      
\usepackage[lofdepth,lotdepth,position=bottom]{subfig}
\usepackage{mathtools}
\usepackage{rotating}
\usepackage{breakurl}
\usepackage{array}
\usepackage{color}
\usepackage{textcomp}
\usepackage{esvect}
\usepackage{amsmath,stackengine}
\usepackage{url}
\usepackage{graphicx}
\usepackage{ccicons}                              

\usepackage{multirow}
\usepackage{amsmath}

\usepackage{comment}
\usepackage{lscape}

\usepackage{amsmath,amssymb,amsfonts}
\usepackage{mathtools}
\usepackage{rotating}
\usepackage{breakurl}
\usepackage{array}
\usepackage{color}
\usepackage{textcomp}
\usepackage{esvect}
\usepackage{amsmath,stackengine}
\usepackage{bbm}
\usepackage{scrextend}
\usepackage{pseudocode}
\usepackage{color}
\usepackage[scientific-notation=true]{siunitx}
\usepackage[textwidth=6in,textheight=9in,left=1.5in,right=1in,top=1in,bottom=1in]{geometry} 
\usepackage[colorlinks=true,
            linkcolor=red,
            urlcolor=blue,
            citecolor=blue]{hyperref} 
\usepackage{rotating}

\titleformat{\paragraph}[hang]{\normalfont\normalsize\bfseries}{\theparagraph}{1em}{}
\titlespacing*{\paragraph}{0pt}{3.25ex plus 1ex minus .2ex}{1em}

\titleformat{\section}
  {\normalfont\large\bfseries}{\thesection}{1em}{} 
\titleformat{\subsection}
  {\normalfont\normalsize\bfseries}{\thesubsection}{1em}{} 
\titleformat{\subsubsection}
  {\normalfont\normalsize\itshape}{\thesubsubsection}{1em}{} 

\usepackage{adjustbox}
\usepackage[document]{ragged2e}
\usepackage{afterpage}
\begin{document}
\singlespacing
\pagenumbering{gobble} 
\newgeometry{top=2in}
\begin{center}
		{\LARGE Data-Driven Intersection Management Solutions for Mixed Traffic of Human-Driven and Connected and Automated Vehicles \par}
    \vfill
    \rule{11.5cm}{1pt}\\
    ~\\
    {A Dissertation \\}
    ~\\
    {Presented to \\}
    ~\\
    {The faculty of the School of Engineering and Applied Science \\}
    ~\\
    {University of Virginia \\}
    ~\\
    \rule{11.5cm}{1pt}\\
    {in partial fulfillment\\}
    ~\\
    {of the requirements for the degree\\}
    ~\\
    ~\\
    {Doctor of Philosophy\\}
    ~\\
    {by\\}
    ~\\
    {Masoud Bashiri}\\
    ~\\
\vfill
    {December\\}
    ~\\
    {2020}
		\thispagestyle{empty}
		
		\newpage
\pagenumbering{arabic}

{\huge \textbf{Approval Sheet}}\\
\vspace{30pt}
This Dissertation is submitted in partial fulfillment of the requirements for the degree of
Doctor of Philosophy (Systems and Information Engineering)

\vspace{10pt}
\begin{flushleft}
\rule{14.5cm}{1pt}\vspace{-3pt}\\
Masoud Bashiri\\
\vspace{10pt}

This Dissertation has been read and approved by the Examining Committee:
\vspace{30pt}
{

\rule{14.5cm}{1pt}\vspace{-3pt}\\
Cody H. Fleming, Ph.D. (School of Engineering and Applied Science)\\~\\~\\
\rule{14.5cm}{1pt}\vspace{-3pt}\\
Madhur Behl, Ph.D. (School of Engineering and Applied Science)\\~\\~\\
\rule{14.5cm}{1pt}\vspace{-3pt}\\
Nicola Bezzo, Ph.D. (School of Engineering and Applied Science)\\~\\~\\
\rule{14.5cm}{1pt}\vspace{-3pt}\\
T. Donna Chen, Ph.D. (School of Engineering and Applied Science)\\~\\~\\
\rule{14.5cm}{1pt}\vspace{-3pt}\\
B Brian Park, Ph.D. (School of Engineering and Applied Science)\\~\\
}

Accepted for the School of Engineering and Applied Science:

\vspace{50pt}
\centering

\rule{12.5cm}{1pt}\vspace{-3pt}\\
Craig H. Benson, Dean, School of Engineering and Applied Science

December 2020

\end{flushleft}

\end{center}

\restoregeometry
\newcommand{\RNum}[1]{%
  \textup{\uppercase\expandafter{\romannumeral#1}}%
}

\null
\vfill
\begin{center}
\copyright Copyright by Masoud Bashiri 2020\\
~\\
All Rights Reserved
\end{center}
\setcounter{page}{3}  
\doublespacing
\chapter*{Abstract}
According to the U.S. Federal Highway Administration, outdated traffic signal timing currently accounts for more than 10 percent of all traffic delays. On average, adaptive signal control technologies improve travel time by more than 10 percent in comparison with traditional signal timing methods. In areas with particularly outdated signal timing, improvements can be 50 percent or more.

With the emergence of connected and automated vehicles and the recent advancements in Intelligent Transportation Systems, Autonomous Traffic Management has garnered more attention. Cooperative Intersection Management (CIM) is among the more challenging traffic problems that pose important questions related to safety and optimization in terms of vehicular delays, fuel consumption, emissions and reliability.

This dissertation proposes two solutions for urban traffic control in the presence of connected and automated vehicles. First a centralized platoon-based controller is proposed for the cooperative intersection management problem that takes advantage of the platooning systems and V2I communication to generate fast and smooth traffic flow at a single intersection. Two cost functions are proposed to minimize total delay and delay variance.

Simulated experiments show that the proposed controller produces schedules that minimize travel delay and variance while increasing intersection throughput and reducing fuel consumption, when compared to traffic light policies. The simulations also verify the positive effect of platooning on fuel consumption and intersection throughput.

Second, a data-driven approach is proposed for adaptive signal control in the presence of connected vehicles. The proposed system relies on a data-driven method for optimal signal timing and a data-driven heuristic method for estimating routing decisions. It requires no additional sensors to be installed at the intersection, reducing the installation costs compared to typical settings of state-of-the-practice adaptive signal controllers. 

The proposed traffic controller contains an optimal signal timing module and a traffic state estimator. The signal timing module is a neural network model trained on microscopic simulation data to achieve optimal results according to a given performance metric such as vehicular delay or average queue length. The traffic state estimator relies on connected vehicles' information to estimate the traffic's routing decisions. A heuristic method is proposed to minimize the estimation error. With sufficient parameter tuning, the estimation error decreases as the market penetration rate (MPR) of connected vehicles grow. Estimation error is below 30\% for an MPR of 10\% and it shrinks below 20\% when MPR grows larger than 30\%.

Simulations showed that the proposed traffic controller outperforms Highway Capacity Manual's methodology and given proper offline parameter tuning, it can decrease average vehicular delay by up to 25\%. 
\chapter*{Acknowledgements}
\doublespacing
I would like to express my sincere gratitude to my advisor Professor Cody Fleming for the continuous support of my Ph.D. study and related research. In four years I have learned a lot from you. Your immense knowledge has been a big help and your motivation inspires me. I feel that under your mentorship, I have become a better scientist, more patient and motivated. Your guidance helped me all the time in research and writing of this dissertation. I could not have imagined having a better advisor and mentor for my Ph.D. study.

I would also like to thank the rest of my thesis committee for their insightful comments, encouragements and especially the hard questions that helped me widen my research.

I thank my fellow labmates in the Coordinated Systems Lab for the stimulating discussions and the fun times we had!

Last but not the least, I would like to thank my family. Mother, I cannot thank you enough for your unconditional love, infinite support and guidance and the sacrifices you made that got me where I am today and shaped my character as a person. To my brothers and sisters: I love you all and I hope to see you soon!

And to the love of my life, my beautiful wife Nazila: From the day you entered my life, suddenly everything is better. Challenges are easier to face, hardships are easier to endure and life is more beautiful. I dedicate this to you. Love You!
{ \hypersetup{linkcolor=blue}
\setcounter{tocdepth}{7}
\tableofcontents
\cleardoublepage
\addcontentsline{toc}{chapter}{List of Tables}
\listoftables
\cleardoublepage
\addcontentsline{toc}{chapter}{List of Figures}
\listoffigures
\cleardoublepage
}

\doublespacing

\chapter{Introduction}\label{chpt:intro}

Traffic Signal Control (TSC) is an important, effective and most widely used method for traffic control in urban settings. TSC systems have gone through various improvements over the past century and yet, intersections are still known to be a major contributor to traffic accidents. According to National Highway Traffic Safety Administration (NHTSA),~\cite{choi2010crash}~40 percent of all crashes that occurred in the United States in 2008 were intersection-related. Moreover, traffic efficiency is reported to be closely correlated with traffic safety on intersections~\cite{chang2003relationship}.

Traffic congestion is also among the more important contributors of~$CO_{2}$ emissions~\cite{grote2016including} which is the largest constituent of transport's greenhouse gas emissions. Vehicle stop times at intersections also contribute to carbon monoxide ($CO$) emissions. The international energy agency reports that when vehicles are idle at an intersection they emit about~5--7 times as much $CO$ as vehicles traveling between~5--10 mph~\cite{statistics2011co2}.

TSC methods can be divided into three main categories of fixed-time (pre-timed) control, traffic-responsive control, and intelligent control. Fixed-time control method relies on predetermined cycle times and splits and is known to produce stable and regular traffic flows. Webster \cite{webster1958traffic} and Miller \cite{miller1963settings} introduced a traffic signal model and analytical formulas for calculation of optimal settings with regard to average vehicular delay.

Macroscopic traffic simulation and signal timing optimization programs have been introduced that build on the pre-timed signal control and utilize search algorithms to optimize cycle length, phasing sequence, splits, and offsets for single intersections or a network. Examples of such programs include, TRANSYT \cite{transyt}, Synchro \cite{husch2003synchro} and PTV Vistro \cite{america2014ptv}. The issue with such methods, however, is that traffic systems are dynamic, therefore, fixed-time signal control cannot adequately respond to real traffic conditions.

With the advancement of sensing technologies such as Piezoelectric sensors and inductive loop systems and more recently, camera systems, real-time traffic-responsive control has become a popular choice for intersection control \cite{tsc_survey}.

Actuated control (AC) is the most widely used traffic-responsive method in practice. Actuated control regulates traffic signal timings according to sensing of traffic flows by detectors such as piezoelectric, magnetic sensors and camera systems that are installed in the network. An AC system performs phase selection and extension based on traffic demands. It is known that an AC system is suitable for traffic scenarios that involve relatively high randomness and low to medium average traffic levels, e.g. under 80\% traffic saturation. However, AC systems are not suitable for higher traffic levels and cannot achieve optimal usage of time and space since they disregard the queues of vehicles on the other phases when performing phase extension and selection.

Sydney coordinated adaptive traffic system (SCAT)~\cite{sims1980sydney} is a plan-selection control system in which a time plan for each intersection is chosen by the overall need of subsystems. Split cycle offset optimization technique (SCOOT)\cite{hunt1982scoot} is an adaptive real-time plan-generating control system, which makes real-time adjustments of splits, cycles, and time offset parameters with a small-step incremental optimization approach.

Traditional signal control strategies based on mathematical traffic-flow models provide many useful ideas and new methods for traffic control applications, but their calculations are often computationally complex. This complexity makes it challenging to meet real-time requirements, and the assumptions of mathematical models of traffic flow are too strict to account for generality of TSC algorithms~\cite{tsc_survey}.

As a response to these issues, intelligent control methods have been introduced. These methods have used technologies in computational intelligence, such as Artificial Neural Networks (ANNs)~\cite{niittymaki2000signal}, Fuzzy Systems~\cite{spall1997traffic} and Reinforcement Learning (RL)~\cite{abdulhai2003reinforcement}.

This dissertation introduces two novel approaches to the problem of cooperative intersection management. The first method proposes leveraging the autonomy and communication capabilities of connected and automated vehicles for more efficient traffic control at urban intersections; the second contribution of this dissertation is a novel technique for signal timing optimization that leverages the precision and ubiquity of traffic data provided by microscopic traffic simulators to build computational models that serve as signal timing optimizer.

These models can be customized for a single intersection or a network for various objectives such as average vehicular delay, queue size, etc. Moreover, the complexity of the model is not a function of traffic level or the size of the intersection, hence, eliminating the computational overhead problem of some of the previous work. The proposed method is complemented with a heuristic approach to dynamic routing decision estimation that relies on information received from Connected Vehicles (CVs).

The rest of this dissertation is organized as follows. Chapter~\ref{chpt:lit_review} provides an overview of the problem's background, motivations and challenges. Chapter~\ref{chpt:cim} introduces a platoon-based approach to Cooperative Intersection Management~(CIM). The proposed data-driven approach for optimal signal timing is presented in Chapter~\ref{chpt:signal}. Chapter~\ref{chpt:tse} introduces a novel traffic state estimation technique in the presence of CAVs. In Chapter~\ref{chpt:method}, the proposed traffic state estimator and the proposed optimal signal timing model are integrated into a novel data-driven adaptive signal controller. And finally, Chapter~\ref{chpt:conclusions} provides a summary of contributions, future work and conclusions to this dissertation.

\chapter{Problem Background, Motivations and Challenges}\label{chpt:lit_review}
\section{Traffic Signal Timing}\label{Sec:lit_signal}
The introduction of traffic lights has helped improve the traffic condition at intersections. Moreover, in recent decades, a number of adaptive traffic light systems have further improved the performance of traffic signals. These systems rely on traffic estimation techniques to adapt the traffic signal settings according to the traffic conditions. Most notable examples of those systems are SCOOT~\cite{hunt1982scoot}, the Sydney Coordinated Adaptive Traffic (SCAT) system~\cite{sims1980sydney}, and RHODES~\cite{mirchandani2001real}.

With the emergence of autonomous ground vehicles and the recent advancements in Intelligent Transportation Systems, Autonomous Traffic Management has garnered more and more attention. Autonomous Intersection Management (AIM), also known as Cooperative Intersection Management (CIM) is among the more challenging traffic problems that pose important questions related to safety and optimization in terms of delays, fuel consumption, emissions and reliability.

\subsection{Motivations \& Challenges}
The motivations and challenges for the CIM problem can be divided into two classes of societal and technical. What follows is a general description of the motivations and challenges of the problem.

\subsubsection{Societal Motivations}
As bottlenecks of traffic flow, intersections are known to be a major contributor to traffic accidents. According to National Highway Traffic Safety Administration (NHTSA),~40\%  of all crashes and~21.5\% of the corresponding fatalities that occurred in the United States in 2008 were intersection-related~\cite{choi2010crash}. Moreover, traffic efficiency is reported to be closely correlated with traffic safety on intersections~\cite{chang2003relationship}.

Traffic congestion is also among the more important contributors of~$CO_{2}$ emissions~\cite{grote2016including} which is the largest constituent of transport's greenhouse gas~(GHG) emissions. Vehicle stop times at intersections also contribute to carbon monoxide ($CO$) emissions. International energy agency reports that when vehicles are idle at an intersection they emit about~5--7 times as much $CO$ as vehicles traveling between~5--10 mph~\cite{statistics2011co2}.

According to several studies, traffic safety and efficiency are closely correlated.~\cite{dias2009relationship} and~\cite{chang2003relationship} report that accident frequency increases with congestion level for intersections. Studies also show that for intersections with low congestion level, possibilities of severe crashes with casualties are higher because of head-on crashes and involvement of vulnerable road users (VRUs). For high congestion intersections, on the other hand, accidents are less serious due to lower speeds but the frequency of accidents are much higher. Therefore, traffic safety and efficiency need to be jointly considered.

Transportation systems today rarely utilize the advantages of autonomous and coordinated driving, however the future of transportation belongs to autonomous driving and if used properly, it can optimize traffic flow, reduce GHG emissions, reduce the number and/or the severity of accidents and as a result reduce the number of fatalities. 
\subsubsection{Non-Technical Challenges}
\begin{itemize}
    \item \textbf{People \& Culture: } Proposing a new solution to a problem with societal aspects will require convincing and training the people that will interact with the system. In the context of intersection control, people have developed habits over a century and any attempt to make drastic changes to the system has to be studied and analyzed from a psychological and social sciences perspective. Steps have to be carefully planned to address society's skepticism, habits and attitude. More specifically, researchers must not forget the important role of human beings as a part of the entire system. This research provides a solution to the CIM problem that requires minimal changes in terms of the interactions of users with the system. This choice will ease the transition from the traditional traffic signals to a new system.
    \item \textbf{Time \& Money: } There are also challenges regarding the implementation costs of a new solution in terms of money and time. "Can we afford to buy and install the required equipment to implement the new system?", "How long would it take to implement?", "What would be the costs of transitioning from the old system to the new system?". These are just some examples of very important questions that have to be answered in order to decide about the feasibility of implementation of a new system in the real-world.
\end{itemize}
\subsubsection{Technical Challenges \& Motivations}
Utilizing Connected and Automated Vehicles (CAVs) abilities in the context of transportation systems has also introduced an array of technical problems and challenges to the industry. This proposal will  address challenges that are specifically important for the Cooperative Intersection Management (CIM) problem.
\begin{itemize}
    \item \textbf{Legacy Systems: } Intersections today are controlled by traditional techniques such as traffic signals and stop signs. New solutions to the CIM problem need to be carefully designed to require minimal changes to the systems that are already installed at intersections. This proposal will address this problem with a strategy that is completely compatible with legacy systems.
    \item \textbf{Coordination Architecture: } From a control theory perspective, the literature on CIM can be divided into two categories of \textit{centralized} and \textit{distributed} methodologies ~\cite{chen2016cooperative}. Centralized CIM involves a coordination unit that communicates with the vehicles, receives information about vehicles' routing decisions and gives instructions to vehicles on how and when to pass through the intersection. Distributed CIM relies on the communications between vehicles through Vehicular Ad-Hoc Network~(VANET) to negotiate and agree on vehicle priorities to pass through the intersection.
    
    The centralized methodology has attracted more research in the past, due to its simplicity, lower risk factors and compatibility with traffic lights methodology, compared to the distributed methodology. A major challenge in designing scheduler algorithms for centralized CIM is scalability with respect to size of the intersection and level of traffic. Meta-heuristics and Mathematical methods such as Linear Programming (LP), Integer Linear Programming, dynamic programming have been widely applied to centralized CIM. These methods usually define space tile and time slot allocations for individual vehicles as the decision variables in their formulation, ignoring optimal vehicle maneuvers through braking, throttle, etc.
    
    Distributed methodologies offload computation onto the vehicles' computer systems. While distributed CIM eliminates the computational problems of the centralized CIM, it relies entirely on the performance of the communication channel and the negotiation protocol, which has  brought up serious questions and concerns about its performance and safety guarantee. With the absence of a central coordination unit, distributed CIM is also not desirable for mixed-traffic scenarios, i.e. the presence of human-driven vehicles, pedestrians and cyclists among fully autonomous vehicles. Due to the above issues, this proposal does not consider distributed control architectures as a candidate for mixed human-driven and autonomous traffic management. 
    
    \item \textbf{Human-Driven Vehicles: } One of the more challenging issues for future CIM strategies is mixed-traffic scenarios. This issue has been left unanswered for the most part in the literature as the focus has been on leveraging the autonomy related features of the CAVs to optimize the traffic throughput. Dresner et. al.~\cite{dresner2008multiagent} proposed a two-phase strategy that involves a traffic light phase to handle human-driven vehicle traffic at an intersection. Simulation results showed that the proposed mechanism provides little or no improvement over today’s traffic signals when less than 90\% of the vehicles are autonomous.
    
    More recently, in~\cite{sharon2017intersection} Sharon proposed another mechanism that complements Dresner's original method to handle mixed traffic scenarios. While simulation results showed significant improvements over the original approach, there are still questions regarding guarantee of safety and the requirement for expensive equipment that need to be addressed.
    \item \textbf{Communications: } VANETs are a subclass of mobile Ad-Hoc networks (MANETs), that is a promising approach for the future of intelligent transportation system (ITS). These networks have no fixed infrastructure and instead rely on the vehicles themselves to provide network functionality. Since the CIM problem is a safety application, connectivity of the network must be maintained and certain required performance guarantees have to be defined with respect to network performance metrics such as coverage and latency. The issues regarding communication are out of scope of this research and will not be further discussed.
\end{itemize}
\section{Proposed Methods}
Chapter \ref{chpt:cim} proposes a platoon-based approach to CIM that addresses the existing scalability issues in the CIM literature. This is achieved through the utilization of vehicle platoons which results in improvements in terms of average vehicular delay, intersection throughput, fuel consumption and computational and communication overhead. 

However, the proposed method still has issues with respect to human acceptance, strict assumptions about the traffic and applicability. Therefore, chapters \ref{chpt:signal}, \ref{chpt:tse} and \ref{chpt:method} build towards an architecture and a set of algorithms that leverage the advancements in connected vehicles technology that relies on the existing infrastructure with minimal requirements for new infrastructure. This results in a scalable signalized approach that is cost effective and accounts for human driving. The proposed method offers significant improvement in system performance over the current state of the practice.

\chapter{Platoon-Based Cooperative Intersection Management}\label{chpt:cim}

\section{Introduction and Related Work}\label{sec:lit_cim}
Traffic Lights and stop signs are the major methods for traffic control  used at intersections. While traffic lights have helped improve traffic flow at intersections, they are still considered inefficient and a contributor to traffic congestion and accidents. Statistically a majority of intersection-related accidents occur in the presence of traffic lights~\cite{choi2010crash}.

In the past two decades, adaptive and smart traffic light controllers have been introduced and deployed, with significant improvements in terms of delay and congestion. It has been shown that the performance of traditional traffic lights can be improved through machine learning based approaches such as Fuzzy logic~\cite{niittymaki2000signal}, neural networks~\cite{spall1997traffic} and Reinforcement Learning~\cite{abdulhai2003reinforcement}, and mathematical models such as Mixed-Integer Linear Programming (MILP)~\cite{lin2004enhanced}. However, traffic lights have remained a major contributor to congestion and traffic accidents. Moreover, while signalized intersections work well with human drivers, they don't necessarily leverage the advantages associated with autonomous vehicles.

Recent advancements of Information Technology and the emergence of Vehicular Adhoc Networks~(VANETs) that support Vehicle to Vehicle (V2V), Vehicle to Infrastructure~(V2I) and Vehicle to Pedestrian (V2P) communications have brought forth opportunities for further advancements of intersection management infrastructure. This includes new non-signalized approaches for the intersection management problem, commonly known as Cooperative Intersection Management~(CIM), where road users, i.e. vehicles communicate with the infrastructure and/or other users to cooperatively coordinate the traffic flow.

In 2014, the Institute of Electrical and Electronics Engineers~(IEEE) published the Wireless Access for Vehicle Environments~(WAVE)~\cite{ieee2013ieee} specification. The standards define architectures based on Dedicated Short Range Communications~(DSRC) for which the Society of Automotive Engineers~(SAE) has specified message types and data elements through various standards. Due to the limited capacity of current V2X communications, the communication complexity is one of the critical issues for CIM~\cite{chen2016cooperative}. 

The introduction of V2V communications has led to a new interest in vehicle platooning, i.e. a group of vehicles that can travel very closely together, safely at a set speed. Vehicle platooning is usually achieved through cooperative adaptive cruise control~(CACC) systems. In CACC, communication between vehicles provides enhanced information so that vehicles can follow their predecessors with higher accuracy, faster response, and shorter gaps. This will result in an enhanced traffic flow.

In \cite{lioris2017platoons} Lioris et al. study the effects of platooning on a network of signalized intersections with fixed time control. The results showed that the network will support an increase in demand by a factor of two or three if all saturation flows are increased by the same factor, while maintaining the same average delay and travel time per vehicle.

This chapter proposes a method to take the findings in \cite{lioris2017platoons} further to take advantage of V2I communications and utilize information received from platoon leader vehicles to achieve optimal intersection control policies.
\section{Cooperative Intersection Management}{\label{sec:CIM}}
Several methods have been proposed to leverage autonomous and connected vehicles for the intersection management. The new methodologies for CIM can be categorized into two classes of Centralized and Distributed methods~\cite{rios2017survey}. 

In centralized methods, a central intersection manager unit receives real-time information and requests from road users and decides how to coordinate the traffic flow, e.g. instruct vehicles how, when and if to pass the intersection. Distributed methods, on the other hand, do not rely on a central control unit. Instead, all vehicles collaboratively plan their trajectories. These methods usually involve negotiation protocols to make decisions on a high level and each user/vehicle makes decisions based on shared objectives given local information from its sensors on a lower level.

\subsection{Centralized Methods}
In~\cite{dresner2004multiagent}, Dresner et al. propose a centralized resource reservation algorithm based on First Come First Serve (FCFS) policy. The control unit receives requests from all vehicles approaching the intersection, simulates the vehicle's movement through the intersection, given the information in the request, and confirms the request if there is no conflict with previously accepted trajectories, otherwise the request would be rejected and the vehicle has to request at a later time. The authors assume constant speed at the intersection and perform simulations comparing results with a traffic light and an overpass. 

It was shown that the proposed method outperformed traffic light in terms of average vehicle delay. The authors further developed their work in~\cite{dresner2005multiagent} to add several methods to improve the performance and overcome major disadvantages of the previous work. Moreover, they improved the work in~\cite{dresner2008multiagent} and~\cite{sharon2017intersection} by adding support for human drivers, considering emergency vehicles that generally have higher priority, and communication schemes for vehicles with different levels of autonomy. 

In~\cite{miculescu2014polling}, a case of two merging roads (one lane) is modeled as a polling system with two queues and one server. The polling system determines the sequence of times assigned to the vehicles on each lane to enter the merging road. The arrival times along with the trajectories of the leading vehicle are then used in a coordination algorithm to generate optimal trajectories for each vehicle in the queue. 

Lee and Park~\cite{lee2012development} derive a nonlinear constrained optimization problem to enhance the performance a traffic signal controller in presence of fully autonomous vehicles. A phase conflict map of the traffic signal is used as part of the optimization problem.

\subsection{Decentralized Methods}
In~\cite{azimi2012intersection} a controller model has been proposed along with different V2V-based intersection management protocols to enhance traffic throughput and safety. Each vehicle runs a collision avoidance algorithm that takes in all safety messages that are being broadcast by surrounding vehicles and detects possible collision. Estimations are then used to generate alternative trajectories to avoid collisions. 

Wu et al.~\cite{wu2015distributed} proposed a decentralized stop-and-go based algorithm that relies on wireless shared information among all approaching vehicles. The vehicle with a shorter estimated arrival time will cross the intersection, while others will need to come to a complete stop until the conflict zone is cleared. Vehicles with non-conflicting turning movements can cross simultaneously.
\subsection{Motivation and Contributions}
The existing approaches are limited with respect to at least one of the following factors: 
\begin{itemize}
    \item unrealistic or infeasible bandwidth requirements for communication
    \item no performance guarantees, i.e. no guarantee that the intersection will behave better than signalized intersections
    \item no formal safety guarantee
    \item scalability with respect to number of cars and lanes
    \item unrealistic assumptions about vehicle behavior
\end{itemize}
Any solutions for the CIM problem has to be compatible with real world communication capacity of vehicular networks, otherwise such solutions will not be feasible. To address this issue I propose a platoon-based approach where vehicles request a pass as a platoon. The intersection manager utilizes an optimization based policy to allocate slots in time and space for any platoon approaching the intersection. Communicating with platoon leaders, instead of every approaching vehicle, decreases the amount of communication needed. This approach also takes advantage of recent advances in platooning and connected vehicle control.

The performance of the solution has to be verified with respect to various performance metrics such as average delay, intersection throughput, average speed, fuel consumption, etc. Microscopic simulations are conducted to verify the performance of the proposed method in comparison to a pre-timed 4-phase traffic light controller. Table~\ref{tab:metrics} shows the performance metrics used in this work.

The solution has to be scalable to growing number of roads/lanes and vehicles. A scalable solution is achieved by only taking the information from the closest platoons to the intersection at each iteration. This guarantees that the computational complexity of the algorithm remains relatively unchanged as the size of the input grows.

Most of the previously published papers make simplistic assumptions about vehicle dynamics, e.g. second order dynamics where the control input generates instant accelerations, disregarding disturbance forces, and oversimplifying the driving force, which is a function of throttle and brake positions among other factors. In this chapter, A model for vehicle dynamics is utilized to generate more realistic results when compared to the relevant literature. These assumptions help design control strategies that are more feasible for use in the real world.
\begin{table}[ht]
\vspace{-6pt}
\centering
\caption{Performance Measure Index}
\label{tab:metrics}
\begin{tabular}{@{}cc@{}}
\toprule
\multicolumn{2}{c}{\textbf{Performance Measure Index}} \\ \midrule
Performance Index                 & Unit               \\ \midrule
Average Delay                     & s                  \\ \midrule
Delay Standard Deviation                    & s                 \\ \midrule
Intersection Throughput           & veh/hour           \\ \midrule
Fuel Consumption                  & ml/veh             \\ \bottomrule
\end{tabular}
\vspace{-8pt}
\end{table}

\section{Platoon-based Intersection Management}
In~\cite{bashiri2017platoon} I proposed that platooning in the vicinity of intersections could reduce communication overhead by allowing platoon leaders to negotiate with the infrastructure and other platoons on behalf of the followers. Moreover, it can help improve the efficiency of \emph{any} scheduling policy by enabling smooth trajectories in the conflict zone. The simulation results showed that the proposed stop-sign based controller outperformed a regular stop sign by~$50\%$ in terms of average delay per vehicle and~$40\%$ in variance of delay. The proposed policies decrease computational complexity by only including one platoon per lane into the schedule.

\subsection{Modifying the Policies to Enhance the Traffic Flow}
\cite{bashiri2017platoon} introduced a reservation-based policy that utilizes cost functions that minimize delay, or a combination of delay and variance, to derive optimal schedules for platoons of vehicles. Such schedules would decrease average delay per vehicle while decreasing the variance in delay due to the fairness of the cost functions and as a result increase intersection throughput and decrease the average fuel consumption in the vicinity of an intersection. The proposed policy guarantees safety by not allowing vehicles with conflicting turning movement to be in the conflict zone at the same time.

In this work, the policies in~\cite{bashiri2017platoon} have been modified, a communication protocol is designed and some of the simplistic assumptions about vehicle dynamics have been removed to achieve the following goals:
\begin{enumerate}
    \item Achieve a solution under realistic assumptions
    \item Improve efficiency in terms of delay and fuel consumption
    \item Design a communication protocol that hides the algorithm from the vehicle agents
\end{enumerate}
\subsubsection{Communication Protocol}
A protocol is designed to ease the Vehicle to Vehicle (V2V) and Vehicle to Infrastructure (V2I) communications. Using this protocol, the intersection manager only has to communicate with the leader of a platoon. Each vehicle is broadcasting its state on a~$10Hz$ frequency, while receiving incoming packets on the same frequency. There are four types of packets designed for the platoon leaders and two types for the infrastructure as follows.
\begin{description}
  \item[Vehicle Message Types:]\ 
    \begin{itemize}
      \item \emph{Request}
      \item \emph{Change-Request}
      \item \emph{Acknowledge}
      \item \emph{Done}
    \end{itemize}
  \item[Infrastructure Message Types:]\ 
    \begin{itemize}
      \item \emph{Acknowledge}
      \item \emph{Confirm}
      \item \emph{Reject}
    \end{itemize}
\end{description}
Approaching platoon leaders send a $Request$ message once they are in a pre-defined proximity of the intersection, and await the response from the manager. Both~\emph{Request} and~\emph{Change-Request} messages consist of the unique Vehicle Identification Number (VIN) of the leader, current position, velocity and acceleration of the leader, estimated arrival time at the conflict zone and the size of the platoon, e.g. number of followers.

Upon receiving a request, the manager sends an~\emph{Acknowledge} message to the sender, runs the scheduler and responds with either an~\emph{Confirm} or a~\emph{Reject} message. The manager expects to receive an~\emph{Acknowledge} from the corresponding vehicle and if such message is not received, it re-sends the message until the acknowledgement is received. The vehicles can send a~\emph{Change-Request} message to the manager if they need to update information in their previous request. Once a vehicle has finished crossing the conflict zone, it is required to send a~\emph{Done} message to the manager, which lets the manager know it can remove the corresponding vehicle's platoon from the schedule.

This protocol is designed to hide the scheduling policy from the vehicle agents, therefore allowing for online changes to the policy without requiring any changes in communications.
\subsubsection{Policy}
The scheduling problem is formulated as the following minimization problem.
\begin{equation}
    \label{eq:argmin}
    arg\min_s \delta(s) = \{s \mid s \in S \wedge \forall s' \in S: \delta(s) \leq \delta(s')\}
\end{equation}
Where $s$ is a schedule of vehicles (platoons), $\delta$ is a cost function designed to penalize total delay and variance in delay, given a schedule, as in Equations~\ref{eq:cost1} and~\ref{eq:cost2}. 
\begin{align}
    \delta_{1}(s) &= \sum_{j=2}^{N} j\left(d(p_{j}) + \sum_{i=1}^{j-1}t_{c}(p_{i})\right)\label{eq:cost1}\\
    \delta_{2}(s) &= \sum_{j=2}^{N} \left(d(p_{j}) + \sum_{i=1}^{j-1}t_{c}(p_{i})\right)\label{eq:cost2}
\end{align}
In the above equations,~$N$ is the number of platoons in the schedule,~$j$ is the platoon's turn,~$d$ is a function that returns the sum of the current delay of the vehicles within a given platoon, and again $s$ is the given schedule. This delay,~$d$, is the difference between the vehicle's original expected arrival time and its expected arrival time assuming it will be the next platoon going through.~$t_{c}$ is a function that computes the additional delay caused by the platoons that are given higher priorities than the~$j$th platoon in the schedule.

The scheduler in Equation~\ref{eq:argmin} essentially simulates all possible schedules given the set of platoons in the queue, and returns a schedule that has the lowest score in terms of the cost function~$\delta_{1}(s)$ or~$\delta_{2}(s)$. For the rest of this chapter,~$\delta_{1}(s)$ and~$\delta_{2}(s)$ will be referred to as~\emph{Platoon-based Variance Minimization}~(\emph{PVM}) and~\emph{Platoon-based Delay Minimization}~(\emph{PDM}), respectively.

The scheduling procedure will be called by the controller every time a change is detected in the set of platoons in the schedule. For example, the procedure is called when a platoon finishes crossing the conflict zone and as a result is removed from the schedule, or when a new platoon enters the communication zone. 

The controller keeps a record of the most recent schedule and sends a message to the platoon that is at the top of the queue. The controller also checks for non-conflicting turning movements in the schedule with that of the platoon at the top of the queue, and lets them cross the intersection simultaneously. The clearance time of the intersection is then updated by the maximum estimated clearance time of the platoons that are crossing the conflict zone. 

Algorithm~\ref{alg:1} shows the intersection management algorithm in pseudo code.
\begin{algorithm}[h]
  \noindent\begin{minipage}{\linewidth}
   \caption{Autonomous Intersection Manager}
    \begin{algorithmic}[1]
    \Function{IntersectionManager}{}
	\While{$True$}
		\State $P = getRequests()$ \footnote{Where P is a map of platoons paired with their respective request information}
		\State $sort(P)$\footnote{Sort the platoon list based on their expected arrival time to make the candidate selection run faster}
		\State $pool = selectCandidates(P)$
		\If{$!pool.isUpdated()$}
			\State $continue$ \footnote{Skip this iteration if the selection pool has not changed}
	    \EndIf
	    \State $[platoons,schedule] = getSchedule(pool)$
		\State $i=1$
        \For{$platoon$ in $platoons$}
        	\State $update(platoon,schedule_{i})$
        	\State $i++$
        \EndFor
        \EndWhile
       \EndFunction
\end{algorithmic}
    \label{alg:1}
\end{minipage}
\end{algorithm}

The algorithm considers at most one platoon for each lane that is in the communication range of the central controller, i.e. the leading platoon in each lane. Such design makes the policy scalable, in that the computational complexity of the scheduler remains suitably low as the number of incoming lanes grow.

\subsubsection{Computational Complexity}
The algorithm considers all permutations of the set of platoons, including possible non-conflicting trajectories in which case simultaneous crossing is considered. These permutations can be modeled as a permutation problem to pick from~$K$ elements without replacement and placing them in sets of~$\{K, K-1, ..., 1\}$ placeholders. Therefore, the worst case computational complexity of such algorithm would be equal to the number of possible permutations, given in Equation~\ref{eq:perms}.
\begin{equation}
    \label{eq:perms}
    T = \sum_{r=1}^{N}\sum_{i=0}^{r-1} (-1)^{i} {r\choose i}(r-i)^N
\end{equation}

In the above equation, $T$ is the total number of possible schedules, $N$ is the number of lanes, $r$ and $i$ are index counters. One may note that the exponential nature of this complexity can only be acceptable for small number of incoming lanes. It can be shown that the algorithm will have the worst case computational complexity of $O(N^N)$. For example, the 4-way intersection considered in this work would only require 75 possible schedules to be considered in the worst case. Due to space constraints, the details and proof of computational complexity are omitted.

A heuristic is proposed for larger intersections to reduce computational complexity. The proposed heuristic does not consider non-conflicting trajectories and therefore reduces the complexity to the number of possible permutations which is exactly $N!$, i.e. computational complexity of $O(N!)$. After a schedule is selected, the controller allows those platoons with non-conflicting trajectories with regard to the selected platoon to cross the intersection simultaneously.

\begin{figure}[H]
    \centering
    \includegraphics[width=.95\columnwidth]{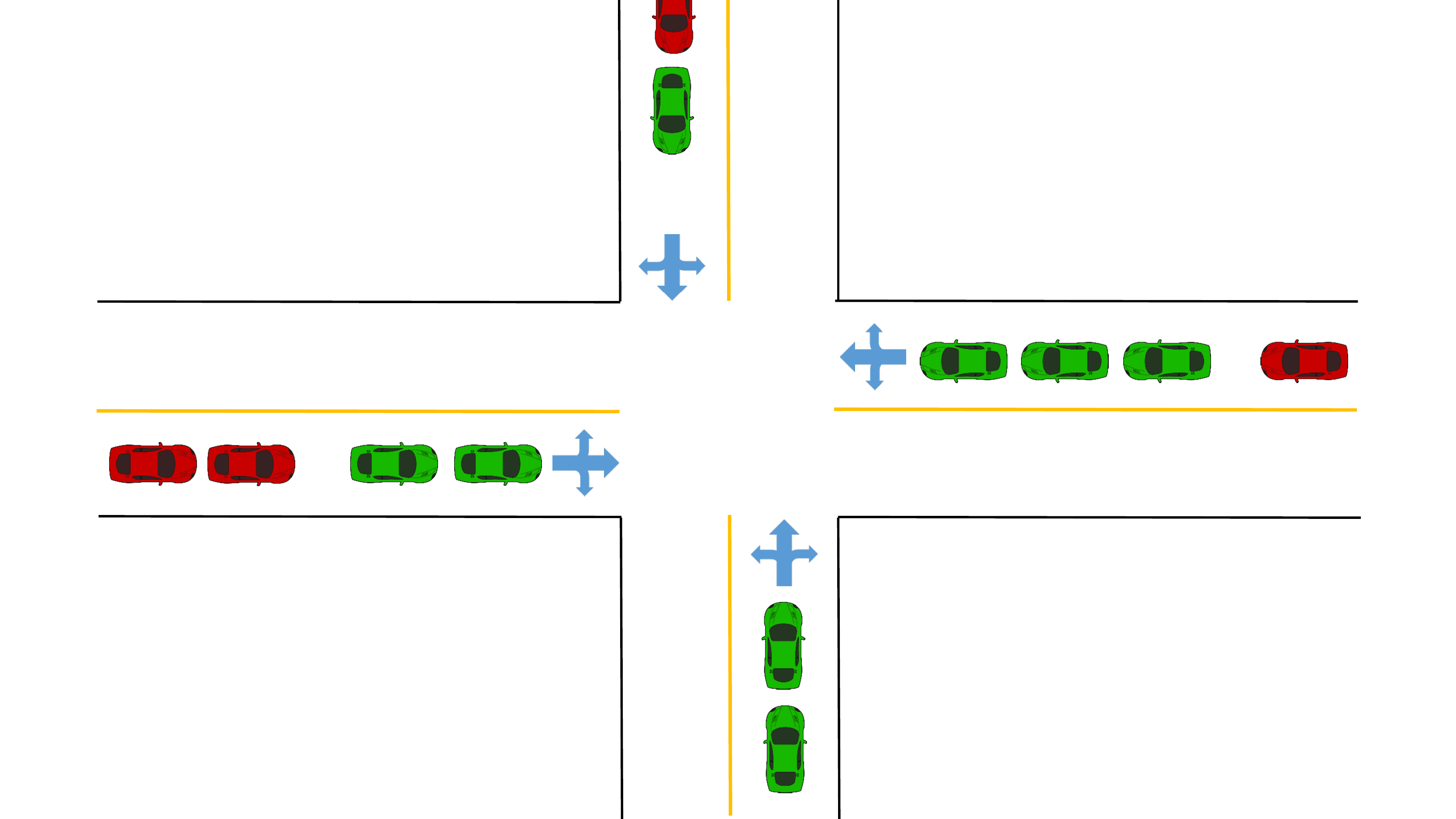}
    \caption{4-way Intersection Geometry}
    \label{fig:intersection}
\end{figure}

Fig.~\ref{fig:intersection} demonstrates how the proposed heuristic ignores the platoons behind the closest platoon to reduce the computational complexity. The platoons/cars in red represent the ignored input to the algorithm. This figure also serves as a visual aid to represent the geometry and turning policy of the intersection that was used for the simulations.

\subsubsection{Simulations}\label{sec:simulations}
\begin{figure}[bh]
    \centering
    \includegraphics[width=.95\columnwidth,trim={0 1.5cm 0 1.5cm},clip]{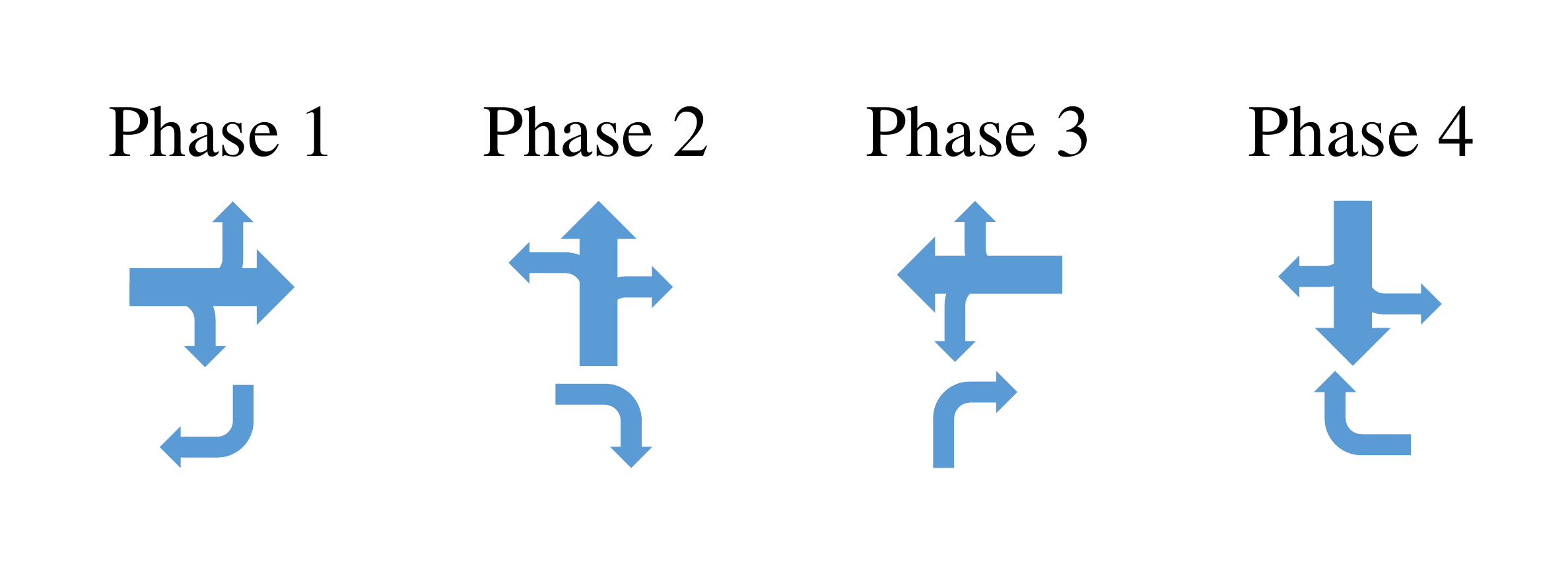}
    \caption{Phase Plan}
    \label{fig:phaseplan}
\end{figure}

\begin{figure}[th]
    \centering
    \includegraphics[width=\columnwidth,trim={0 0.5cm 0 0.5cm},clip]{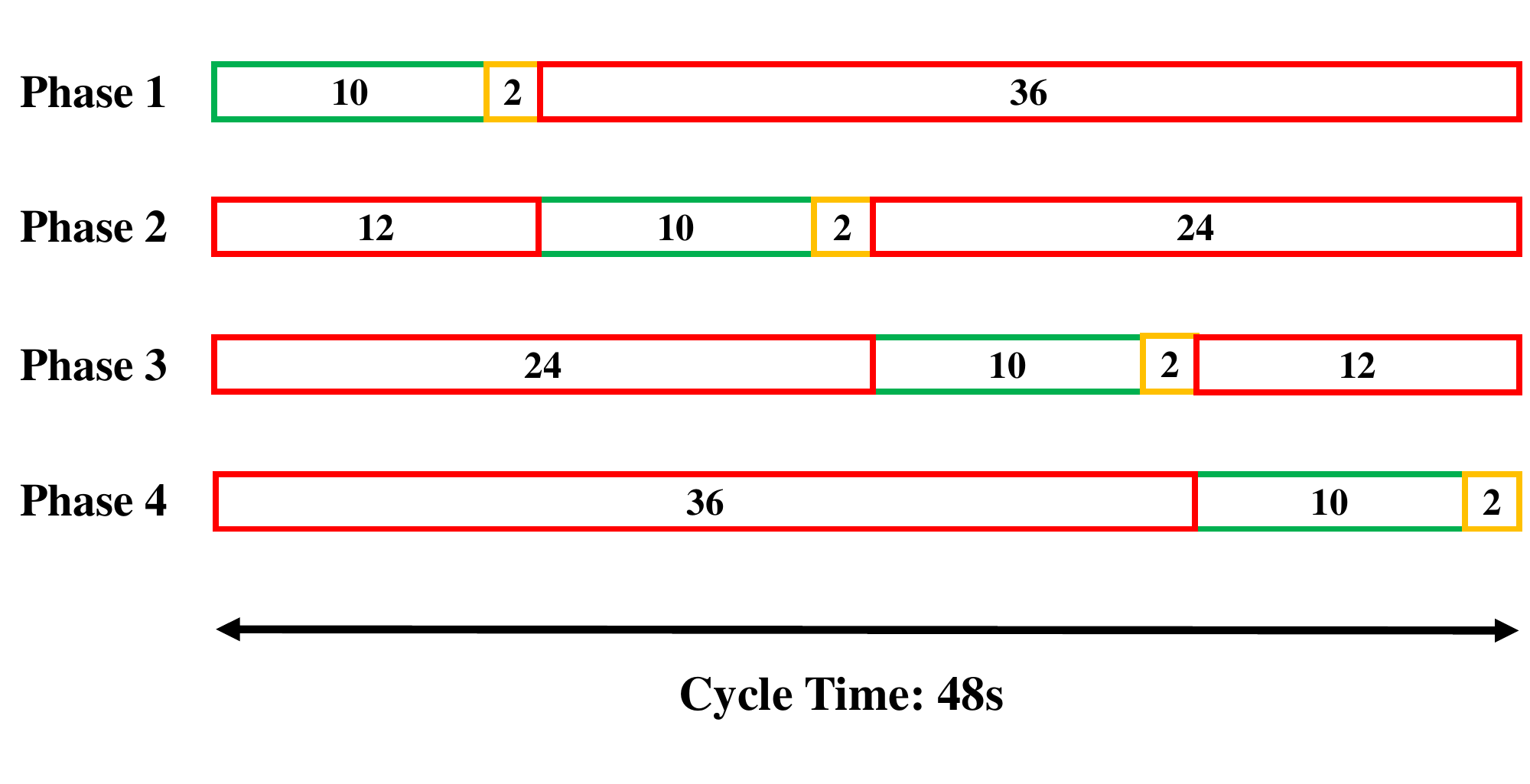}
    \caption{Timing Diagram}
    \label{fig:timing}
\end{figure}

A fixed-time 4-phase traffic light controller is tuned for the 4-way intersection shown in Fig.~\ref{fig:intersection}. The phase plan and timing diagram of the baseline traffic light policy are shown in Figs~ \ref{fig:phaseplan} and~\ref{fig:timing} respectively.

To compare the performance of the two proposed methods against the baseline policy, 20 scenarios are designed by choosing different parameter settings in terms of incoming traffic flow and maximum platoon size. Each policy is evaluated for every variation of the two parameters in table~\ref{tab:parameters}. The incoming traffic flow is equal on all approaches with $70\%$ of the traffic going straight,~$20\%$ turning right and~$10\%$ turning left. To minimize the influence of randomness, each simulation is run for 60 minutes.

\begin{table}[h]
\centering
\caption{Simulation Parameters}
\label{tab:parameters}
\begin{tabular}{@{}ccc@{}}

                  Parameter Name   & Set of Values       & unit    \\ \midrule
Traffic Level        & \{300,400,500,600\} & veh/hour/lane \\ \midrule
Simulation Time        & \{3600\} & s \\ \midrule
Maximum Platoon Size & \{1,2,3,4,5\}       & veh     \\ \bottomrule
\end{tabular}
\end{table}
Recorded videos of simulation for~\emph{PVM}\footnote{\href{https://youtu.be/RtN0f7BlFyg}{https://youtu.be/RtN0f7BlFyg}} and~\emph{PDM}\footnote{\href{https://youtu.be/qHGv9LF72NA}{https://youtu.be/qHGv9LF72NA}} are available.

\begin{figure}[H]
\centering
    \includegraphics[width=0.7\textwidth]{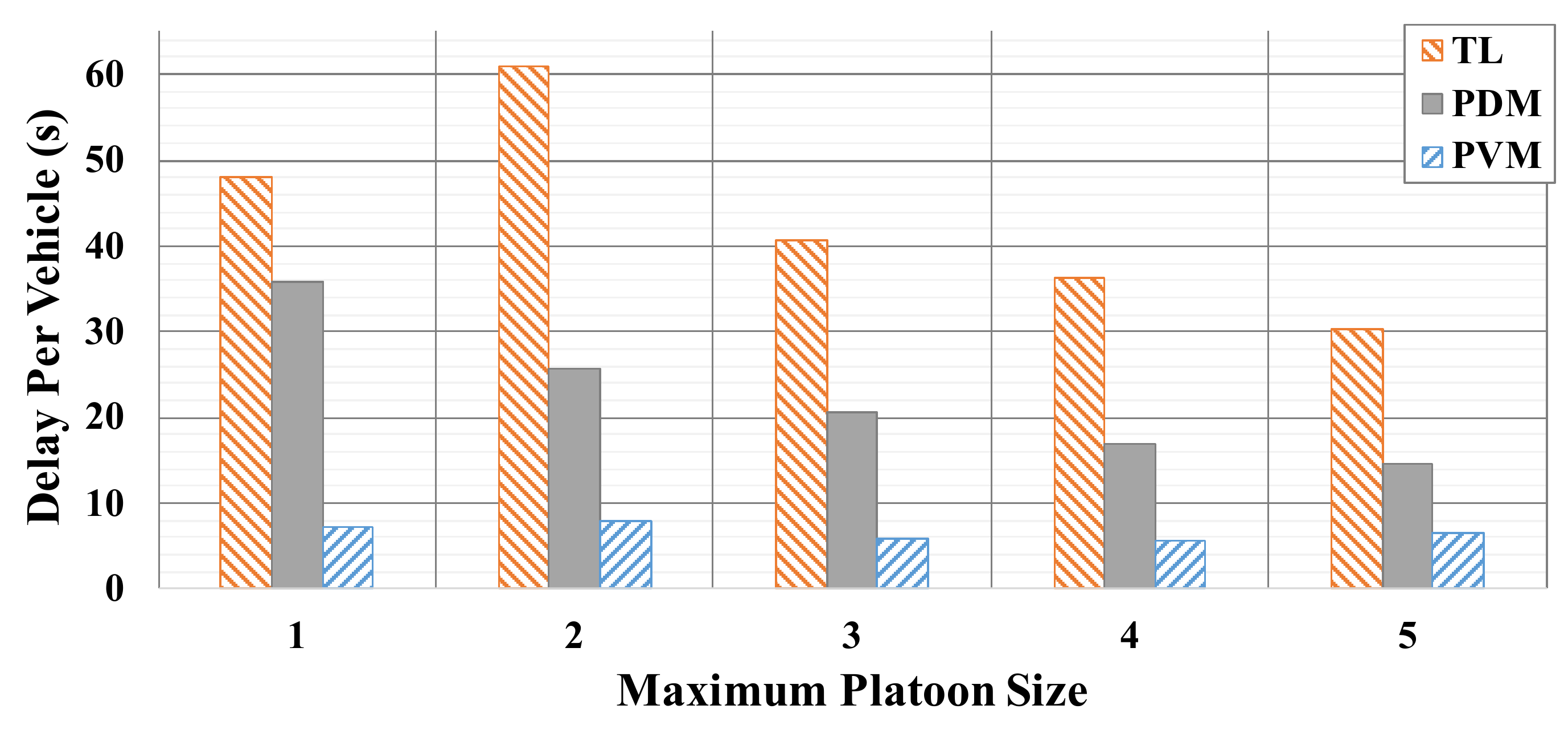}
    \caption{Average Delay Per Vehicle}\label{fig:1}
    
    \vspace*{\floatsep}
    
    \centering
     \includegraphics[width=0.7\textwidth]{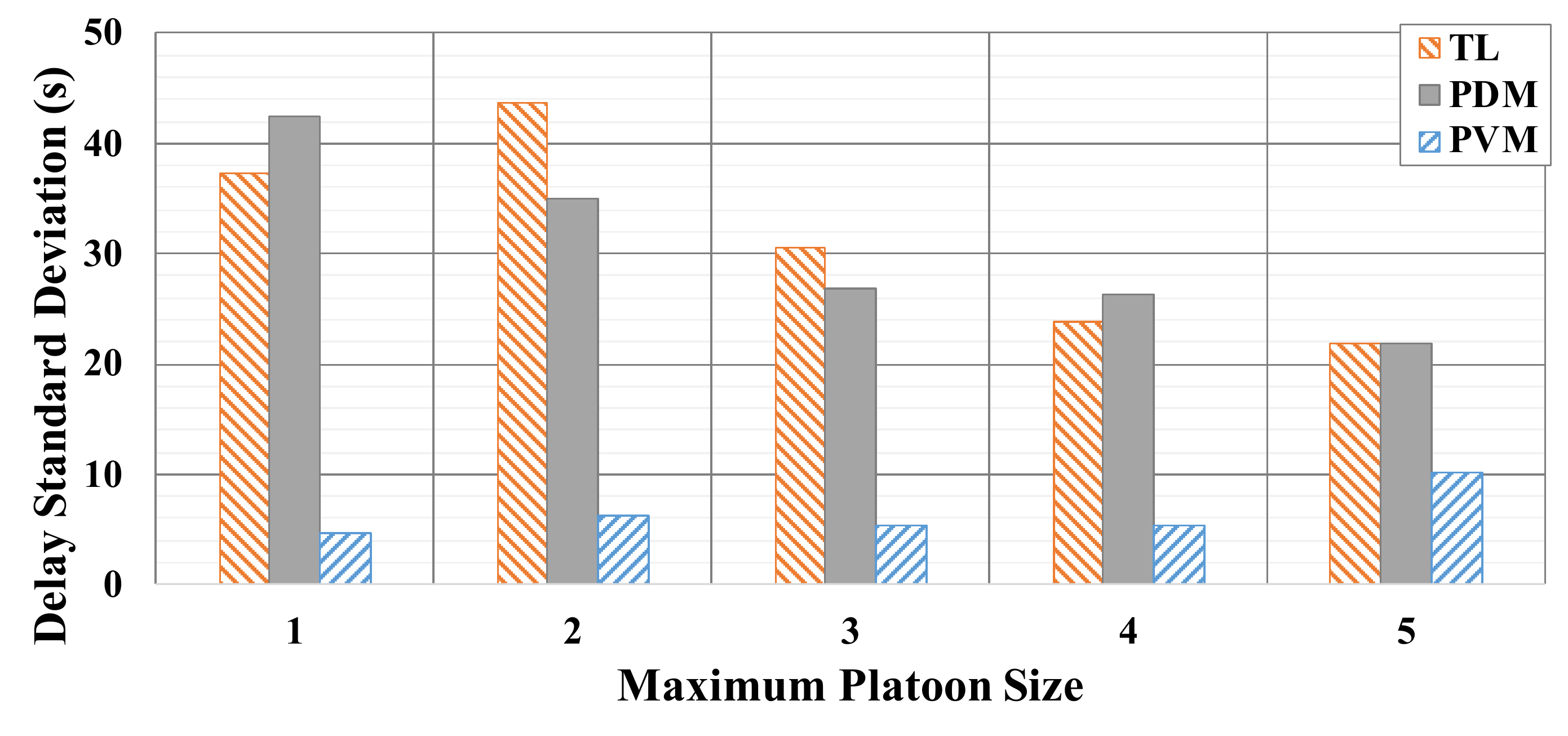}
    \caption{Delay Standard Deviation}\label{fig:2}
    
        \vspace*{\floatsep}

    \includegraphics[width=0.7\textwidth]{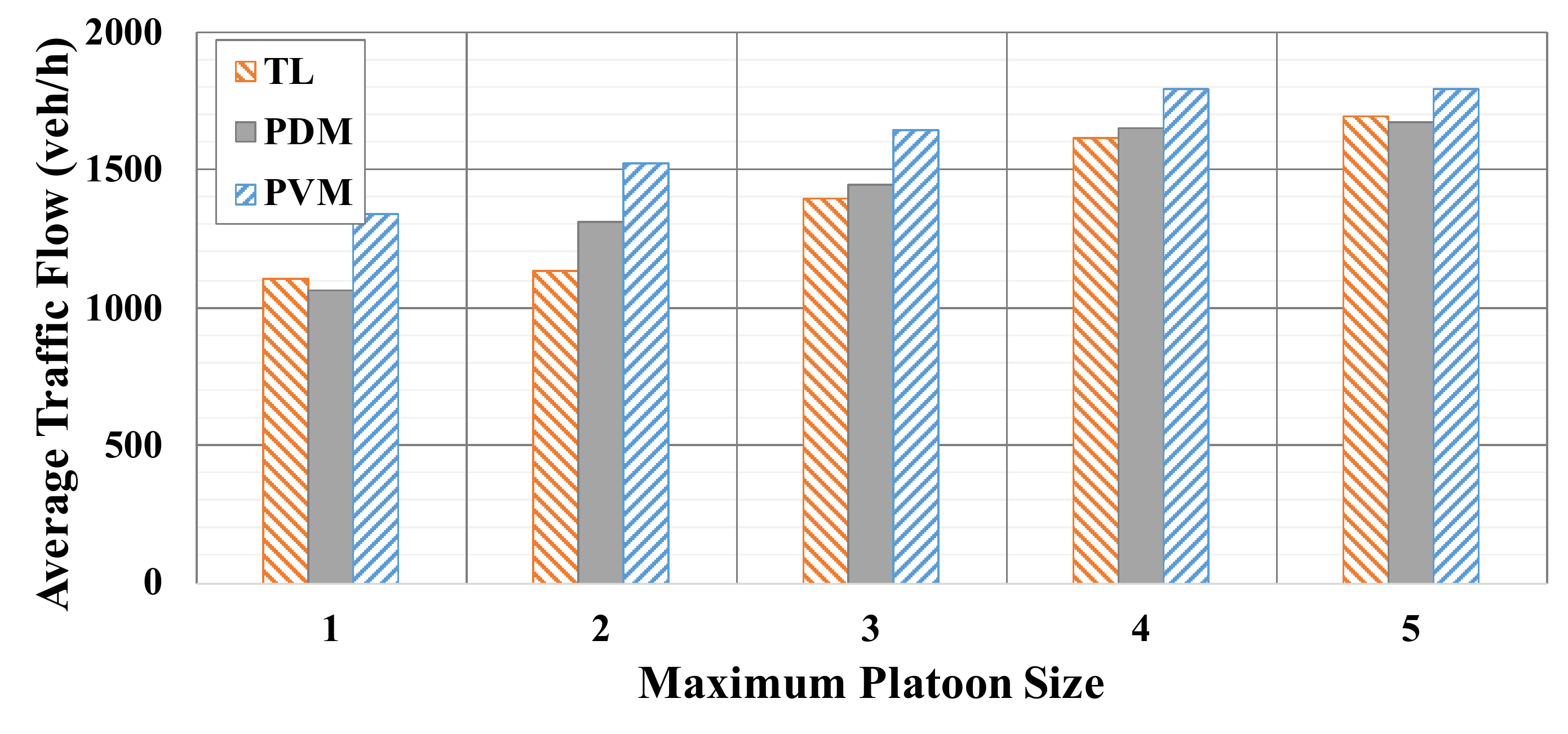}
    \caption{Intersection Traffic Flow}\label{fig:3}
    
            \vspace*{\floatsep}

    \includegraphics[width=0.7\textwidth]{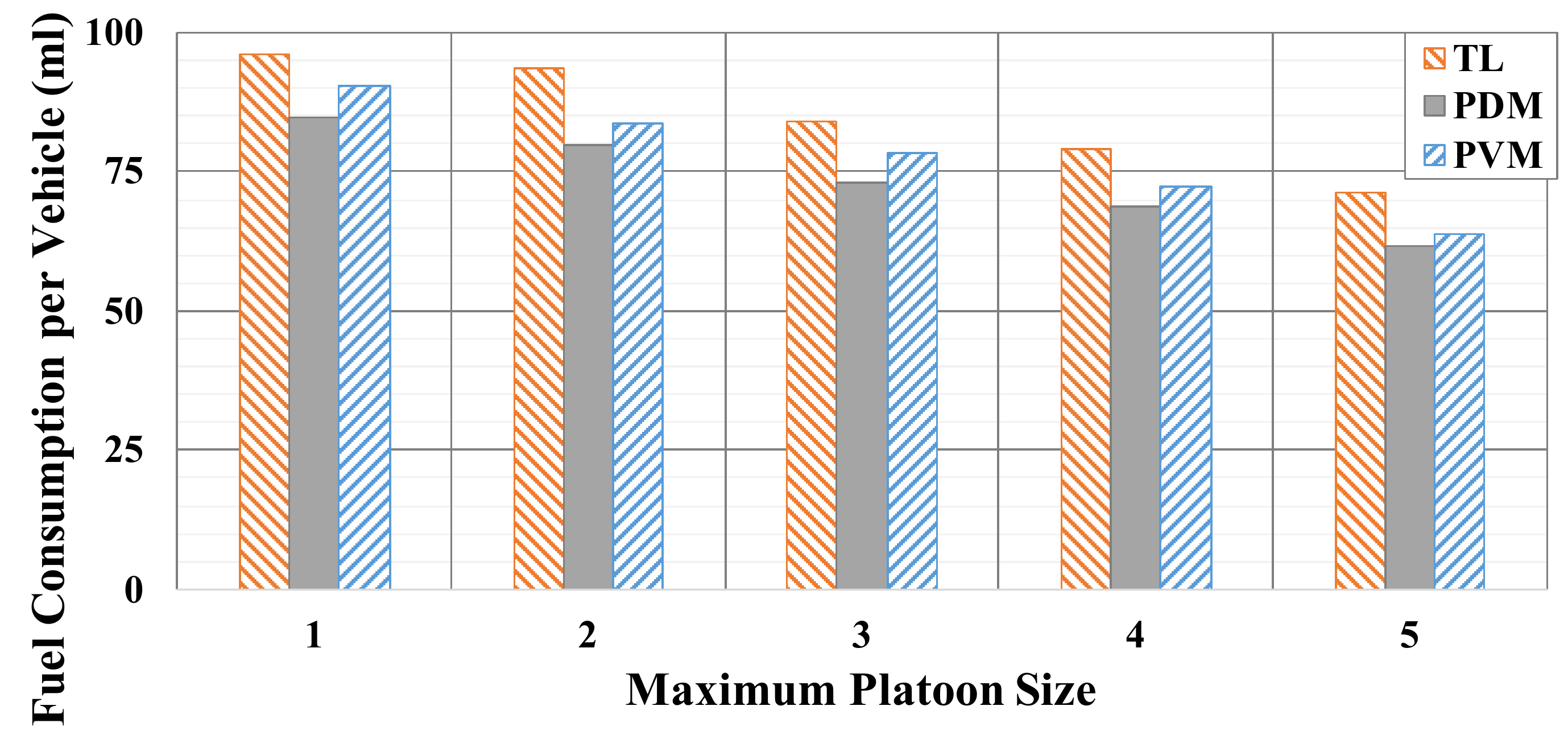}
    \caption{Fuel Consumption}\label{fig:4}
    
\end{figure}

Figs~\ref{fig:1},~\ref{fig:2},~\ref{fig:3} and~\ref{fig:4} demonstrate the results in terms of delay per vehicle, delay standard deviation, intersection capacity and fuel consumption, respectively. To conserve space and promote readability, the results are aggregated over the set of values for the traffic level parameter.

According to Fig.~\ref{fig:1}, the proposed methods significantly outperform the traffic light controller in terms of average delay per vehicle. Fig.~\ref{fig:2} shows the computed standard deviation of delays throughout the entire simulations for each policy and the set of maximum platoon sizes. It can be seen that the~\emph{PVM} method significantly decreases the standard deviation compared to the traffic light, while as expected,~\emph{PDM} does not show a meaningful improvement in terms of delay variance as its cost function is designed to solely decrease total delay. 

One may also note that the maximum platoon size and the~\emph{PVM} and the traffic light performance are negatively correlated, which confirms the positive effect of platooning on the performance of any type of traffic controller.

Fig.~\ref{fig:3} shows that platoon size and intersection capacity are positively correlated for all policies. For the traffic light policy, larger platoons result in more smooth trajectories with shorter headways, and as a result increases the outgoing traffic flow. Larger platoons also help the proposed policies select better schedules in terms of total delay and variance, which ultimately will increase the outgoing traffic flow. In the simulations, \emph{PVM} policy outperformed traffic light for all incoming traffic flows and platoon sizes.

Fig.~\ref{fig:4} shows fuel consumption per vehicle as a function of platoon size. As expected, platoon size and fuel consumption are strongly correlated. \emph{PVM} and~\emph{PDM} policies outperform traffic light by~$8\%$ and~$13\%$ on average, respectively. This result can be explained by the shorter idle times generated by~\emph{PVM} and~\emph{PDM} compared to the traffic light policy. According to the fuel consumption model adopted in the simulations, the vehicles consume fuel at a rate of~$0.15~ml/s$ when idle.
\begin{table*}[h]
\caption{Aggregated Results}
\label{tab:final}
\centering
\begin{tabular}{p{0.2\linewidth}p{0.10\linewidth}p{0.10\linewidth}p{0.10\linewidth}p{0.15\linewidth}p{0.15\linewidth}}
\hline
 & Traffic Light & PVM               & PDM      & PVM-Improvement & PDM-Improvement \\
 \hline
Delay(s)              & $43.26$       & $\bm{6.56}$   & $22.71$  & $\bm{6.6\times}$     & $1.9\times$    \\ 
Capacity(veh/h)        & $1388$      & $\bm{1617}$ & $1426$ & $\bm{13.8\%}$      & $2.7\%$ \\ 
FCPV(ml/v)             & $84$      & $77$ & $\bm{73}$ & $8\%$ & $\bm{13\%}$      \\ 
STDEV(s) & $31.44$         & $\bm{6.37}$     & $30.48$    & $\bm{4.9\times}$ & $3\%$  \\    
\hline
\end{tabular}
\vspace{-11pt}
\end{table*}

To compare the overall performance of the policies, results from all configurations are aggregated into table~\ref{tab:final}. All the metrics in this table are averaged over the set of incoming traffic flows that range from~$500$ to~$800~v/h/l$. Traffic flows are identical for each approach.

The~\emph{PVM} method outperforms the traffic light policy on all four metrics. More notably, it decreased average delay per vehicle by factor of~$6.56\times$ and decreases the standard deviation to~$4.9\times$, resulting in faster and more reliable traffic flows. The~\emph{PVM} policy also increased the intersection capacity by~$13.8\%$ compared to traffic light.

The~\emph{PVM} and~\emph{PDM} policies both outperform the traffic light in terms of fuel consumption by~$8\%$ and~$13\%$ respectively~\cite{bashiri2018paim}.
\section{Conclusions}\label{sec:Conclusion_3}
In this chapter, a centralized platoon-based controller was proposed for the cooperative intersection management problem that takes advantage of the platooning systems and V2I communication to generate fast and smooth traffic flow at a single intersection. A simple communication protocol was designed for V2I communication and two policies were introduced for the controller to minimize total delay and delay variance according to the cost functions tailored for platoons of vehicles.

According to the simulation results, the proposed controller minimizes travel delay and variance while increasing intersection throughput and reducing fuel consumption, when compared to traffic light policies. The simulations also verify the positive effect of platoon size on fuel consumption and intersection throughput.

There are several limitations to the proposed method that became a motivations for the research leading to the methods proposed in the next chapters. The main limitation is the assumption of market penetration rate of CAVs being at 100\% which is unrealistic at least for the next few decades. The proposed method cannot adapt to the presence of human-driven vehicles or other vehicle road users incapable of communicating to the controller.

Other limitations of this work include the simplistic assumptions about platoon formation/merging and vehicle dynamics and control, lack of consideration for emergency situations such as accidents or communication issues, and no formal guarantee for safety. The latter is addressed in the next chapter by adapting a traffic signal approach to the problem which is known to be formally safe by not granting the simultaneous entrance to vehicles with conflicting trajectories.

\chapter{Optimal Signal Timing}\label{chpt:signal}
This chapter will focus on the optimal signal timing problem. We identify the main limitations of previous work and discuss how the proposed data-driven method addresses each of these concerns.

\section{Motivations}\label{sec:signal_motivation}
As mentioned in the previous chapter, CIM methods (including the proposed method in Chapter~\ref{chpt:cim}) have several limitations:
\begin{itemize}
    \item Unrealistic or infeasible bandwidth requirements for communication
    \item No performance guarantees, i.e. no guarantee that the intersection will behave better than signalized intersections
    \item No formal safety guarantee
    \item Scalability with respect to number of cars and lanes (High Computational Complexity)
    \item Unrealistic assumptions about vehicle behavior
\end{itemize}

Another issue with the~CIM approach is the long transition period within which autonomous and human-driven vehicles will co-exist on the roads, which calls for introduction of novel methods that can handle mixed-traffic situations, while leveraging autonomy and communication.

This chapter proposes a traffic signal based approach to guarantee safe trajectories with no conflict of paths. It is based on realistic assumptions about v2I communications, relying on basic safety messages (BSM) as defined in SAE J2735 message set dictionary~\cite{dsrc}. BSMs convey basic vehicle state information necessary to support vehicle safety applications. The proposed method is capable of adopting BSMs to improve its traffic state estimations.

The proposed method trains a learning model based on data generated through VISSIM, an industry standard traffic simulation software that uses detailed and realistic vehicle dynamics and driver behavior models. Microscopic models provide further detailed information about individual vehicles in traffic when compared to macroscopic models that rely on aggregating individual vehicle state information.

Moreover, the complexity of the proposed model is not a function of traffic level or the size of the intersection, hence, eliminating the scalability issues of some of the previous work in the CIM literature.
\section{Data-Driven Signal Timing}\label{sec:ddst}
To achieve optimal signal timing plans, we propose a data-driven method that relies on building a model utilizing recorded traffic data from the network. As a traffic signal based method, it provides the safety measures inherent to a traffic signal; namely no vehicles with conflicting trajectories will be allowed to cross the intersection simultaneously. It relies on Vissim's vehicle dynamics and driving agent model, which are more realistic compared to the macroscopic models and second order dynamics models in the literature. Moreover, the proposed signal timing model's computational complexity is agnostic to the level of traffic or the size of the network, making it a suitable for large and busy urban networks.

The proposed method consists of three stages of \textit{Scenario Design}, \textit{Data collection and Preparation}, and \textit{Model training}. The entire source code of this approach is available online\footnote{\href{https://github.com/ashkanbashiri/data_driven_signal_control}{source code}}. The data generation process is described in pseudo code in algorithm~\ref{alg:dg}. 

\begin{algorithm}[ht]
\caption{Data Generation Algorithm}
\label{alg:dg}
\begin{algorithmic}
    \State $T \gets getTrafficValues()$
    \State $C \gets getCycles()$
    \State $L \gets getLostTimes()$
    \ForAll{$t \in T$} 
        \ForAll{$c \in C$}
            \ForAll{$l \in L$}
                \State $G\gets genGreenTimes(t,c,l)$
                \ForAll{$g \in G$}
                    \State $result\gets simulate(t,c,l,g)$
                    \State $save(result)$
                \EndFor
            \EndFor
        \EndFor
    \EndFor
\end{algorithmic}
\end{algorithm}

\subsection{Scenario Design}
Scenario design is the process of developing scenarios to cover traffic conditions that the network is expected to experience. The first step in designing scenarios is to identify all the variables that affect the traffic controller's performance. Incoming traffic flows, cycle length, green times and intersection lost time were considered as the main variables contributing to the intersection controller's performance. Scenarios were designed for three isolated~$4-way$ intersections to cover traffic flows in the range of under-saturated conditions to saturated conditions, various total lost times in the range of~$(4s,24s)$ and all combinations of the scenario variable values. The interactions between traffic flow variables were designed to cover a range of scenarios from highly unbalanced to completely balanced with regards to lane group traffic volumes. The value ranges used in the simulations are presented in table~\ref{tbl:sim_params}.

\begin{table*}[ht]
\centering
\caption{Data collection scenario settings}
\label{tbl:sim_params}
\resizebox{\textwidth}{!}{%
\begin{tabular}{@{}ccccc@{}}
\toprule
\multicolumn{5}{c}{Input Variables} \\ \midrule
 & Notation & Range of Values & Number of Values & unit \\ \midrule
Total traffic & F & $(0.3, 1) \times s \times w$ & 7 & veh/hour \\ \midrule
Lane group traffic flow & $f_{i}$ & $(0.1,0.7)\times F,  \sum f_{i}=F$ & 296 & veh/hour \\ \midrule
Total lost time & L & 8 & 1 & s \\ \midrule
Cycle length & C & \{40, 50, ..., 190, 200\} & 17 & s \\ \midrule
Effective Green time & $g_{i}$ &$ \{\frac{1}{4}, \frac{f_{i}}{F}, rnd_{1}, rnd_{2}\} \times (C-L) $& 4 & s \\ \midrule

\multicolumn{5}{c}{Simulation Settings} \\ \midrule
\multicolumn{2}{c}{Parameter} & Notation & Value & Unit \\ \midrule
\multicolumn{2}{c}{Simulation time} & T & 60 & Minutes \\ \midrule
\multicolumn{2}{c}{Saturation flow rate (ideal)} & s & 1900 & veh/hour/lane \\ \midrule
\multicolumn{2}{c}{Intersection width (\# of lanes)} & w & \{4,4,6\} & lane \\ \midrule
\multicolumn{2}{c}{\# of Scenarios} & Sc & 140896 & - \\ \midrule
\multicolumn{2}{c}{\# of optimal scenarios (dataset size)} & $n_{opt}$ & 2072 & - \\
\bottomrule
\end{tabular}%
}
\end{table*}

\subsection{Intersection Model}
\begin{figure}[!htbp]
\hspace*{\fill}

    \centering
    \begin{minipage}{0.4\textwidth}
        \centering
    \includegraphics[width=3in]{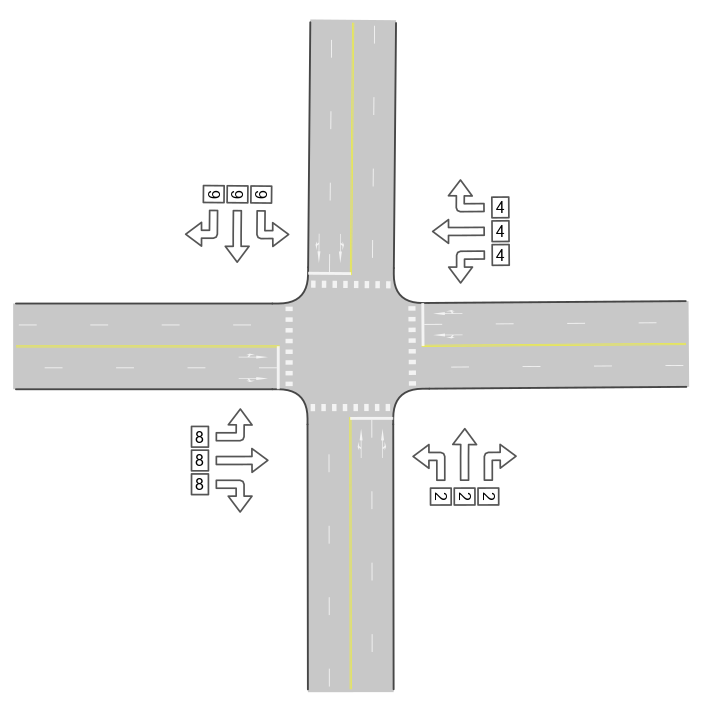}
    \caption{Intersection Model 1 in Vistro}
    \label{fig:snapshot}
    \end{minipage}\hfill
    \begin{minipage}{0.4\textwidth}
        \centering
        \vspace{0.8in}
    \includegraphics[width=3in]{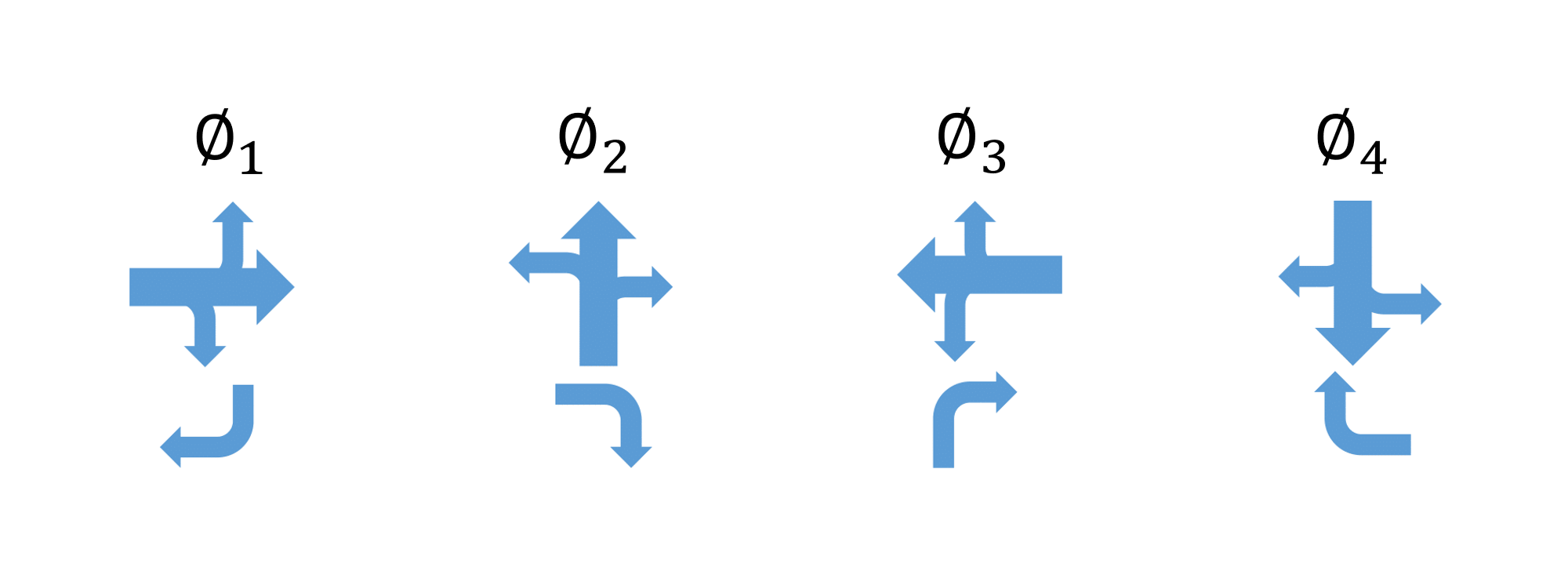}
    \vspace{0.7in}
    \caption{Phase Diagram of the Signal Controller 1}
    \label{fig:phase_diag}
    \end{minipage}
\hspace*{\fill}
\end{figure}

\begin{figure}[!htbp]
\hspace*{\fill}

    \centering
    \begin{minipage}{0.4\textwidth}
       \centering
    \includegraphics[width=3in]{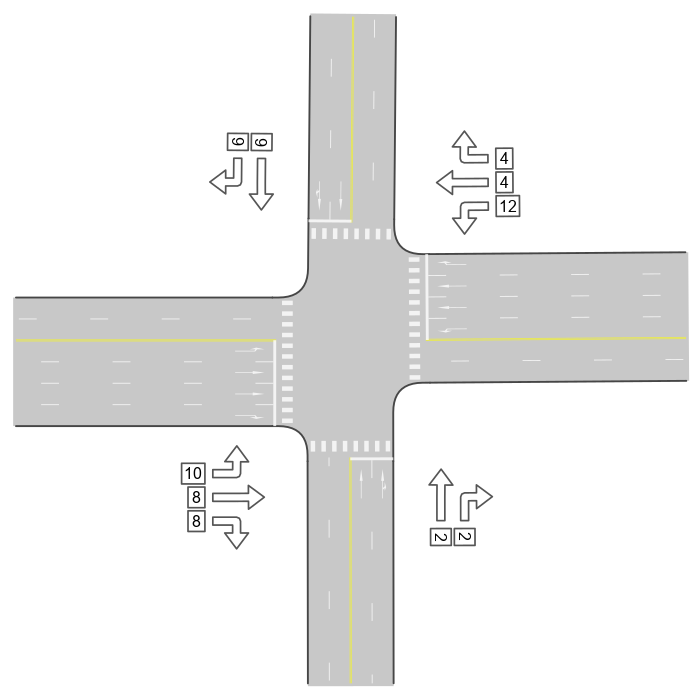}
    \caption{Intersection Model 2 in Vistro}
    \label{fig:snapshot2}
    \end{minipage}\hfill
    \begin{minipage}{0.4\textwidth}
     \centering
             \vspace{0.8in}

    \includegraphics[width=3in]{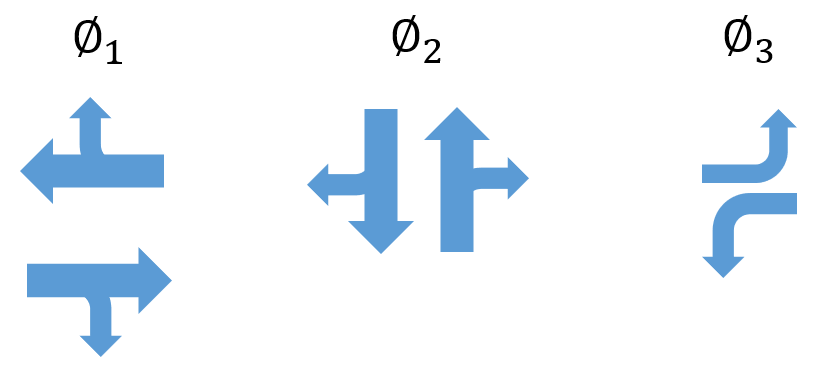}
        \vspace{0.7in}

    \caption{Phase Diagram of the Signal Controller 2}
    \label{fig:phase_diag2}
    \end{minipage}
    \hspace*{\fill}

\end{figure}

\begin{figure}[!htbp]
\hspace*{\fill}

    \centering
    \begin{minipage}{0.4\textwidth}
     \centering
    \includegraphics[width=3in]{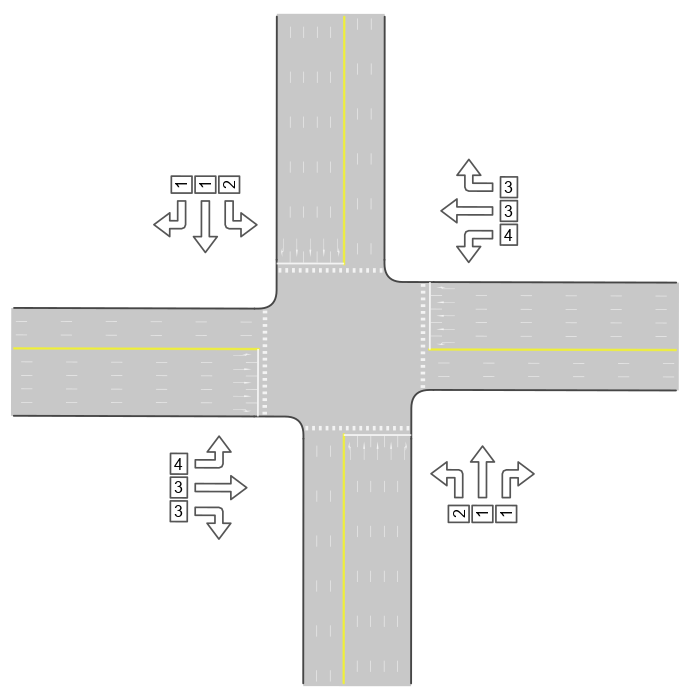}
    \caption{Intersection Model 3 in Vistro}
    \label{fig:snapshot3}
    \end{minipage}\hfill
    \begin{minipage}{0.4\textwidth}
      \centering
              \vspace{0.7in}

    \includegraphics[width=3in]{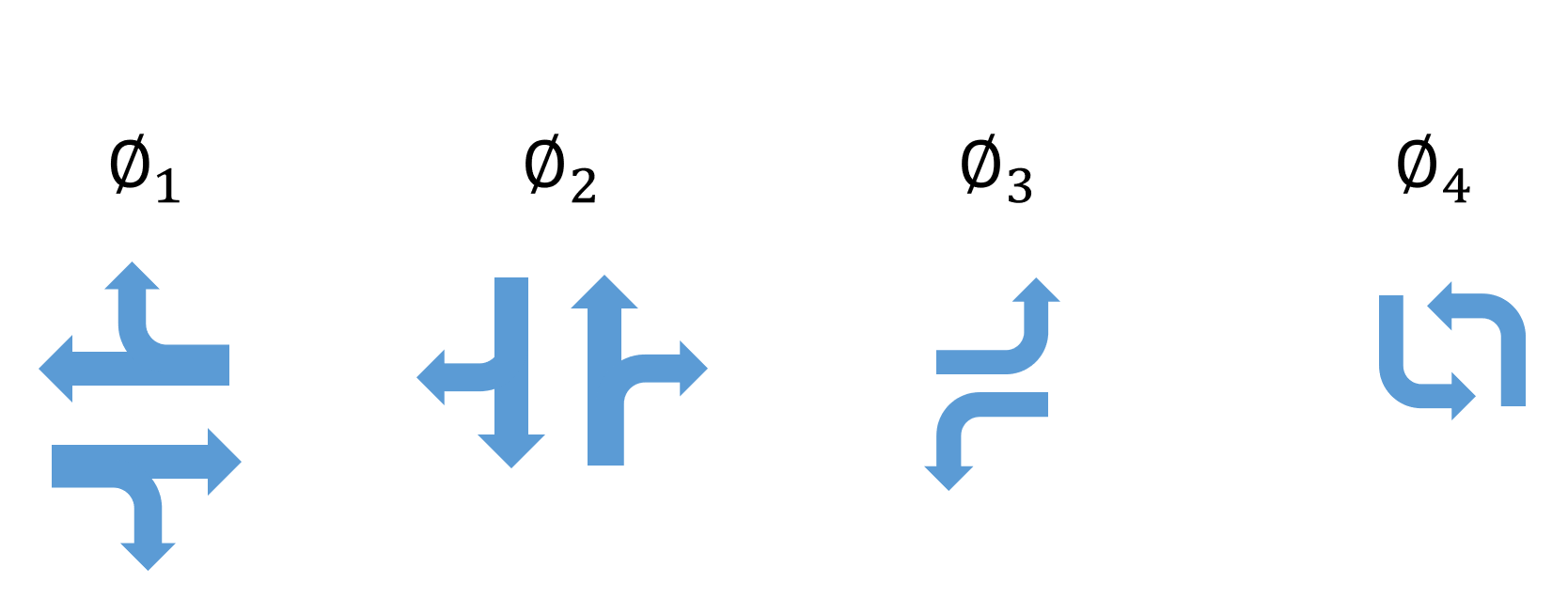}
            \vspace{0.9in}

    \caption{Phase Diagram of the Signal Controller 3}
    \label{fig:phase_diag3}
    \end{minipage}
        \hspace*{\fill}

\end{figure}

Three intersection models were designed for this work. Figs.~\ref{fig:snapshot},~\ref{fig:phase_diag},~\ref{fig:snapshot2},~\ref{fig:phase_diag2},~\ref{fig:snapshot3} and~\ref{fig:phase_diag3} demonstrate the intersection models in PTV Vissim~\cite{vissim} and the phase diagrams of the signal controllers for the three intersections respectively.

The first intersection is controlled by a 4-stage signal controller, where each stage is dedicated to one leg of the intersection. The second intersection is a larger network with a different setup in terms of vehicle routes and is controlled by a 3-stage signal controller as depicted in Fig.~\ref{fig:phase_diag2}. The third intersection is controlled with a~$4-phase$ signal controller with two protected phases for the left turn movements.

\subsection{Data Collection}
All scenarios were simulated in Vissim through the Component Object Model (COM) interface. Vissim reports various performance metrics at the end of each simulation. All reported metrics along with the configurations were recorded for each of the simulation scenarios. The collected data would later be filtered to contain only best results with regard to a selected performance metric and be set up as the training set for the learning model.

\subsubsection{Filtering}
The final step of data collection is to filter the raw data to only include best set of variables in terms of the selected objective. In this work, average vehicular delay is selected as the performance metric to optimize. However, one has the choice to select other performance metrics that are reported for any simulation in the Vissim software.

Table~\ref{predictors} lists the predictor variables and possible target variables (metrics). One should note that the choice of recording multiple target variables makes it possible to build multiple optimal cycle models based on one or a combination of these variables.
\begin{table}[ht]
\caption{Predictor variables and possible target variables.}
\centering
\begin{tabular}{|c|c|c|}
\hline
\multicolumn{3}{|c|}{\textbf{Predictor Variables}}       \\ \hline
Variable Name                & Notation & Unit  \\ \hline
Lane Group Traffic Flow      & $f_i$     & 1/sf  \\ \hline
Lost Time                    & $L$        & s     \\ \hline
Total Traffic Flow           & $F$        & 1/sf  \\ \hline
\multicolumn{3}{|c|}{\textbf{Target Variables}}          \\ \hline
Delay Per Vehicle            & $d$        & s     \\ \hline
Stop Delay Per Vehicle       & $d_s$     & s     \\ \hline
Fuel Consumption per Vehicle & $fc$       & liter \\ \hline
Intersection Throughput        & $t$        & veh/s \\ \hline
Delay Standard Variation     & $d_{std}$   & s     \\ \hline
\end{tabular}
\label{predictors}
\end{table}

\subsection{Signal Timing Model}
The next step is to build a model that estimates appropriate green times given the estimated values of the predictor variables, i.e. the traffic flows and total lost time of the intersection. This problem is known as multi target regression~(MTR).

It should be noted that while the predictors and target variables above are well established as the input/output variables to a signalized intersection controller, one could restructure the problem definition and establish a different set of predictor and target variables. For example, an end-to-end solution to the problem could consider the predictor variables to be total incoming traffic and some type of sensor readings or communicated information from vehicles. However, this dissertation adopts a modular design where traffic flow estimation and signal timing are divided into two separate modules. The choice of modular design offers several benefits compared to a monolithic design. The benefits include minimizing safety-related issues, reducing costs to deploy and maintain and fewer resource requirements.

Reinforcement learning (RL) is among popular approaches in the CIM literature where an intersection or a network of intersections are controlled using reinforcement learning technique. The intersection controllers learn value functions with the aim of choosing policies that optimize metrics such as the average delay and congestion.

\cite{arel2010reinforcement} proposed a method that utilises the Q-Learning algorithm \cite{watkins1989h} with a feedforward neural network for value function approximation. A two ring structure is assigned to each of the intersections with each ring consisting of four phases. The policies enable the controller to select an action among the eight-phase combination schemes.

The above approach has proved to be effective in simulations; however, defining the problem as selecting optimal policies makes the solution hard to interpret and as a result does not provide much insight to a traffic engineer. Furthermore, real-time selection of phase-combinations could cause confusion making the approach unappealing to drivers. The proposed approach in this dissertation however, provides a signal timing model that is easy to gain insights from and makes no sudden changes to the policy.

Given the problem definition and the fact that all input and output variables are scalars, a fully connected neural network is a natural candidate for the learning model. However, to ensure the performance, several models were built using techniques such as linear regression \cite{lin_reg}, gradient boosting decision trees \cite{gbdt} and deep neural networks \cite{deep_nn}. After comparing the accuracy and robustness of the trained models, a neural network (Multi Layer Perceptron) was selected to build the model.

\subsubsection{Learning Model}
To select the exact structure of the multi-layer perceptron we compared several models with an emphasis on maximizing the convergence rate and minimizing the computational overhead in the training phase.

\begin{figure}[ht]
    \centering
    \includegraphics[width=4in]{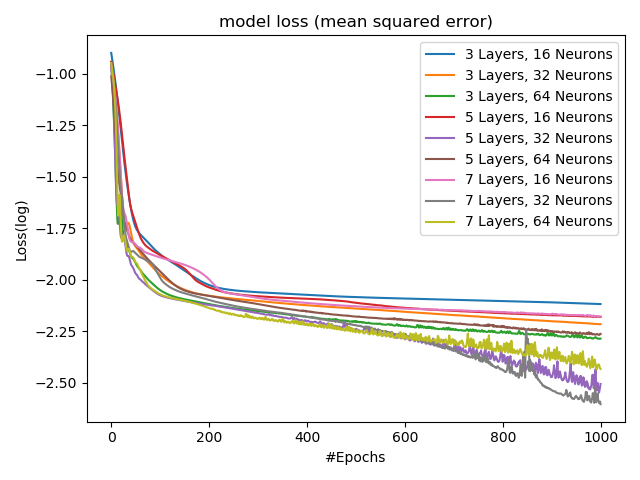}
    \caption{Model's training loss}
    \label{fig:model_loss}
\end{figure}

Fig.~\ref{fig:model_loss} compares the convergence rate of different model structures for the first intersection. The graph represents logarithmic loss values (as defined in Equation~\ref{eq:mse}) throughout~$1000$ epochs. According to Fig.~\ref{fig:model_loss} the model with~$7$ hidden layers and~$32$ neurons in each layer has the highest convergence rate, however, we picked the model with the second fastest convergence due to its lower computational complexity and the fact that it achieves similar performance when compared using the test set. 

Similar learning models were selected for the other two intersections, with the exception of the second model's number of neurons in the output layer that was set to~$3$ to represent the three phases of the signal controller.

\begin{figure*}[ht]
  \centering
  \includegraphics[width=\textwidth]{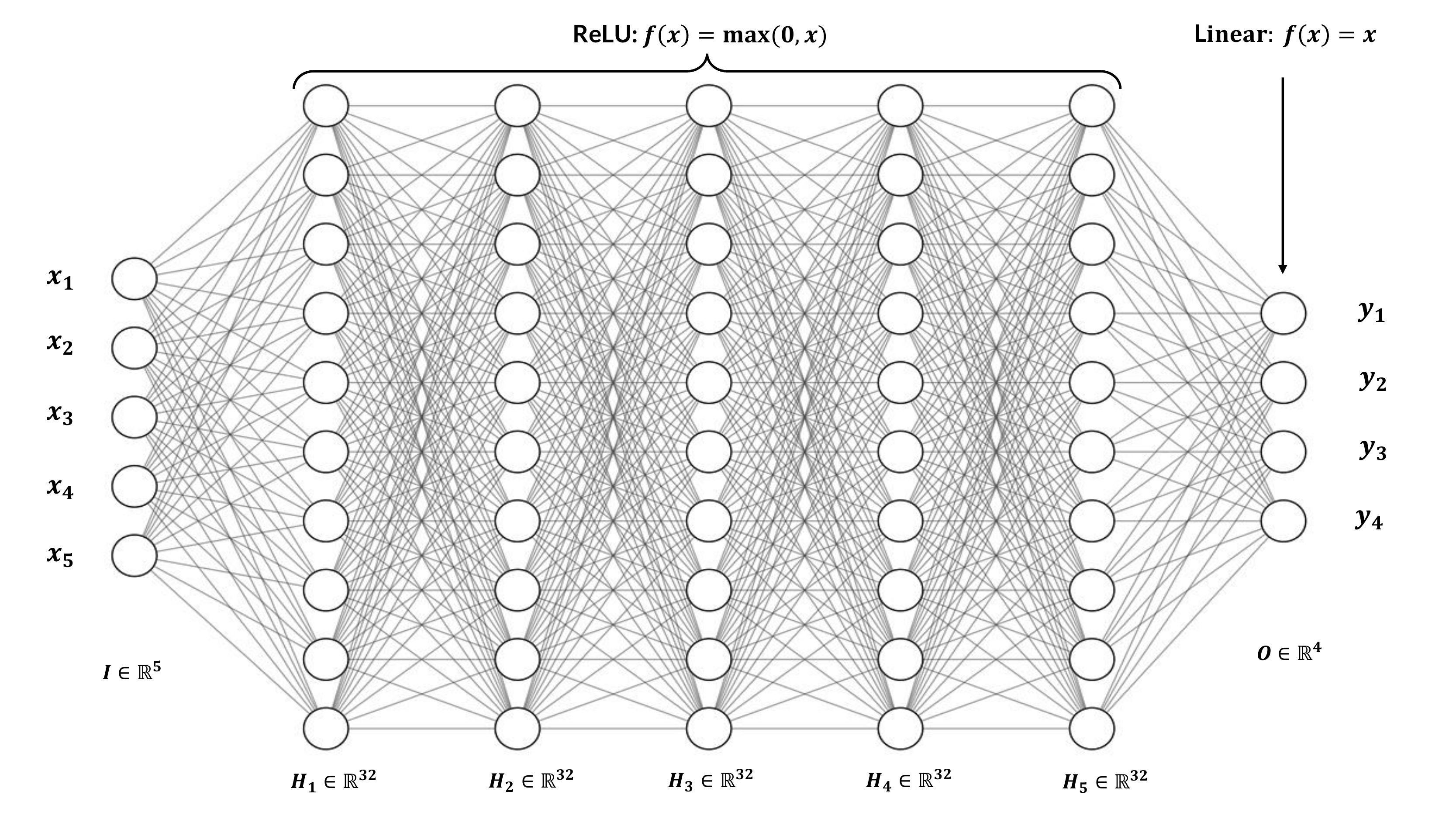}
  \caption{Schematic Diagram of the Training Model}
  \label{fig:nn_arch}
\end{figure*}

Fig.~\ref{fig:nn_arch} demonstrates the structure of the selected multi layer perceptron (MLP) that was designed as the learning model. It should be noted that while the general structure of the network and the chosen activation functions remain the same regardless of the intersection model, the exact number neurons at the output layer are different for a given intersection, e.g. they match the number of phases designed for the intersection that the network is designed for. The numbers shown in this figure represent the parameters optimized for the first intersection's signal timing model. 

Furthermore, the number of epochs for training was separately determined based on the early stopping method for each intersection. Each intersection's data set was divided into 70\% training set and 30\% validation set. The network was trained with the training set and evaluated with the hold out validation set until the network's performance (MSE) stopped improving. The epoch at which the model stops improving (stuck in minima) was selected as the number of epochs to train on the entire dataset for final simulations. 

The network consists of an input layer with five neurons that feed the input variables into the network, an output layer with four neurons that output the four estimated optimal green times, and finally five hidden layers of 32 neurons. The number of hidden layers and the number of neurons in each layer was carefully chosen to avoid an unnecessarily complex structure while preserving an acceptable performance in terms of the training loss. In addition, the selection of a shallower model is less likely to result in overfitting.

Rectified linear unit (ReLU) was selected for activation of hidden layer neurons. The choice of ReLU helps with the sparsity of the activation of neurons but mapping negative values to zero. This will help in decreasing the number of neurons firing, especially when compared to sigmoid type activation functions such as $tanh$ that cause almost all neurons to fire.

Since the objective is to build a regression model, mean squared error (Equation~\ref{eq:mse}) was selected as the loss function. Furthermore, the input variables are normalized, and linear activation (a.k.a. no activation) is utilized on the output layer.
\begin{equation}
    \label{eq:mse}
     MSE = \frac{1}{n}\Sigma_{i=1}^{n}{(y_{i} - \Tilde{y}_{i})^2}
\end{equation}

In the above equation, $y_{i}$ and $\Tilde{y}_{i}$ represent ground truth and predicted values for the $i_{th}$ phase's green time.

\subsubsection{Computational Complexity of The Model}
We divide the computational complexity of the proposed model into two categories of \emph{training complexity} and~\emph{testing complexity}. The \emph{training complexity} refers to the computational complexity of training the model which involves the complexity of the feed-forward pass algorithm and the back-propagation algorithm. The combined computational complexity can be written as:
\begin{equation}
    \label{eq:train_cc}
    O(nt\times \sum_{i=1}^{L-1} \| l_i \| \times \|l_{i+1}\|)
\end{equation}
Where,~$n$ is the number of epochs,~$t$ is the number of training samples,~$L$ is the number of layers, and~$\|l_{i}\|$ refers to the number of neurons in the~$i$th layer. Omitting the constants in equation~\ref{eq:train_cc} results in:
\begin{equation}
    \label{eq:train_cc_trimmed}
    O(nt)
\end{equation}
One should note that the time complexity calculated above is for the training of the model and will not affect the performance of the model when deployed. The time complexity of the model in testing and deployment phase, here referred to as the~\emph{testing complexity} can be written as the following:
\begin{equation}
    \label{eq:test_cc}
    O(\sum_{i=1}^{L-1} \| l_i \| \times \|l_{i+1}\|)
\end{equation}
Equation~\ref{eq:test_cc} shows that the proposed model has a linear time complexity in deployment phase, i.e. it is real-time responsive.
\subsection{Handling Input Noise}\label{sec:input_noise_sensitivity}
An important aspect of a regression model is its robustness against input noise. We specially care about this because the input to our model will be estimates of the traffic's state through sensor readings and communications from CAVs as will be discussed in the next chapter.

We perform variance-based sensitivity analysis~\cite{sens} on the first intersection's trained model to understand how the uncertainty in the model's predictions can be apportioned to the uncertainty in the predictors. Same analysis was performed on the other intersection models that yielded similar results. Variance-based sensitivity analysis allows full exploration of the input space, which in the case of our model includes the estimations of incoming traffic flows as well as the estimated total lost time.

\subsubsection{First Order and Total Order Indices}
Given a model of the form~$Y = f(X_{1},X_{2},...,X_{k})$, with $Y$ a scalar, a variance based first order effect for a generic factor $X_{i}$ can be written as:
\begin{equation}
    \label{eq:foe}
    V_{X_{i}}(E_{X_{\sim i}}(Y|X_{i}))
\end{equation}
In equation~\ref{eq:foe},~$X_{i}$ is the~$i-th$ factor and~$X_{\sim i}$ denotes all factors but~$X_{i}$. The inner expectation operator can be interpreted as the mean of output~$Y$ taken over all possible values of~$X_{\sim i}$, while keeping~$X_{i}$ fixed; and the outer variance is taken over all possible values of~$X_{i}$. The associated sensitivity measure called first-order sensitivity index, is written as equation \ref{eq:foi}. First-order index measures the main contribution of an input parameter to the output without considering its interaction with the other input parameters.
\begin{equation}
    \label{eq:foi}
    S_{i} = \frac{V_{X_{i}}(E_{X_{\sim i}}(Y|X_{i}))}{V(Y)}
\end{equation}
To understand the total contribution of each input variable, we also study the total effect index defined in equation~\ref{eq:toi}.
\begin{equation}
    \label{eq:toi}
    \begin{split}
    S_{Ti} &= \frac {E_{X{\sim i}}(V_{X_{i}}(Y|X_{\sim i}))}{V(Y)} \\
    &= 1 - \frac {V_{X_{\sim i}}(E_{X_{i}}(Y|X_{\sim i}))}{V(Y)}
    \end{split}
\end{equation}
The total effect index captures the contribution of all terms in the variance decomposition which do include~$X_{i}$.
\subsubsection{Analysis Results}
Input samples were generated to cover the range of values that were considered in the data generation phase. Since all input and output values were normalized before used in training, a range of~(0,1) was selected for each of the input variables. Since the output of the trained model is a vector of size four, representing the four green times, sensitivity indices were averaged over the output scalars.

\begin{figure}[ht]
    \centering
    \includegraphics[width=0.8\textwidth]{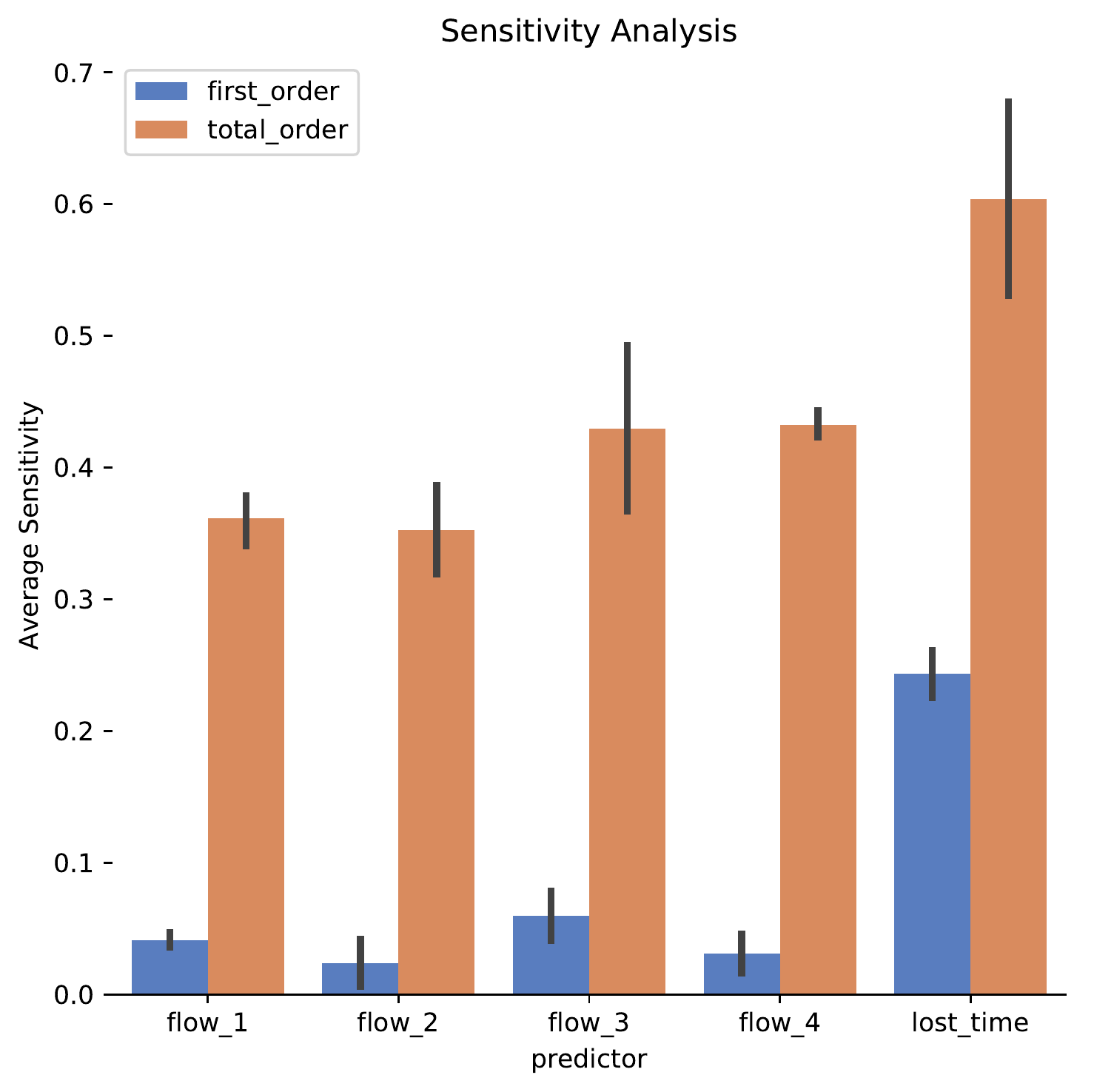}
    \caption{Sensitivity analysis results of the model.}
    \label{fig:sens}
\end{figure}

Fig.~\ref{fig:sens} demonstrates the contributions of input variables to the uncertainty in the predicted optimal green times in the form of first-order and total-order index as defined in equations~\ref{eq:foi} and~\ref{eq:toi}.
The first-order sensitivity index for traffic flows is significantly low which suggests low sensitivity to error in estimations of incoming traffic. Moreover, cycle lost time is the main contributor to the variance in estimated green times. This makes intuitive sense, since green times are expected to grow as the lost time increases. The sensitivity analysis results verify the model's robustness against noisy traffic flow estimations which is the case in real-world scenarios. It should be noted that high sensitivity to the \emph{lost\_time} input variable will not take away from the model's robustness since this variable is a part of the design of the intersection and its controller and does not include any variable noise in practice.

\section{HCM Analysis and Comparisons}{\label{sec:comp}}
A set of scenarios were designed to compare the performance of the trained models with that of the HCM methodology. The scenarios were designed to reflect different levels of traffic volume and the imbalance of the incoming traffic. The results show that the trained models were able to outperform the HCM model in terms of average vehicular delay. 

\subsection{HCM Methodology and Analysis of its Optimal Cycle Length Formula}
\begin{figure*}[!ht]
    \centering
    \includegraphics[width=\textwidth]{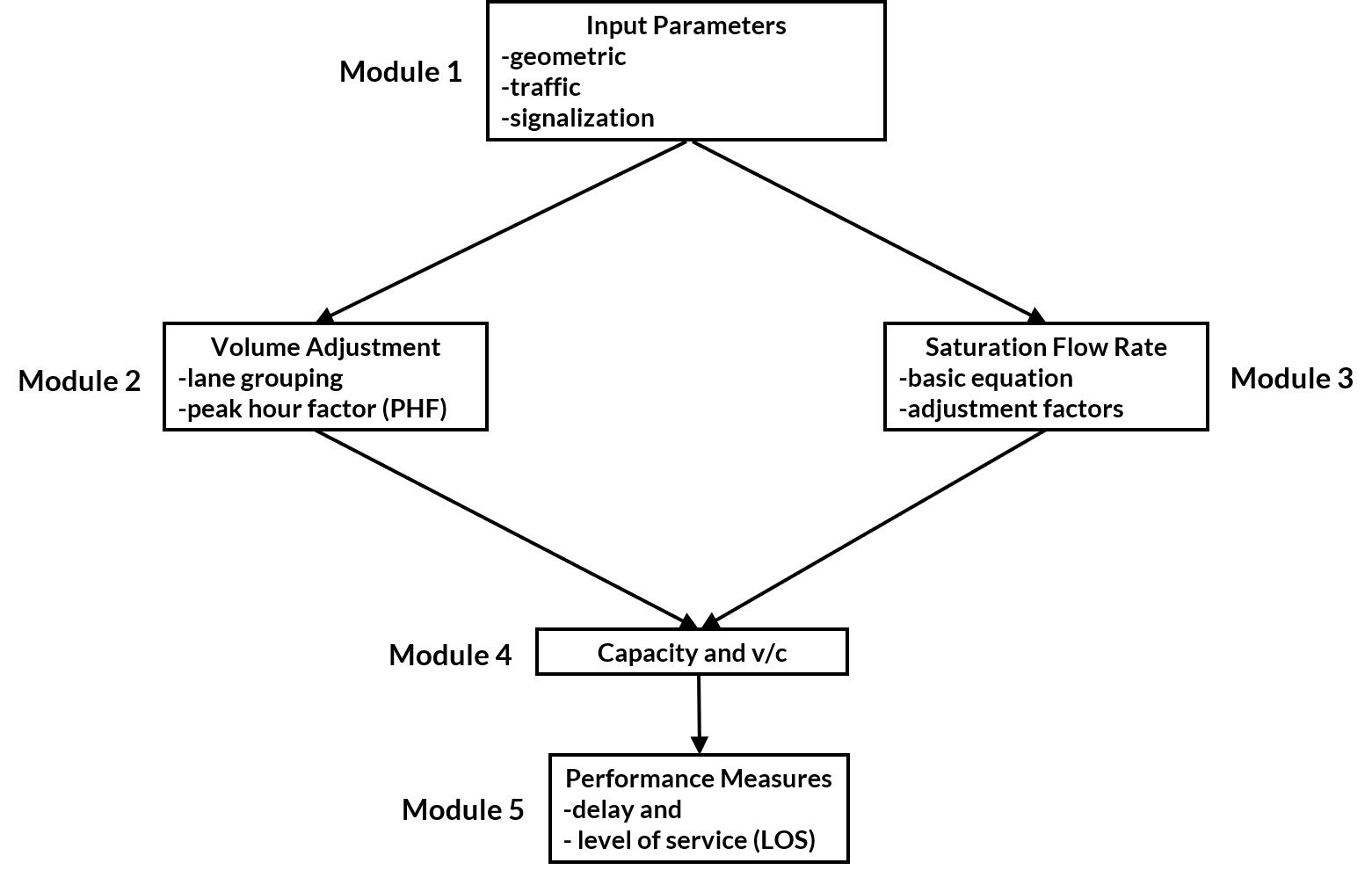}
    \caption{HCM Model Structure}
    \label{fig:hcm}
\end{figure*}

The proposed model's performance was compared to that of the HCM analysis procedure~\cite{hcm6} which provides estimates of saturation flow, capacity, delay, level of service~(LOS), and back of queue by lane group for each approach. PTV Vistro macroscopic simulator offers a selection of analysis methodologies such as the Highway Capacity Manual~(HCM)~2000,~2010 and~6th Edition. HCM~6th edition was used as the compared methodology. Fig.~\ref{fig:hcm} demonstrates the HCM intersection analysis model structure that consists of five modules.

Module~1~(Input) considers parameters such as traffic, geometric and signalization conditions. Module~2 is concerned with conversion of hourly demand volumes to peak 15-min flow rates and establishing the lane groups. Module~3 estimates saturation flow rates~$(s_{i})$ for each of the established lane groups. Volume to saturation flow ratios~$(v/s)_{i}$ and capacities~$(c_{i})$ for lane groups are computed in Module~4 to determine critical lane groups. Signal timings are modified according to the computed~$v/s$ ratios of the critical lane groups. The last module estimates performance metrics such as individual and aggregated control delays for each lane group.

HCM offers two objective functions for analysis of the signalized intersection; (1) balancing volume to capacity ratio and (2) minimizing critical movement delay. The latter was selected for the comparisons, since it yielded better results in terms of average delay.

The HCM method utilizes Webster's Formula~\cite{webster1958traffic} to estimate the optimal cycle length~(equation~\ref{cycle_length}):
\begin{equation} \label{cycle_length}
    C_{opt} = \frac{1.5L+5}{1-\sum_{i=1}^{n}(v/s)_{ci}}
\end{equation}
In the above equation,~$C$ is the estimated optimum cycle length to minimize average control delay.~$L$ and~$(v/s)_{ci}$ are total lost time per cycle and flow ratio for critical lane group~$i$, respectively.Moreover,~$v$ is the estimated flow rate for critical lane group~$i$ and~$s$ is saturation flow rate.

The average control delay experienced by all vehicles that arrive in the analysis period~($T$) can be determined by equation~\ref{delay_estimation}:
\begin{equation} \label{delay_estimation}
    d = d_1(PF) + d_2 + d_3
\end{equation}
where $PF$ is uniform delay progression adjustment factor and $d_1, d_2, d_3$ indicate uniform control delay, incremental delay, and initial queue delay, respectively. The uniform delay and incremental delay can be estimated as:
\begin{align} \label{uniform_delay}
    d_1 &= \frac{0.5C(g/C)^2}{1-[\min(1,1/X)g/C]}\\
    d_2 &= 900T \Big[(X-1) + \sqrt{(X-1)^2 + \frac{8klX}{cT}} \Big]
\end{align}
Where~$g$ denotes effective green time for lane group, and $X = v/c$ is ratio or degree of saturation for lane group, where $c$ shows the approach capacity.~$k$ and~$l$ are incremental delay and upstream filtering/metering adjustment factors.\\

The third term of lane group control delay refers to the delay due to a residual queue identified in a previous analysis period and persisting at the start of the current analysis period. This delay can be estimated as the following general form:
\begin{equation} \label{initial_queue_delay}
    d_3 = \frac{1800Q_b(1+u)t}{cT}
\end{equation}
where 
\[ t = \begin{cases} 
      0 & Q_b=0 \\
      \min\{T,\frac{Q_b}{c[1-\min(1,X)]}\} & o.w. 
   \end{cases}
\]
\[ u = \begin{cases} 
      0 & t<T \\
      1-\frac{cT}{Q_b[1-\min(1,X)]} & o.w. 
   \end{cases}
\]
$Q_b$,~$t$ and~$u$ are initial queue, duration of unmet demand and delay parameter, respectively.\\

In the following, an analytical study of HCM's formula for estimating the optimal cycle length (i.e. equation \ref{cycle_length}) is presented. This analysis formally studies the sensitivity of HCM's formula for optimal cycle length to its parameters.\\

We build our analysis based on the assumption that the values set for saturation flow rate parameter~$s$ for each lane group have been included with a margin of safety (i.e. saturation flow rates are underestimated). The margin of safety is shown by~$\alpha$ and the measured parameters are:
\begin{align}
    s_i &= (1-\alpha)s_i^r; \quad  0<\alpha<1, \quad \forall i \in \{1, \dots, n\}
\end{align}

$s^r$ is the actual measured values in the field.
\begin{theorem}
\label{theorem:1}
Involving margin of safety~$\alpha$ in~$s$ will cause an over-estimation in cycle length.
\end{theorem}
\begin{proof}
Based on the assumption, the estimated cycle length can be written as:
\begin{equation}
    C = \frac{1.5L+5}{1-\frac{1}{1-\alpha}\sum_{i=1}^{n}(v/s^r)_{ci}}
\end{equation}
and because, $\frac{1}{1-\alpha}>1$, it is concluded that
\begin{equation}
    C > C^r = \frac{1.5L+5}{1-\sum_{i=1}^{n}(v/s^r)_{ci}}
\end{equation}
\end{proof}

\begin{theorem}
\label{theorem:2}
Involving margin of safety $\alpha$ in $s$ will increase the cycle length over-estimation rate with respect to critical lane group's volume, $v_{ci}$.
\end{theorem}
\begin{proof}
\label{proof:2}
To calculate the rate of change in cycle length over-estimation with respect to critical lane group volume, we need to obtain $\frac{\partial}{\partial v_{ci}} (C - C^r)$: 
\begin{multline}
        \frac{\partial}{\partial v_{ci}} (C - C^r) =(1.5L + 5)\\
    \bigg(\frac{1/s_{ci}}{(1-\sum_{i=1}^{n}(v/s)_{ci})^2} -\frac{1/s^r_{ci}}{(1-\sum_{i=1}^{n}(v/s^r)_{ci})^2} \bigg)=\\
    \bigg(\frac{1/((1-\alpha)s_{ci}^r)}{(1- \frac{1}{1-\alpha} \sum_{i=1}^{n}(v/s^r)_{ci})^2} -\frac{1/s^r_{ci}}{(1-\sum_{i=1}^{n}(v/s^r)_{ci})^2} \bigg)
\end{multline}
where $1- \frac{1}{1-\alpha} \sum_{i=1}^{n}(v/s^r)_{ci} < 1-\sum_{i=1}^{n}(v/s^r)_{ci}$ and $\frac{a}{(1-\alpha)s_{ci}^r} > \frac{1}{s_{ci}^r}$. 
\\

These inequalities result in $\frac{\partial}{\partial v_{ci}} (C - C^r) > 0.$
\end{proof}

We conclude from~\ref{theorem:1} and~\ref{theorem:2} that any underestimation of lane group saturation flow rates will result in an over-estimation of the optimal cycle length. Moreover, as shown in proof~\ref{proof:2}, the overestimated cycle length values grow at a larger rate compared to an ideal estimation with the increase of the incoming traffic flows.

In the following section, the above claims are validated using microscopic simulation of multiple scenarios under the assumption of a fixed estimation of saturation flow rate~$(s=1900 \frac{v}{h})$.

\subsection{Comparisons}
Nine scenarios were designed for comparisons. The scenarios include three levels of traffic with respect to volume (low, medium, high) and three levels of traffic with respect to the imbalance of the incoming traffic (low, medium, high). Tables~\ref{tbl:scenarios1},~\ref{tbl:scenarios2}  and~\ref{tbl:scenarios3} depict the details of designed comparison scenarios for intersection~1, intersection~2 and intersection~3 respectively.

\begin{table*}[!htbp]
\centering
\caption{First intersection comparison scenarios}
\label{tbl:scenarios1}
\resizebox{\textwidth}{!}{%
\begin{tabular}{@{}cccccccc@{}}
\toprule
\textbf{Scenario \#} & \textbf{Total Traffic (veh/h)} & \textbf{Phase 1} & \textbf{Phase 2} & \textbf{Phase 3} & \textbf{Phase 4} & \textbf{Traffic Level} & \textbf{Traffic Imbalance} \\ \midrule
1 & \multirow{3}{*}{2000} & 500 & 500 & 500 & 500 & Low & Low \\ \cmidrule(r){1-1} \cmidrule(l){3-8} 
2 &  & 600 & 600 & 400 & 400 & Low & Medium \\ \cmidrule(r){1-1} \cmidrule(l){3-8} 
3 &  & 200 & 200 & 400 & 1200 & Low & High \\ \midrule
4 & \multirow{3}{*}{3000} & 750 & 750 & 750 & 750 & Medium & Low \\ \cmidrule(r){1-1} \cmidrule(l){3-8} 
5 &  & 900 & 900 & 600 & 600 & Medium & Medium \\ \cmidrule(r){1-1} \cmidrule(l){3-8} 
6 &  & 300 & 300 & 600 & 1800 & Medium & High \\ \midrule
7 & \multirow{3}{*}{4000} & 1000 & 1000 & 1000 & 1000 & High & Low \\ \cmidrule(r){1-1} \cmidrule(l){3-8} 
8 &  & 1200 & 1200 & 800 & 800 & High & Medium \\ \cmidrule(r){1-1} \cmidrule(l){3-8} 
9 &  & 400 & 400 & 800 & 2400 & High & High \\ \bottomrule
\end{tabular}%
}
\end{table*}

\begin{table*}[!htbp]
\centering
\caption{Second intersection comparison scenarios}
\label{tbl:scenarios2}
\resizebox{\textwidth}{!}{%
\begin{tabular}{@{}ccccccc@{}}
\toprule
\textbf{Scenario \#} & \textbf{Total Traffic (veh/h)} & \textbf{Phase 1} & \textbf{Phase 2} & \textbf{Phase 3} & \textbf{Traffic Level} & \textbf{Traffic Imbalance} \\ \midrule
1 & \multirow{3}{*}{4000} & 2400 & 1200 & 400 & Low & Low \\ \cmidrule(r){1-1} \cmidrule(l){3-7} 
2 &  & 2000 & 1600 & 400 & Low & Medium \\ \cmidrule(r){1-1} \cmidrule(l){3-7} 
3 &  & 1600 & 1600 & 800 & Low & High \\ \midrule
4 & \multirow{3}{*}{6000} & 3600 & 1800 & 600 & Medium & Low \\ \cmidrule(r){1-1} \cmidrule(l){3-7} 
5 &  & 3000 & 2400 & 600 & Medium & Medium \\ \cmidrule(r){1-1} \cmidrule(l){3-7} 
6 &  & 2400 & 2400 & 1200 & Medium & High \\ \midrule
7 & \multirow{3}{*}{7600} & 4560 & 2280 & 760 & High & Low \\ \cmidrule(r){1-1} \cmidrule(l){3-7} 
8 &  & 3800 & 3040 & 760 & High & Medium \\ \cmidrule(r){1-1} \cmidrule(l){3-7} 
9 &  & 3040 & 3040 & 1520 & High & High \\ \bottomrule
\end{tabular}%
}
\end{table*}

\begin{figure}[H]
    \centering
    \includegraphics[width=4in]{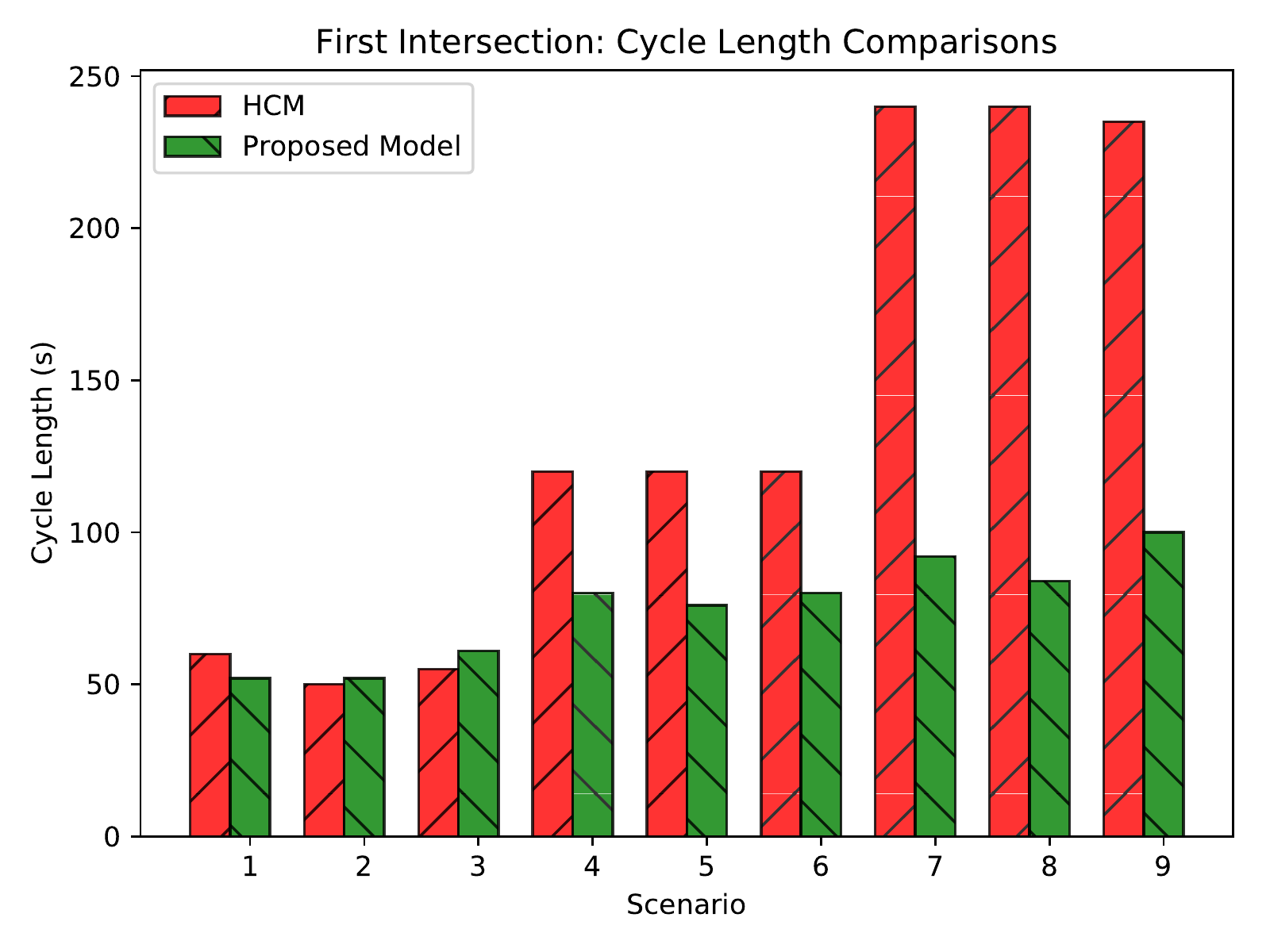}
    \caption{First Intersection: Estimated Cycle Lengths}
    \label{fig:cycles1}
\end{figure}

\begin{figure}[H]
    \centering
    \includegraphics[width=4in]{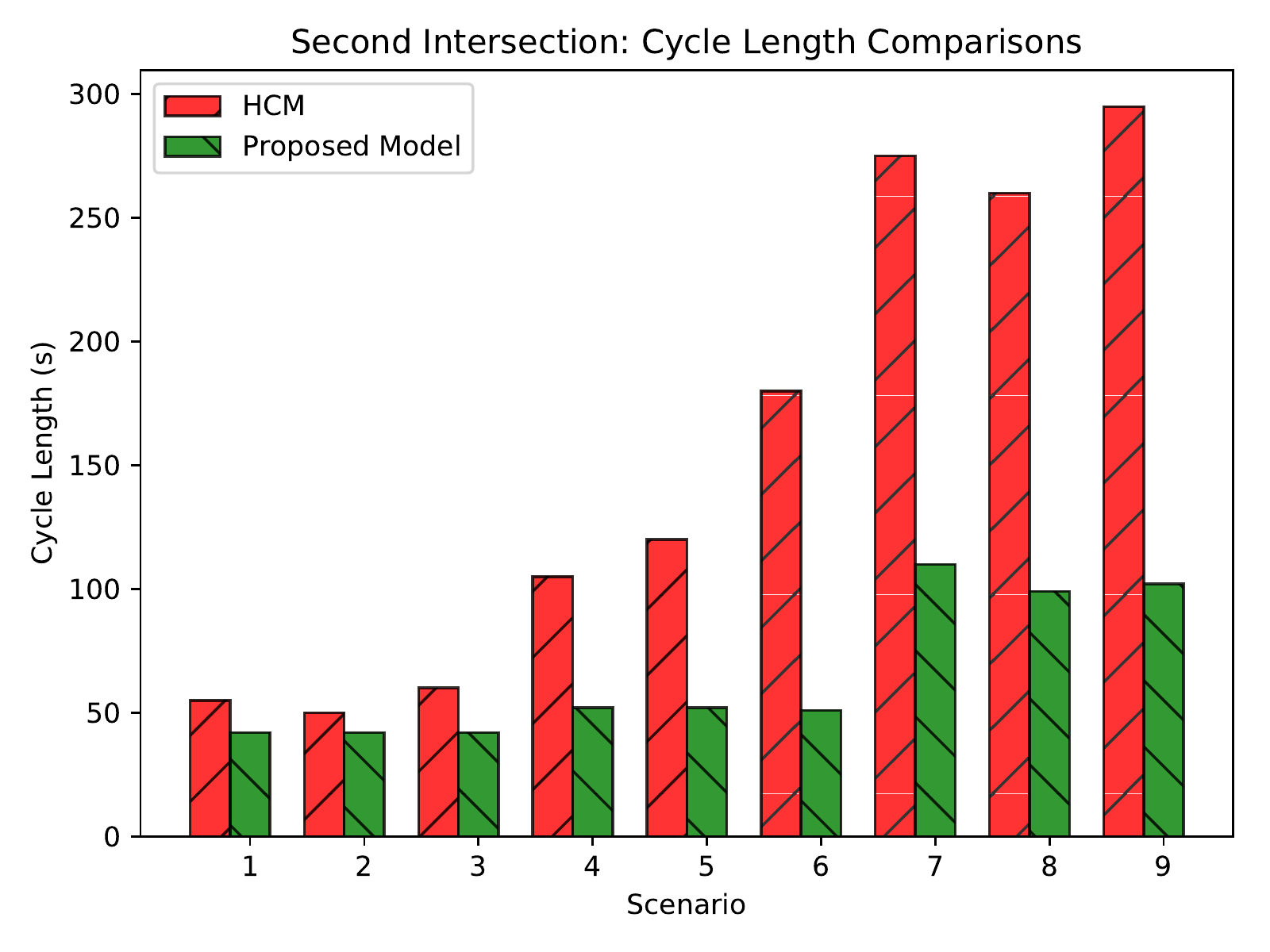}
    \caption{Second Intersection: Estimated Cycle Lengths}
    \label{fig:cycles2}
\end{figure}

\begin{figure}[H]
\centering
    \includegraphics[width=4in]{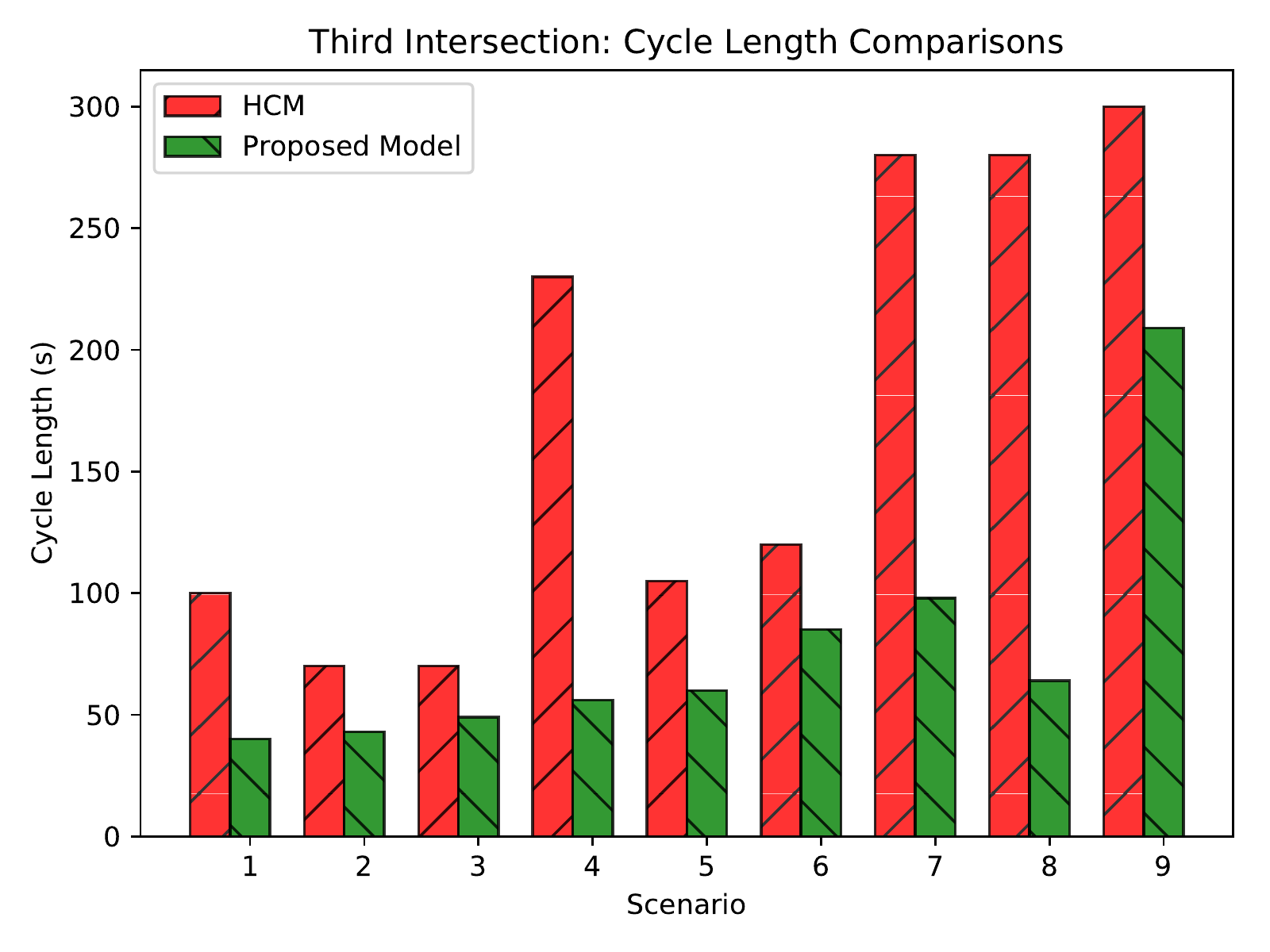}
    \caption{Third Intersection: Estimated Cycle Lengths}
    \label{fig:cycles3}
\end{figure}

\begin{table*}[!htbp]
\centering
\caption{Third intersection comparison scenarios}
\label{tbl:scenarios3}
\resizebox{\textwidth}{!}{%
\begin{tabular}{@{}cccccccc@{}}
\toprule
\textbf{Scenario \#} & \textbf{Total Traffic (veh/h)} & \textbf{Phase 1} & \textbf{Phase 2} & \textbf{Phase 3} & \textbf{Phase 4} & \textbf{Traffic Level} & \textbf{Traffic Imbalance} \\ \midrule
1 & \multirow{3}{*}{5000} & 1250 & 1250 & 1250 & 1250 & Low & Low \\ \cmidrule(r){1-1} \cmidrule(l){3-8} 
2 &  & 1000 & 1500 & 1000 & 1500 & Low & Medium \\ \cmidrule(r){1-1} \cmidrule(l){3-8} 
3 &  & 500 & 1000 & 500 & 3000 & Low & High \\ \midrule
4 & \multirow{3}{*}{7500} & 1875 & 1875 & 1875 & 1875 & Medium & Low \\ \cmidrule(r){1-1} \cmidrule(l){3-8} 
5 &  & 1500 & 2250 & 1500 & 2250 & Medium & Medium \\ \cmidrule(r){1-1} \cmidrule(l){3-8} 
6 &  & 750 & 1500 & 750 & 4500 & Medium & High \\ \midrule
7 & \multirow{3}{*}{12000} & 3000 & 3000 & 3000 & 3000 & High & Low \\ \cmidrule(r){1-1} \cmidrule(l){3-8} 
8 &  & 2400 & 3600 & 2400 & 3600 & High & Medium \\ \cmidrule(r){1-1} \cmidrule(l){3-8} 
9 &  & 1200 & 2400 & 1200 & 7200 & High & High \\ \bottomrule
\end{tabular}%
}
\end{table*}
Figs.~\ref{fig:cycles1},~\ref{fig:cycles2} and~\ref{fig:cycles3} demonstrate the estimated cycle lengths from the proposed method and the HCM model for the three intersections.

\begin{figure}[H]
    \centering
    \caption{Phase Split Comparisons for Intersection 1}
            \vspace{-30pt}

        \subfloat[][Low Traffic Volume]{
        \includegraphics[width=0.75\textwidth]{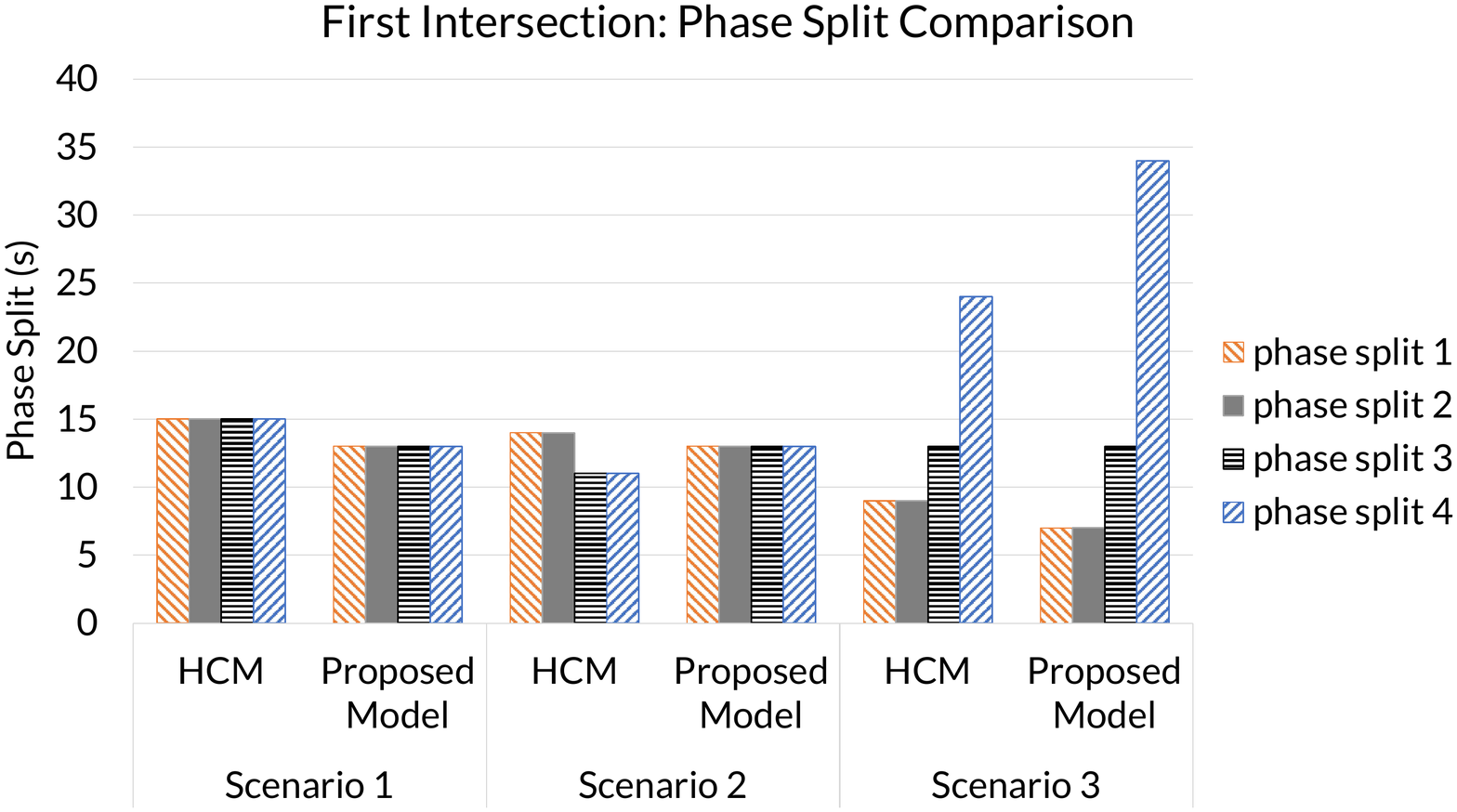}}
        \vspace{-30pt}
        \subfloat[][Medium Traffic Volume]{

        \includegraphics[width=0.75\textwidth]{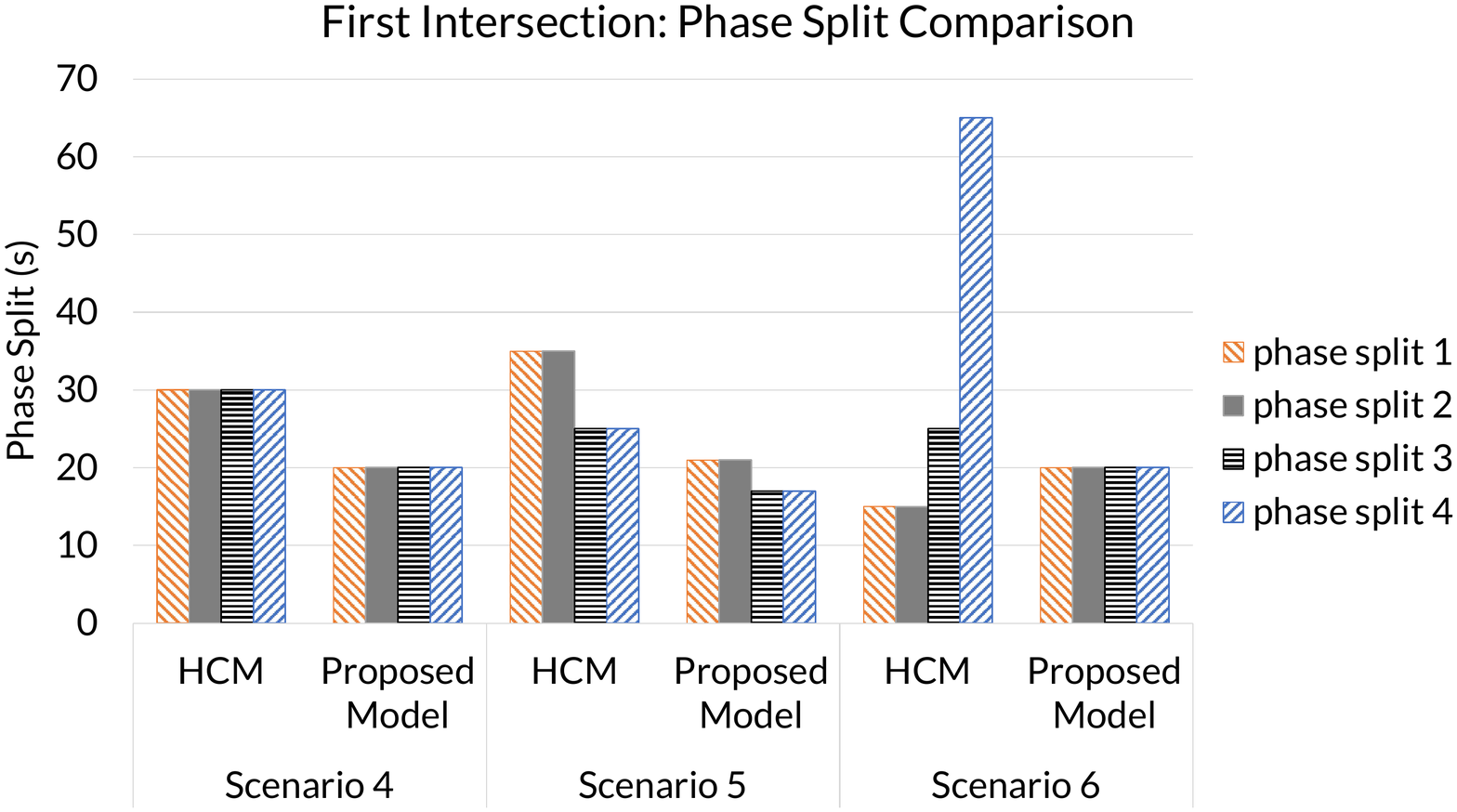}}
        \vspace{-30pt}
        \subfloat[][High Traffic Volume]{
        \includegraphics[width=0.75\textwidth]{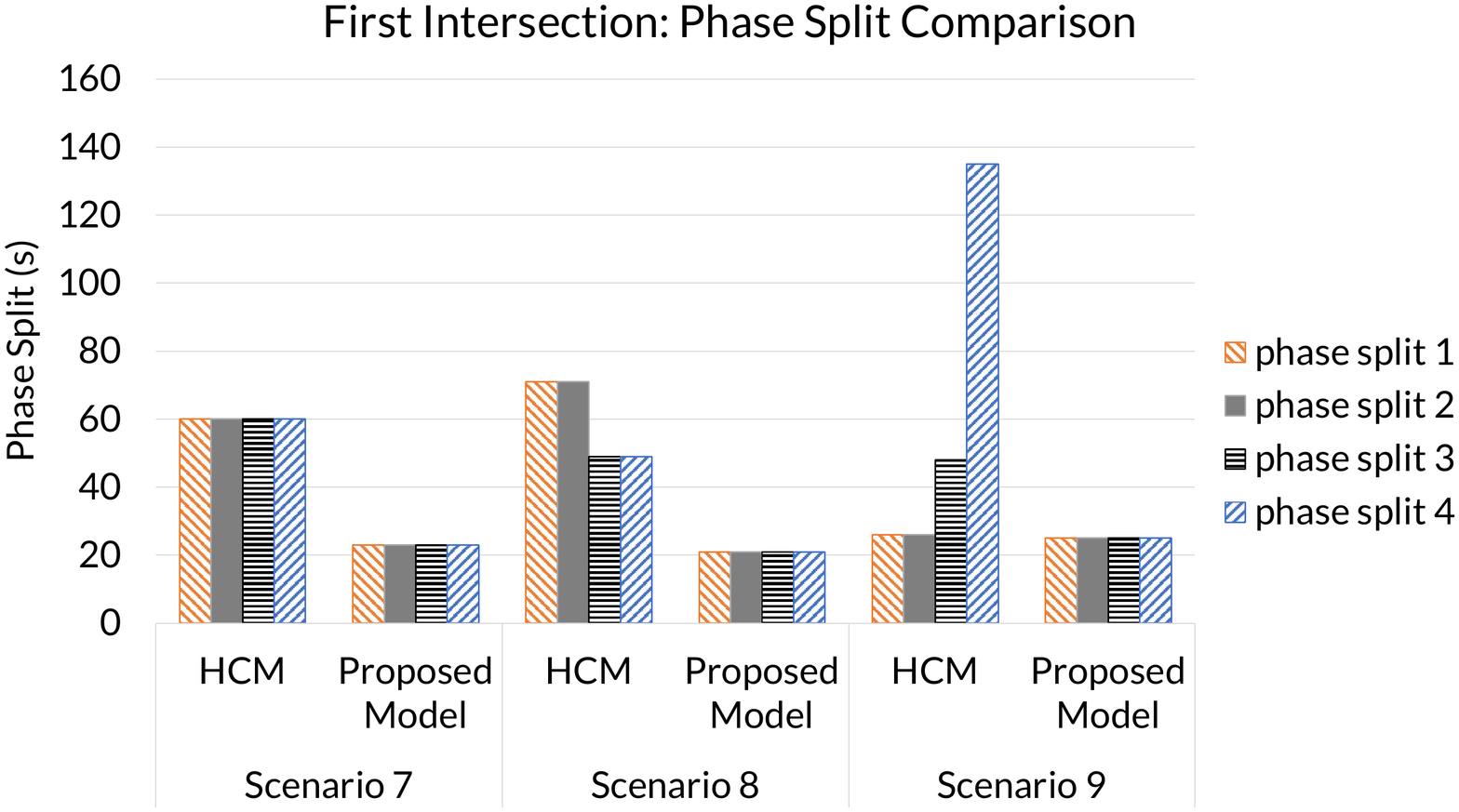}
        }
    \label{fig:int1_phase}
\end{figure}

\begin{figure}[H]
    \centering
    \caption{Phase Split Comparisons of Intersection 2}
            \vspace{-30pt}

        \subfloat[Low Traffic Volume]{\includegraphics[width=0.75\textwidth]{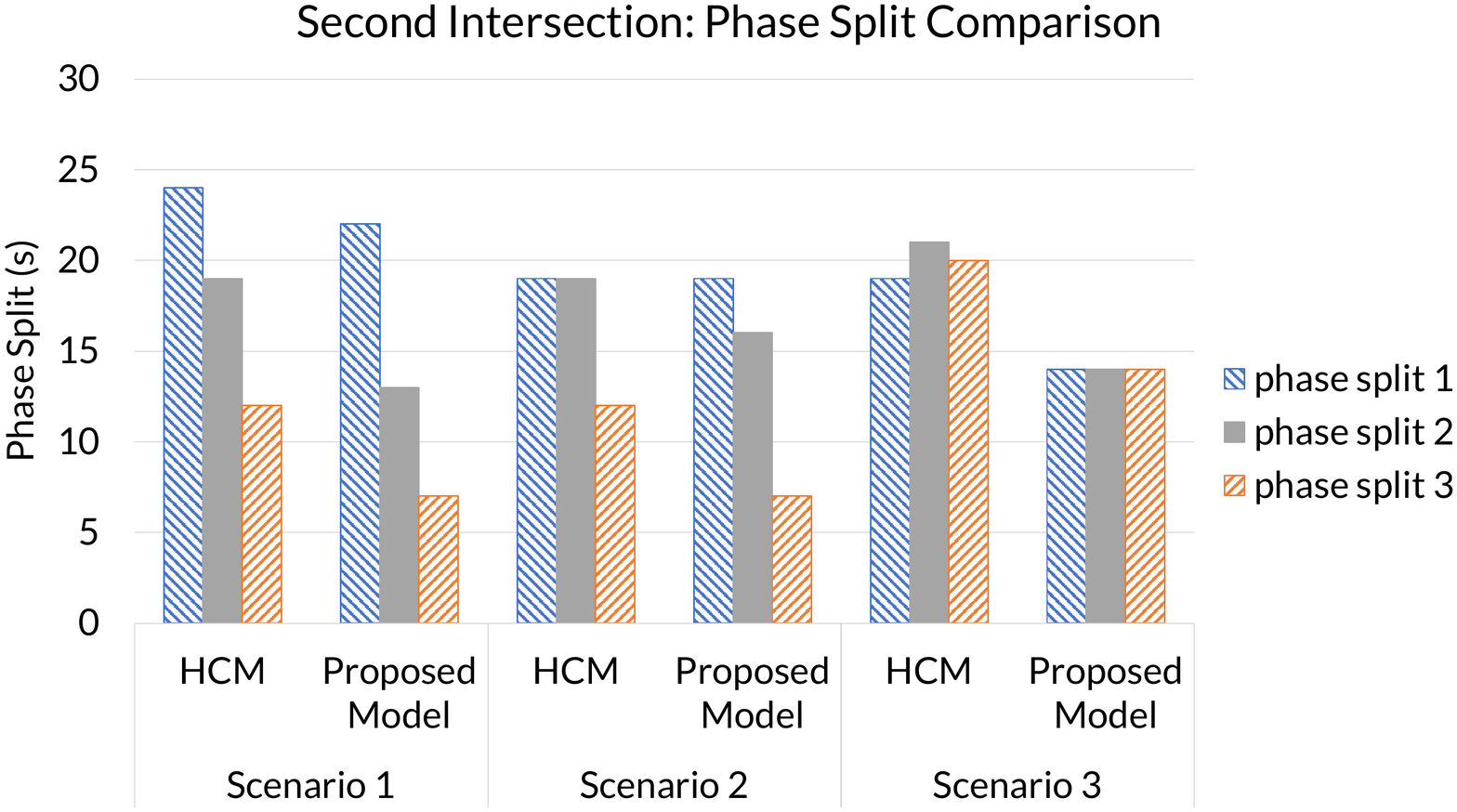}}
                \vspace{-30pt}

        \subfloat[Medium Traffic Volume]{\includegraphics[width=0.75\textwidth]{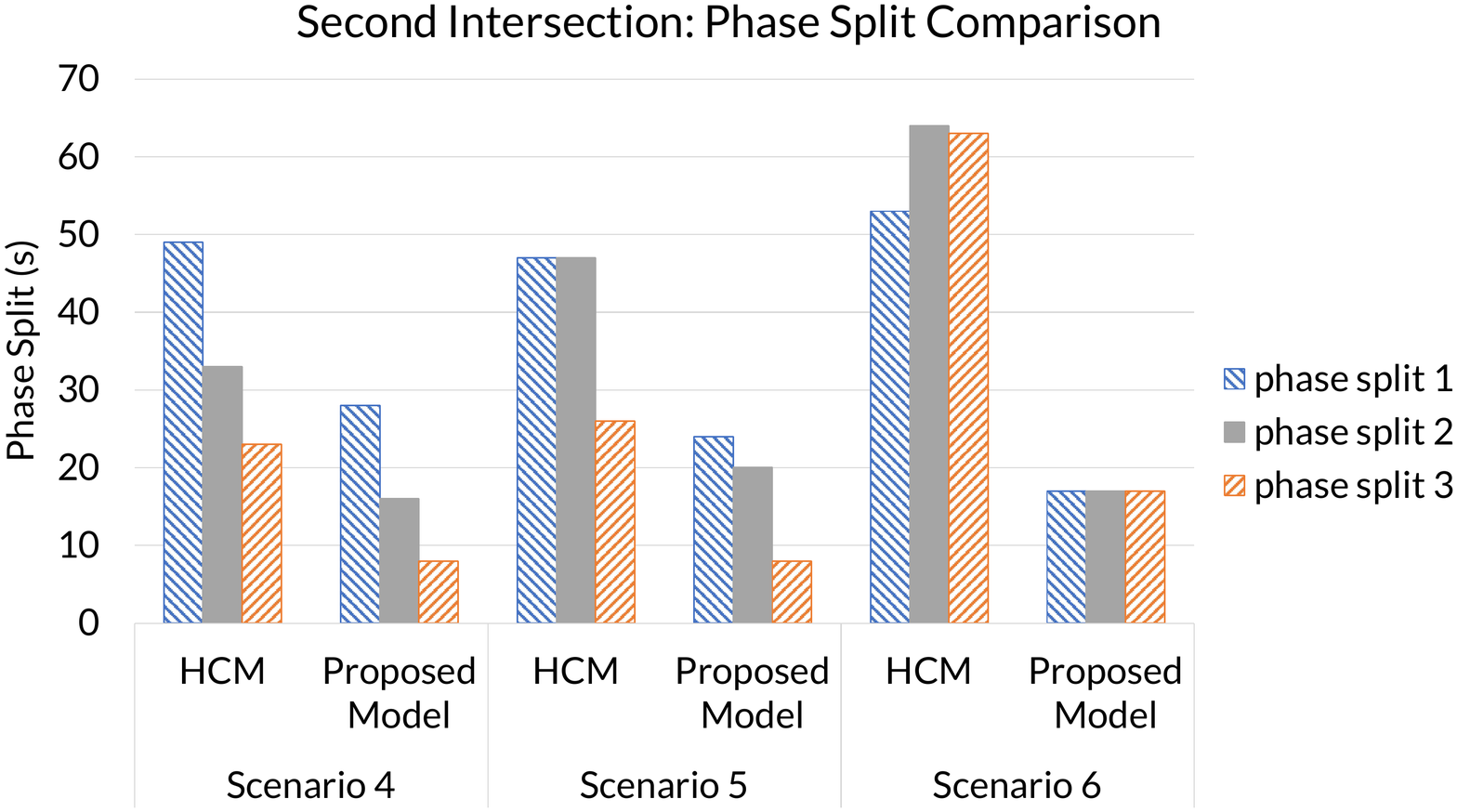}}
                \vspace{-30pt}

        \subfloat[High Traffic Volume]{\includegraphics[width=0.75\textwidth]{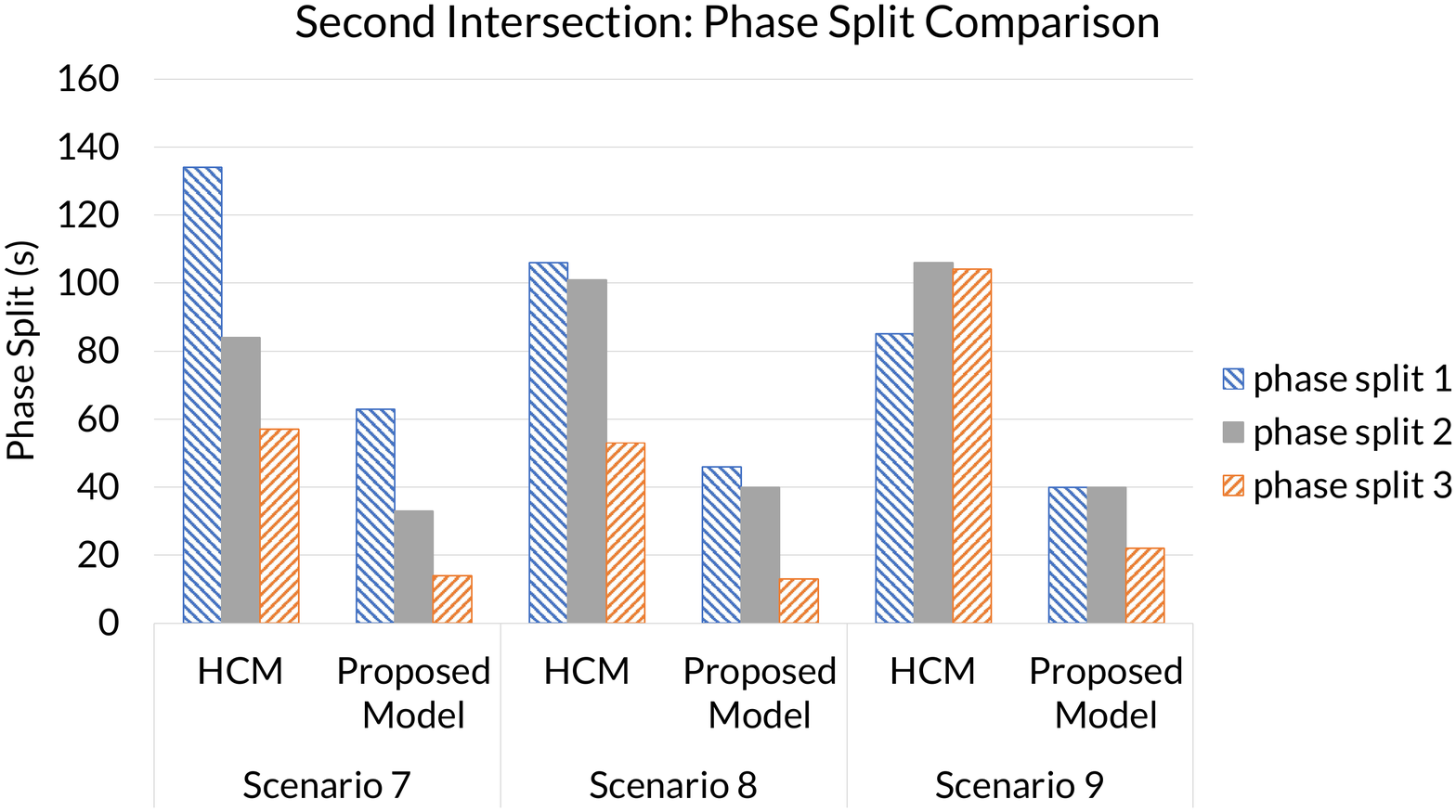}}
    \label{fig:int2_phase}
\end{figure}

\begin{figure}[H]
    \centering
    \caption{Phase Split Comparisons for Intersection 3}
    \vspace{-30pt}
        \subfloat[Low Traffic Volume]{\includegraphics[width=0.75\textwidth]{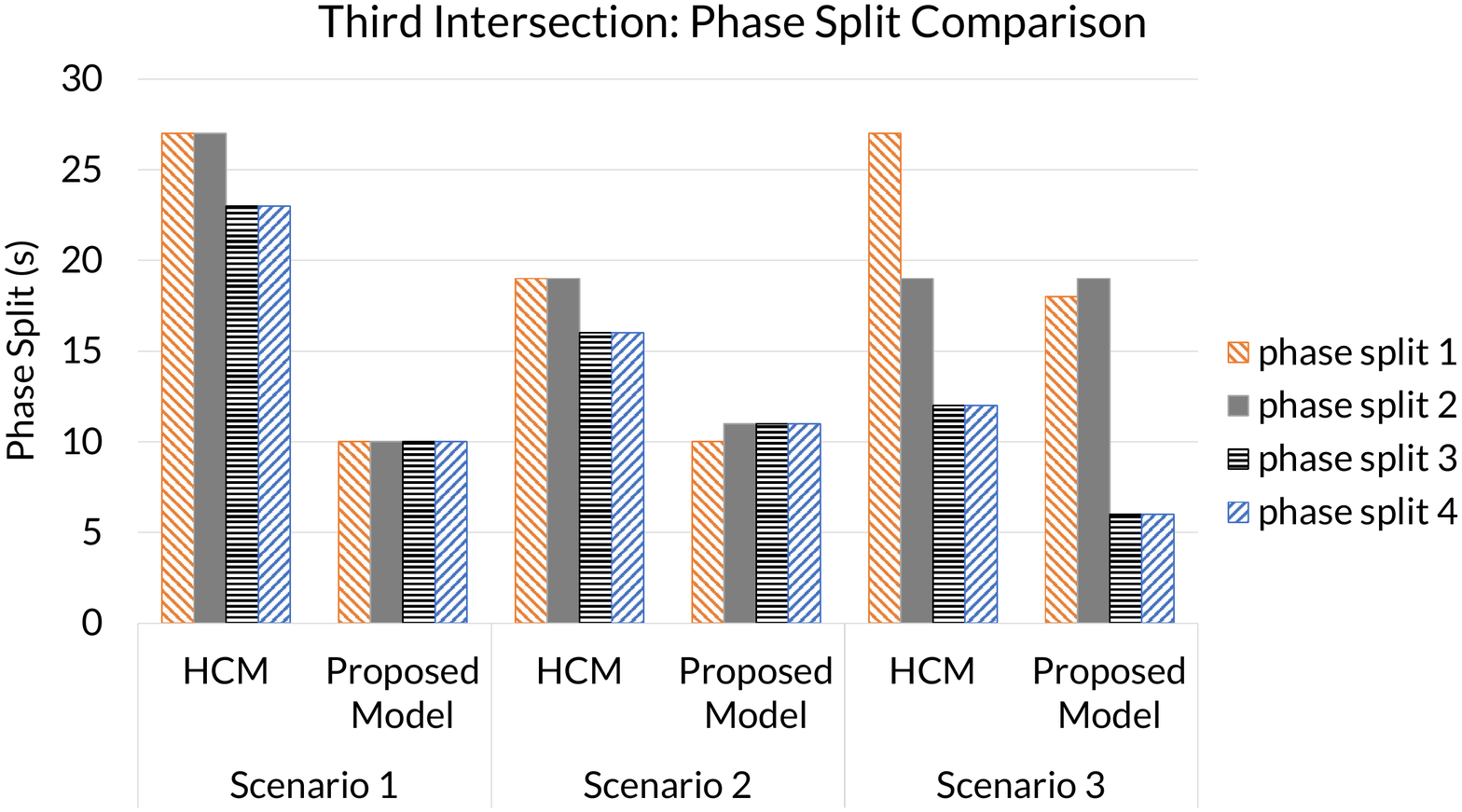}}
        \vspace{-30pt}
        \subfloat[Medium Traffic Volume]{\includegraphics[width=0.75\textwidth]{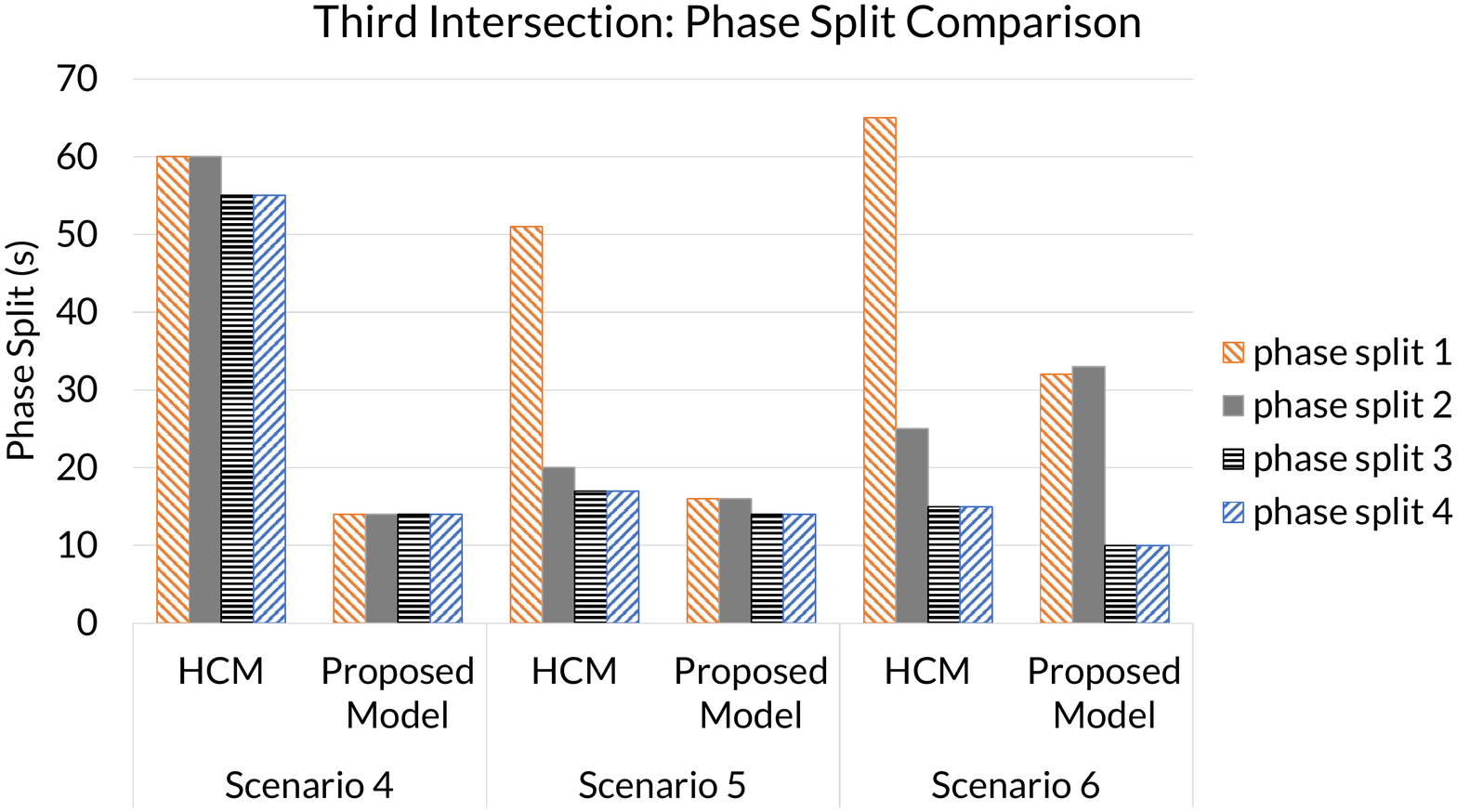}}
        \vspace{-30pt}
        \subfloat[High Traffic Volume]{\includegraphics[width=0.75\textwidth]{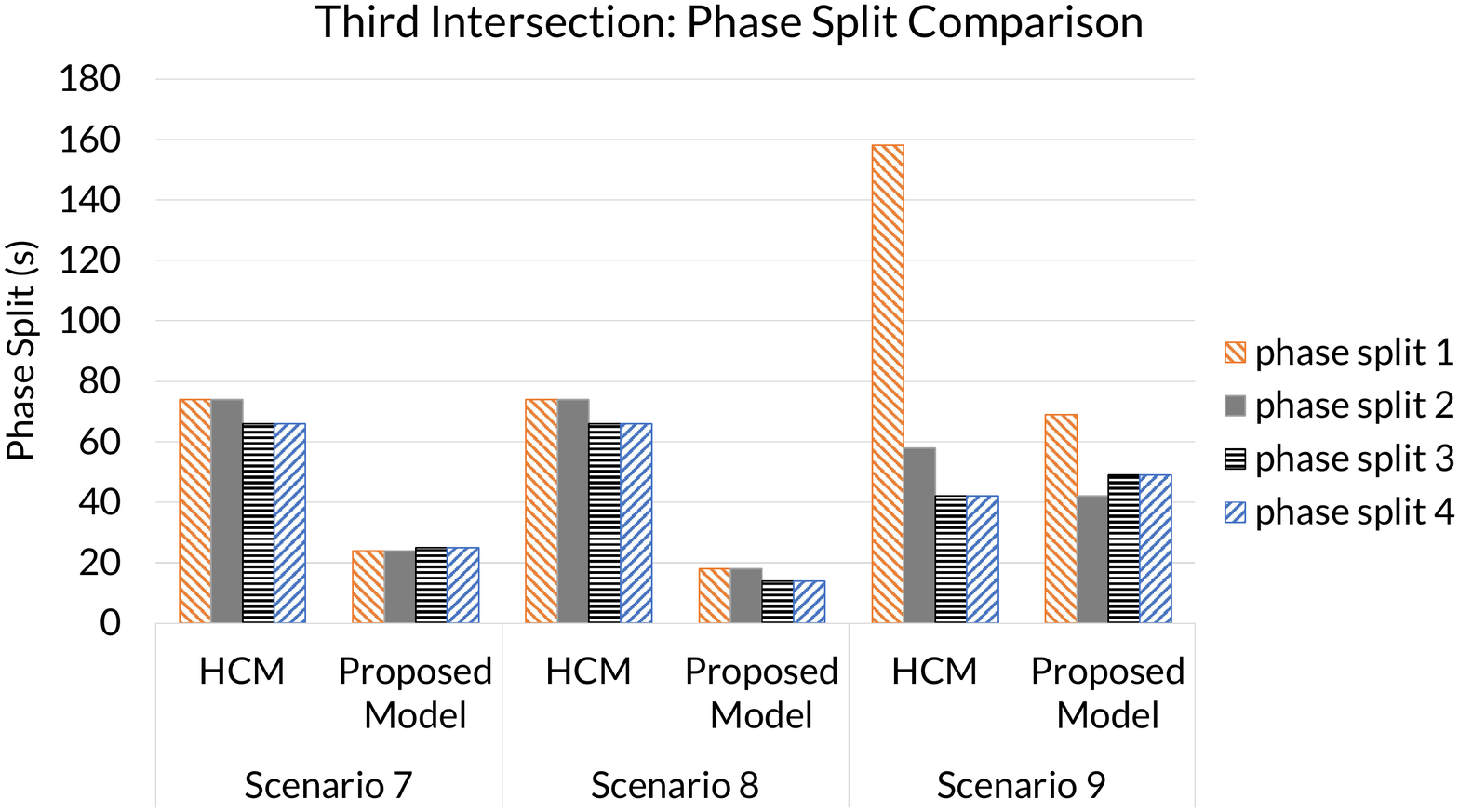}}
    \label{fig:int3_phase}
\end{figure}

Figs.~\ref{fig:int1_phase},~\ref{fig:int2_phase} and~\ref{fig:int3_phase} compare the two models' suggested phase splits for the three intersection models.

\begin{figure}[H]
    \centering
    \includegraphics[width=4in]{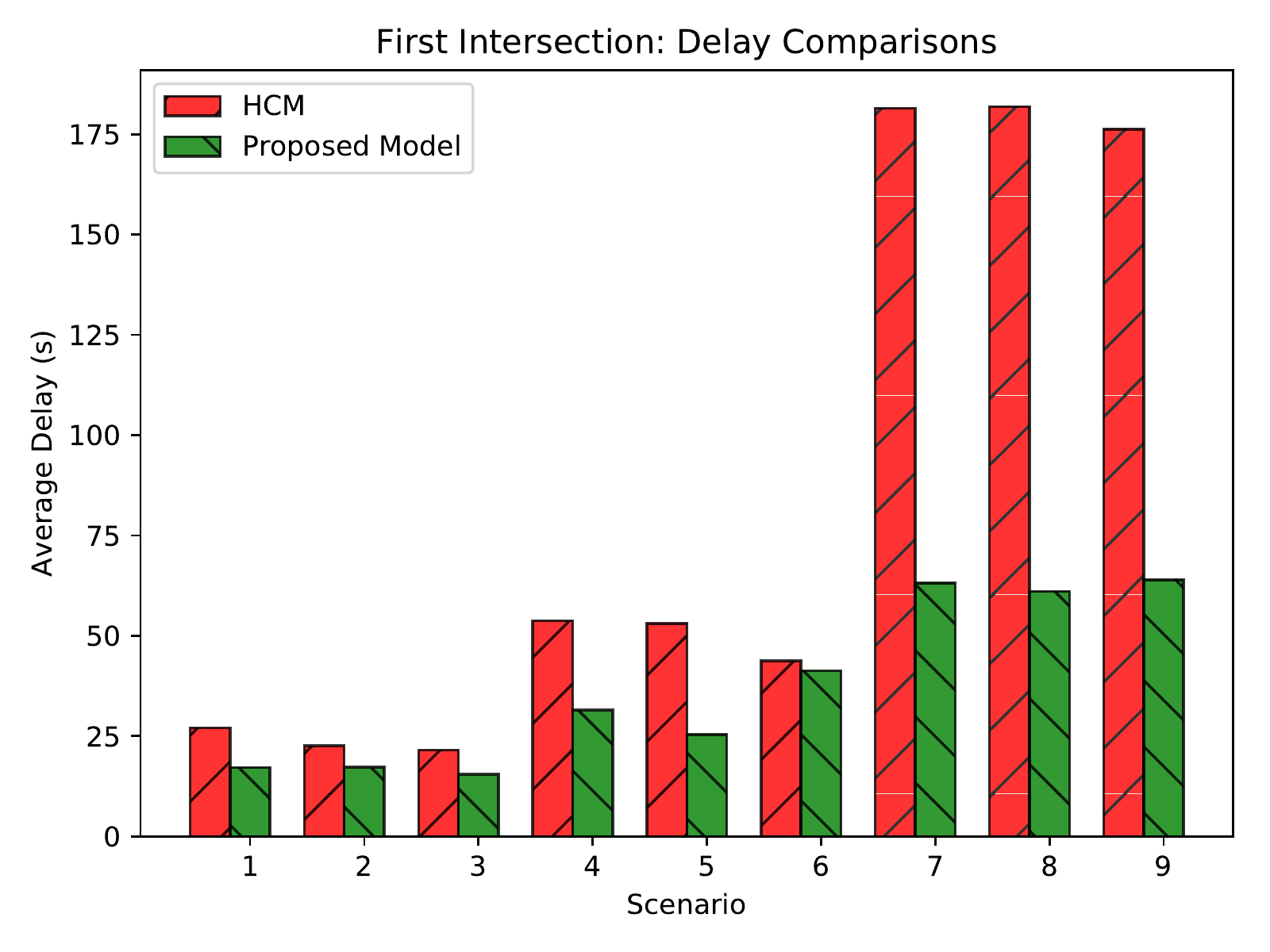}
    \caption{First Intersection: Average Vehicular Delays}
    \label{fig:delays1}
\end{figure}

\begin{figure}[H]
    \centering
    \includegraphics[width=4in]{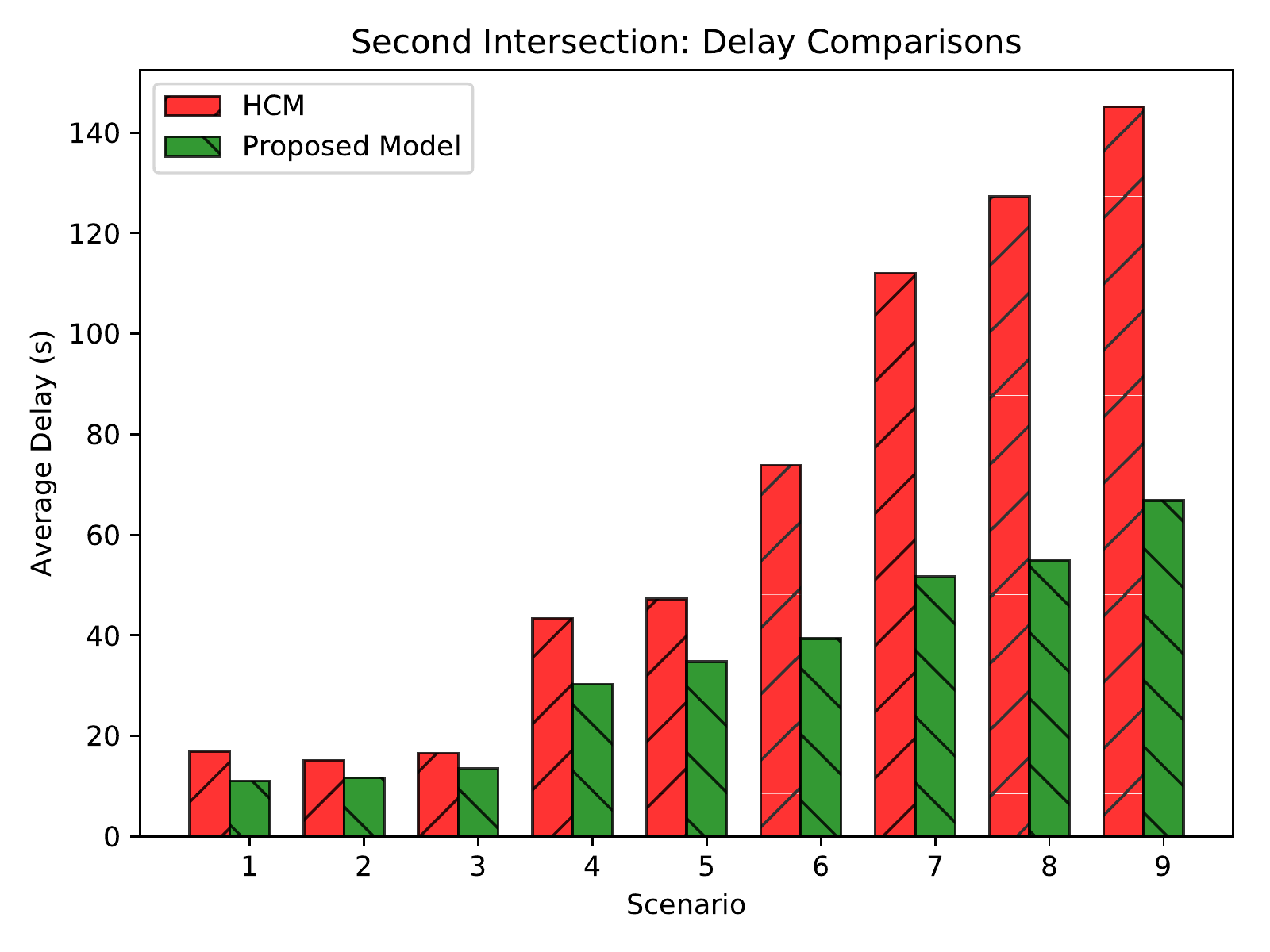}
    \caption{Second Intersection: Average Vehicular Delays}
    \label{fig:delays2}
\end{figure}
\begin{figure}[H]
\centering
    \includegraphics[width=4in]{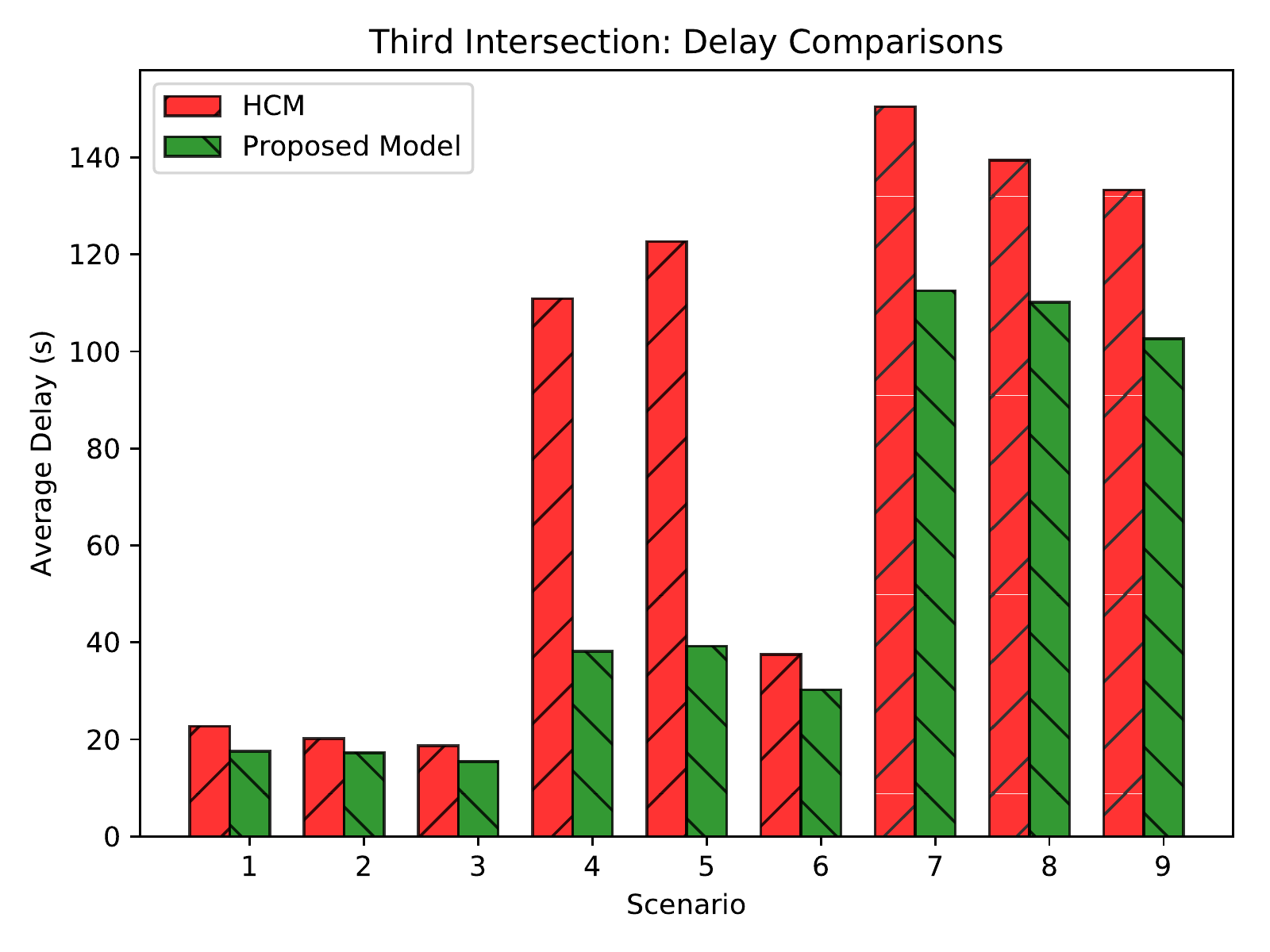}
    \caption{Third Intersection: Average Vehicular Delays}
    \label{fig:delays3}
\end{figure}

The analytical models are known for their over-estimations of the optimal cycle length which can be verified in the above figures. On average, the proposed model's cycle lengths were~$45\%$,~$58\%$ and~$48\%$ shorter than HCM's suggested cycle lengths for intersections~$1$,~$2$ and~$3$ respectively.\\

It should be noted that in practice, the cycle length determined from the HCM equation would be checked against reasonable minimum and maximum values. These values are set by the local jurisdiction. However, to assure fair comparisons we skip this process for both compared models.\\

The disparity in optimal cycle length estimations are even more significant in over-saturated traffic conditions as well as certain situations unique to each intersection. For example, HCM's estimations of the optimal cycle length blows up to large numbers when the estimated traffic volume of the left turn movements are relatively large, e.g. scenarios~$\#4$ and~$\#7$. This is caused by HCM's overestimation of volume to capacity ratio for the left turn movements.

Figs.~\ref{fig:delays1},~~\ref{fig:delays2} and~~\ref{fig:delays3} compare the calculated average vehicular delays of the proposed model and the HCM model. Delays are computed through microscopic simulations with identical settings. HCM's own estimations of delay have been omitted due to their over-estimations caused by the inclusion of incremental delay. Vissim's delay measurements are used for comparisons which are more precisely calculated through microscopic simulations. See appendix~\ref{first_appendix} for detailed comparisons of HCM's delay estimation and delays computed through microscopic simulation.

The above figures show that the proposed model outperforms the HCM model in all scenarios for all three intersection models. The proposed model's performance is even more significant in mid to high traffic volumes as the HCM's estimation of the optimal cycle length starts to blow up, causing traffic jams in certain lane groups, hence longer delays in such traffic situations.

The simulation results showed that the proposed model decreased average vehicular delay by~$55\%$,~$47\%$ and~$36\%$ compared to the HCM model for intersections~1,~2 and~3 respectively. Furthermore, the results suggest that the proposed model outperforms the HCM model similarly in all traffic conditions.

\section{Conclusions}\label{sec:conclusions_4}
This chapter introduced a novel data-driven approach to the optimal signal timing problem. The proposed method relies on recorded data from various traffic scenarios through microscopic simulations. Multiple models can be built with different objectives. The proposed model can be customized to each network's unique geometrical and traffic conditions, outperforms existing state-of-the-practice pre-timed signal control model and provides new insights into the the traffic control problem.
\subsection{Applications in Traffic Engineering}
Although the proposed method was introduced as a signal timing model in this chapter, it can be utilized in a number of applications.
\begin{itemize}
    \item \textbf{Signal Timing Model} The proposed model can be utilized as a replacement to existing state-of-the-practice pre-timed signal timing methods such as the HCM model.
    \item \textbf{Signal Timing Helper} A trained model can be utilized as a helper method to any signal timing planner. For example, the model can provide customized cycle length and green time ranges to the signal timing planner.
    \item \textbf{Adaptive Signalization Method} With proper modifications, the proposed model can be applied in the context of real-time traffic responsive signal control. These modifications include readjusting the training scenarios as well as the structure of the learning model. A traffic state estimator would be required to provide the input to the model. We are currently working on designing a real time traffic state estimator that leverages CAV information received through BSM messages to estimate the routing decisions of the incoming traffic. The traffic state estimator module coupled with the proposed traffic signal model, form a real-time traffic responsive controller.
\end{itemize}

\chapter{Traffic State Estimation}\label{chpt:tse}
Traffic state estimation (TSE) is the process of the inference of traffic state variables such as flow, density and the speed of the traffic on road segments using partially observed traffic data. Due to technical limitations, traffic measurements are sparse and sometimes non-existing. Therefore, the main challenge of TSE is to deal with the shortage of measurements and the associated noise. Numerous studies have proposed TSE methods relying on different approaches to the problem~\cite{seo2017traffic}.

Since the publication of Wang's seminal paper~\cite{wang2005real} on freeway traffic state estimation in 2005, there have been considerable advances in the field. Wang developed a general approach based on the extended Kalman filter~\cite{jazwinski1969adaptive} to the real-time estimation of the complete traffic state in freeway stretches. The paper presented a general stochastic macroscopic traffic flow model of freeway stretches. The authors organized a macroscopic traffic flow model along with a measurement model in a compact state-space form, which made it possible to apply estimation methods such as the Kalman filter. The authors designed a traffic state estimator based on the model in the state-space form using the extended Kalman filtering method. One of the main contributions of this paper was applying the extended version of the Kalman filter method designed to be applicable to nonlinear systems. The authors also studied various factors such as the sensitivity of the estimator to the initial values of the estimated model parameters and dynamic tracking of time-varying model parameters.

\begin{figure}
    \centering
    \includegraphics[width=4in]{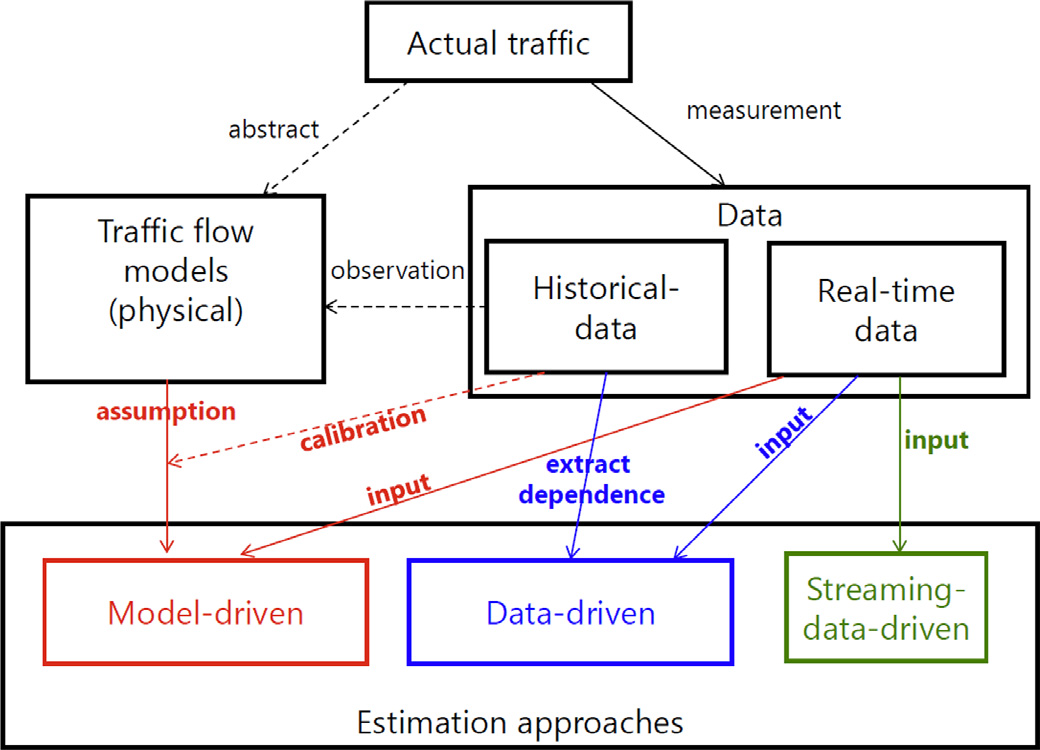}
    \caption{Traffic State Estimation with Assumptions and Input~\cite{seo2017traffic}}
    \label{fig:tse}
\end{figure}

Typically, TSE methods can be characterized according to the three choices of \emph{estimation approach}, \emph{traffic flow model}, and \emph{input data}. Figure~\ref{fig:tse} demonstrates a typical architecture of Traffic State Estimators including their assumptions and input~\cite{seo2017traffic}.

The rest of this chapter is organized as follows. Section~\ref{tse:highway} provides a brief overview of the previous work on TSE in highways. Section~\ref{tse_proposed} introduces the proposed streaming data-driven approach to traffic state estimation; Finally, Section~\ref{sec:conclusion_5} provides conclusions.

\section{Traffic State Estimation on Highways}\label{tse:highway}
The literature on traffic state estimation can be divided into two classes of model-driven and data-driven methods. Model-driven methods rely on well-studied traffic flow models and produce state estimators for such models. Data-driven methods, on the other hand, generally rely on historical data and machine learning methods to learn traffic flow trends. Data-driven methods do not rely on specific traffic flow models, instead utilize the power of machine learning techniques to deal with non-linearity which are among the main disadvantages of model-driven approach. However, these methods cannot react to irregular traffic patterns and unexpected events.
\subsection{Model-Driven TSE on Highways}\label{tes:mdl}
Data assimilation techniques are widely used in the literature on model-driven TSE and among those, Kalman Filter Techniques are the most popular. The Kalman filter is a recursive algorithm that works in a two-step process: \emph{State Prediction} and \emph{Measurement Update}. In the prediction step, the filter produces estimates of the current state variables, along with their uncertainties. Once the outcome of the next measurement (corrupted with noise) is observed, these estimates are updated using a weighted sum of the measurements, with more weight being given to estimates with higher certainty. Kalman filters in general are based on the state-space model (Equations~\ref{eq:syseq} and \ref{eq:measeq}).
\begin{equation}
    x_t = f_t(x_{t-1},v_t)
    \label{eq:syseq}
\end{equation}
\begin{equation}
    y_t = h_t(x_t,\omega_t)
    \label{eq:measeq}
\end{equation}
Where:
\begin{addmargin}[3em]{3em}
\begin{description}
    \item $x_t$ is a state vector
    \item $f_t$ is a system model
    \item $v_t$ is system noise
    \item $y_t$ is an observation vector
    \item $h_t$ is an observation model
    \item and, $\omega_t$ is observation noise at time t
\end{description}
\end{addmargin}

Equation~\ref{eq:syseq} is known as the system (process) model and Equation~\ref{eq:measeq} is known as the measurement (observation) model. A Kalman filter estimates the most probable state variables with respect to a system model, available observation and noise. The main objective of the Kalman filter is to obtain the $x_t$ that maximizes $p(x_t | y_1, y_2,...,  y_t)$.
\paragraph*{Using Kalman filter methods in the context of TSE}
Kalman Filter and its variants are the most widely used state estimation method in TSE literature. Due to the fact that the traffic flow models are nonlinear, the standard KF is rarely used; However, Extended Kalman filter (EKF), Unscented Kalman filter (UKF)~\cite{wan2000unscented} and Ensemble Kalman filter (EnKF)~\cite{evensen2003ensemble} have been applied to the problem. A common mistake in the literature is to use EKF technique on non-differentaible models such as the Godunov discretization of the LWR model, including Cell Transition Model (CTM)~\cite{blandin2012sequential}.

\subsection{Data-Driven Traffic State Estimation}
Data-driven TSE methods can be further divided into~\emph{historical data-driven} and~\emph{streaming data-driven} approaches. Both historical and streaming date-driven approaches rely on traffic information obtained from a wide range of sensor types either installed on the road or on vehicles. Section~\ref{tse:intro-sensors} provides a brief overview of sensor technologies used for traffic state estimation.

\subsubsection{Traffic Information Collection Techniques}\label{tse:intro-sensors}
Fixed sensors have become an integral part of many urban traffic control systems in the 21st century and have helped collect valuable traffic information to support the development of the first-generation intelligent transportation systems~\cite{kurzh2015}. However, fixed sensors can only provide the information of traffic flow measured at discrete spatial points~\cite{SUN2013}, and we need to build special models to estimate the traffic states at other spatial locations. Because not any traffic flow model or estimation method is perfect, we always expect to fuse additional traffic information to increase estimation accuracy~\cite{berkow2009}. 

Most early attempts in this direction studied probe vehicle based traffic monitoring systems using wireless location technology~\cite{smith2004,ou2011}. These approaches sampled a portion of vehicles as they traversed the network. Wireless location technology and Global Position System (GPS) were used to record the latitude and longitude of the probe vehicles sampled along with their trips. Then, specific measurements of traffic flow (i.e., queue length and travel time) were estimated based on the sampled data~\cite{ban2011,cheng2012}. 

Compared with fixed sensors, these approaches were able to provide more information of the traffic flow, which could improve our knowledge on traffic states and detect errors of fixed sensors~\cite{li2014diagnosing}. Conventional probe vehicle based data have limited scope and time. Since the last decade, the connected vehicles~(CVs) based techniques have achieved significant advances, making it possible to access a great deal of more accurate, and multi-dimension information of traffic flow in real time~\cite{massaro2016car}. The newly available high-resolution trajectory data of individual vehicles could increase our understanding of traffic flow states, which are critical to traffic control. The V2I based data collection techniques have not been widely implemented. Some other technologies, especially the mobile sensing technique~\cite{hoh2008virtual,sun2013privacy}, could also provide high precision trajectory data, which is similar to the location data that CVs could provide. 

There are two major differences between the two approaches that are directly influenced by the differences between the probe vehicle data and the Connected Vehicle data. First, the resolution levels of trajectories and the penetration-rates of sampling vehicles for probe vehicle approaches are much lower than those for CVs based approaches. As a result, we usually need to divide the roads into grids and use the data collected by probe vehicles to estimate the average traffic flow states (e.g. average speed, average density) within each grid~\cite{he2017mapping,jiang2017traffic,ran2016using}. In contrast, with the high-resolution data collected by CVs, we can characterize the trajectories of each sampled vehicles and may also derive the trajectories of other neighboring vehicles that are not sampled~\cite{xie2018generic}. 

Second, constrained by the resolution level and the amount of data, most probe vehicle approaches are not suitable for real time traffic control. Instead, such data were usually used to estimate freeway traffic flow states or vehicle travel times~\cite{montanino2015}. In contrast, real time data collected by CVs and mobile sensing can be used to real-time traffic control.
\subsubsection{Historical Data-Driven Traffic State Estimation}
A historical data-driven TSE method is defined as a TSE method that directly relies on historical-data rather than explicit traffic flow models. It mainly uses statistical and ML approaches, and infers real-time traffic states based on dependence found in historical-data. 

Compared with the model-driven approach, the data-driven approach does not require explicit theoretical assumptions. Additionally, data-driven approaches often rely on relatively simple models; therefore, their computational cost for estimation can be remarkably low. However, a dependency on historical-data means that the methods can fail if irregular events or a long-term trend occurred. In this case, the computational cost for training and learning can be significantly high. Moreover, the method can be considered as a "black box", which means that it is difficult to obtain deductive insights.

Imputation methods have developed to complement missing data caused by malfunctions of detectors and communication. Statistical approaches were often used in early studies. These approaches relied on data from neighboring detectors and time periods.~\cite{smith2003exploring} compares several heuristic methods based on historical-data with a statistical method using a data augmentation technique. A linear regression model using data from spatial neighbor detectors~\cite{chen2003detecting}, and autoregressive integrated moving average~(ARIMA) using time series dataset~\cite{zhong2004estimation} are developed. 

To consider greater complexity of traffic data, several studies have developed methodologies that depend more substantially on historical data, such as ML and Bayesian statistics.~\cite{ni2005markov} proposed a time series based model incorporating to Bayesian network to improve bias and robustness of estimation. Kernel regression (KR)~\cite{yin2012imputing}, fuzzy c-means (FCM)~\cite{tang2015hybrid}, k-nearest neighbors~(kNN)~\cite{tak2016data}, and probabilistic principal component analysis~(PPCA)~\cite{li2013efficient} have been used to incorporate more spatial-temporal information.~\cite{tan2014robust} developed a method based on robust principal component analysis~(RPCA) to consider capacity of traffic flow as well as the temporal correlation. 

Tensor-based methods using Tucker decomposition (TD)~\cite{tan2013tensor} have also been used to consider spatial-temporal information.~\cite{ran2016tensor} proposed tensor completion algorithm~(HaLRTC) to improve imputing performance. Deep learning technique has also been applied in~\cite{duan2016efficient}. Bayesian state-space modeling has also been applied to estimate the hidden parameters of the traffic state.~\cite{polson2017bayesian} used the Bayesian particle filter to estimate traffic state corresponding to free-flowing, breakdown, and recovery regimes.

Several studies have attempted to estimate flow using probe vehicle data. These studies have explored statistical relations between flow and other variables. The parameters of TSE models~(i.e., concepts similar to Flow diagram) are estimated from historical stationary data, and then streaming mobile data are used for TSE. For example, speed~\cite{anuar2015estimating,neumann2013dynamic}, variance of speed~\cite{blandin2012individual,bulteau2013traffic}, and other variables in extended floating car data known as~\emph{xFCD} (e.g. range sensors or cameras on car)~\cite{wilby2014lightweight} are used for such estimation. 

It is known that machine learning can be good at predicting nonlinear phenomena often found in the transportation field~\cite{karlaftis2011statistical}. However, relatively few studies have used machine learning for TSE. This might be because the physics of link traffic flow has been well understood in previous studies, and the available models' performance is relatively high for estimation and subsequent control compared with data-driven methods.

\subsubsection{Streaming-Data Driven Approaches}
A streaming-data-driven TSE method is defined as a TSE method that relies on streaming data and weak assumptions, such as random sampling condition and conservation law. It does not rely on strong assumptions characterized by empirical relation, such as partial differential equation models and flow diagrams, nor dependency in historical-data (as in historical data-driven approaches). This means that the method requires less a priori knowledge and no historical data, making it robust against uncertain phenomena and unpredictable incidents.

In~\cite{wardrop1954method}, Wardrop and Charlesworth proposed the moving observer method that relies on vehicles counting their respective neighbors (near by vehicles). A modernized version of the moving observer method was recently proposed in~\cite{florin2016variant}.~\cite{seo2015estimation} proposed a probe vehicle-based estimation method for obtaining volume-related variables such as flow and density by assuming that a probe vehicle can measure the spacing to its leading one using~\emph{xFCD}~(e.g. range sensors on vehicles).

The Conservation Law~(CL) is a principle in traffic flow theory that relates traffic flow and traffic density and considered as physically reasonable without empirical justification~\cite{papageorgiou1998some}~(equation~\ref{eq:conserv}). 

\begin{equation}
\frac{\partial k}{\partial t} + \frac{\partial q}{\partial x} = 0
\label{eq:conserv}
\end{equation}

Vehicle trajectories obtained from probe vehicle data provide disaggregated information about the traffic. However, such trajectories can be useful information to be incorporated with the conservation law, albeit by ignoring lane-changing and other unexpected behaviour to conclude that the number of vehicles between two probe vehicles will remain the same. 

Ignoring lane-changing can be a reasonable assumption if a sufficiently wide area is considered where effects of lane-changing can be ignored. This idea has been the basis of several streaming-data driven TSE methods.~\cite{coifman2003estimating} used disaggregated data from travel time information and detectors to estimate density. Its notable feature is that it considers and estimates lane-changing flow as well.~\cite{astarita2006motorway,qiu2010estimation} used aggregated and disaggregated probe vehicle data and a limited number of detectors at the entrance and the exit.~\cite{bekiaris2016highway} combined this approach with the KF, whose system model represents the CL. ~\cite{seo2015traffic} combined disaggregated spacing probe vehicle data with the CL.

At an age of near ubiquitous cell phone and on-vehicle sensor penetration, and with the massive emergence of connected (automated) vehicles, this approach might become prevalent in the near future. The limitation of streaming-data-driven approach is that they may require massive streaming data in order to perform accurate estimation. In addition, the approach itself does not have any future prediction capability.

\section{The Proposed Method}\label{tse_proposed}
This section proposes a traffic state estimation method that is tailored for estimating dynamic routing decisions at a single intersection. The proposed method relies on real-time information acquired from the CAVs that attempt to cross the intersection. It should be noted that this method is designed to complement the proposed data-driven signal control method introduced in section~\ref{chpt:signal}.
\subsection{Estimating Routing Decisions}
\begin{figure}[H]
    \centering
    \includegraphics[width=0.75\textwidth]{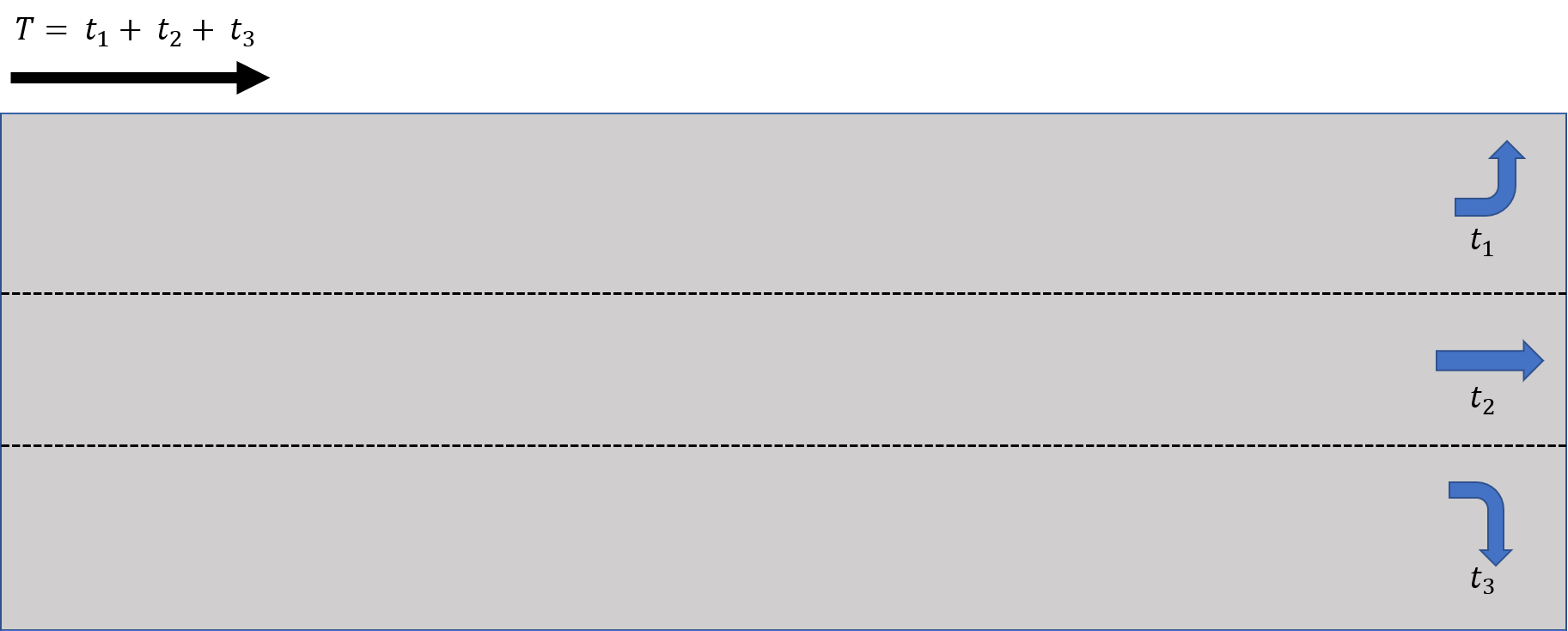}
    \caption{Abstraction of a Road with Three Available Routing Decisions}
    \label{fig:tse_concept}
\end{figure}

This component is designed to estimate the lane group traffic flows~($f_{i}$) as inputs to the signal optimizer introduced in Chapter~\ref{chpt:signal}. Here we assume that the total incoming traffic flow from each leg of the intersection is known either through fixed sensors or communications from neighboring signal controllers. However, the routing decisions of the incoming traffic is unknown. The problem of estimating routing decisions for each leg of the intersection is defined as below:

\begin{definition}
\textbf{Estimate Routing Decisions} Given the total incoming traffic flow~($T$) of a road, estimate portions of traffic that will take each routing decision available~($t_{i}$) such that $\sum t_i = T$.
\end{definition}
Consider a one-way road as in Fig.~\ref{fig:tse_concept} where the vehicles have three routing decision choices to pick from~(i.e. Left Turn, Right Turn, Straight). Assuming the incoming traffic volume is known, the problem is defined as estimating the portions of traffic that select any given routing decision with minimum error.
\subsubsection{Utilizing CAV Information}
CAVs are capable of communicating their intentions to the intersection manager through V2I communication technology. Such information can be directly applied to estimate the traffic routing decisions. A key factor influencing the precision of such estimations is the percentage of CAV vehicles on the road, usually referred to as the Market Penetration Rate~(MPR). Fig.~\ref{fig:tse_static} demonstrates the results of estimating the static routing decisions of a road similar to the one depicted in Fig.~\ref{fig:tse_concept}, utilizing CAV information containing the CAV's routing decision. The estimates are updated according to equation~\ref{eq:tse_update}.
\begin{equation}
    \label{eq:tse_update}
    \tilde{t_{i}} = \frac{t_{cv,i}}{T_{cv}}
\end{equation}
In the above equation, $\tilde{t_{i}}$ is the estimated portion of traffic taking the $i$th routing decision, $t_{cv,i}$ is the portion of the CV traffic that has taken the $i$th routing decision and $T{cv}$ is the total CV traffic.
\begin{figure}[H]

\centering
\subfloat[MPR = 10\%]{
{\includegraphics[width=0.7\textwidth]{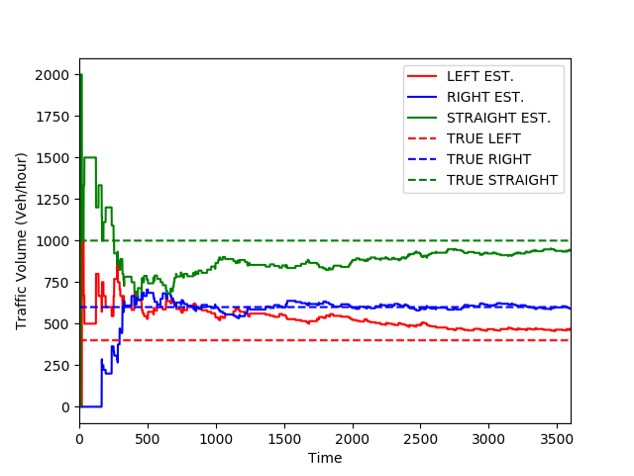}}}
\vspace{-5pt}
\subfloat[MPR = 20\%]{
{\includegraphics[width=0.7\textwidth]{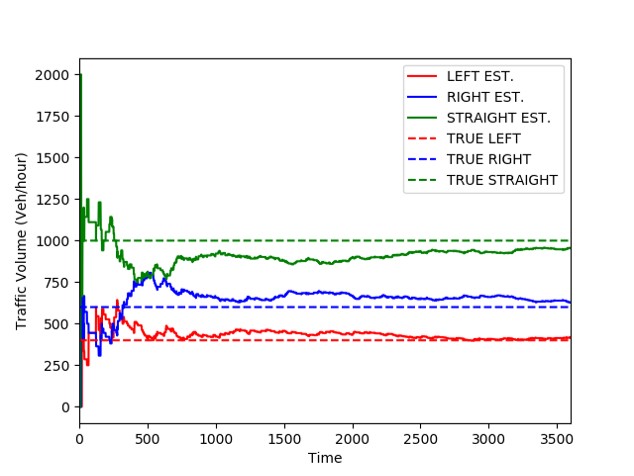}}}
\vspace{-5pt}
\subfloat[MPR = 30\%]{
{\includegraphics[width=0.7\textwidth]{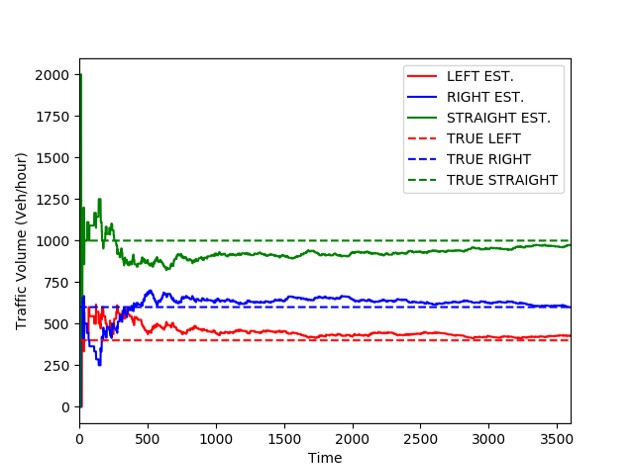}}}

\caption{Estimating Static Routing Decisions with CAV Data}
\label{fig:tse_static}
\end{figure}

\begin{figure}[H]
\ContinuedFloat
\centering
\subfloat[MPR = 40\%]{
{\includegraphics[width=0.7\textwidth]{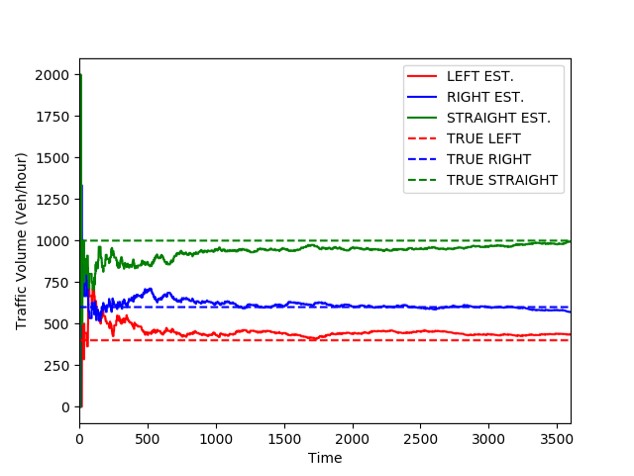}}}
\vspace{-5pt}
\subfloat[MPR = 50\%]{
{\includegraphics[width=0.7\textwidth]{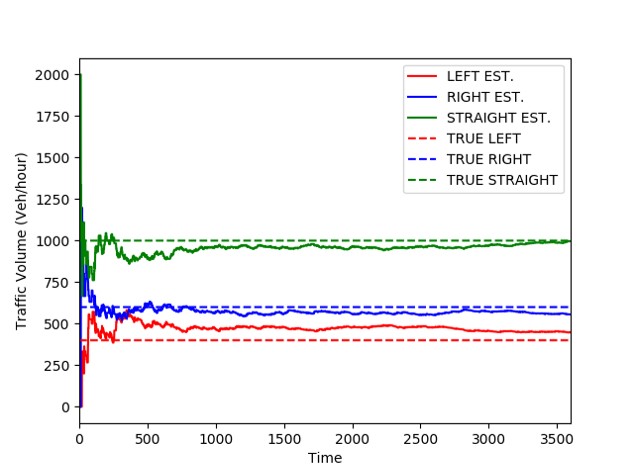}}}
\vspace{-5pt}
\subfloat[MPR = 60\%]{
{\includegraphics[width=0.7\textwidth]{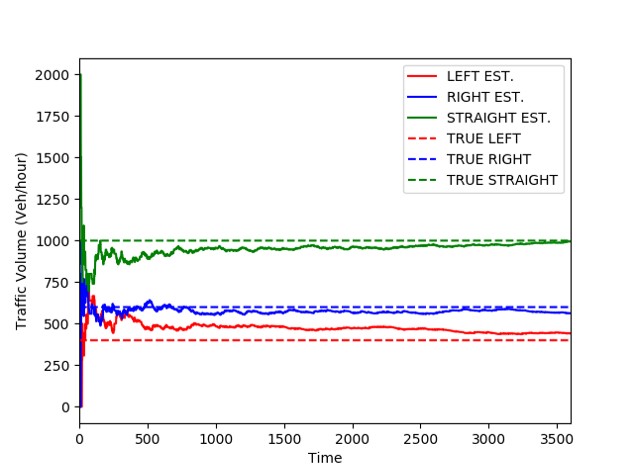}}}

\caption{Estimating Static Routing Decisions with CAV Data}
\end{figure}

\begin{figure}[H]
\ContinuedFloat
\centering
\subfloat[MPR = 70\%]{
{\includegraphics[width=0.6\textwidth]{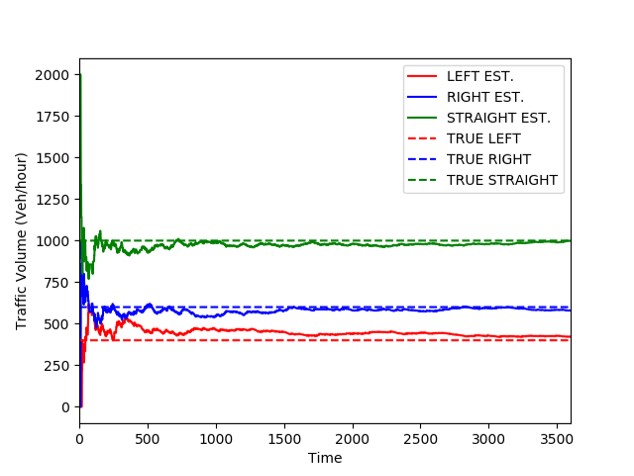}}}
\vfill
\subfloat[MPR = 80\%]{
{\includegraphics[width=0.6\textwidth]{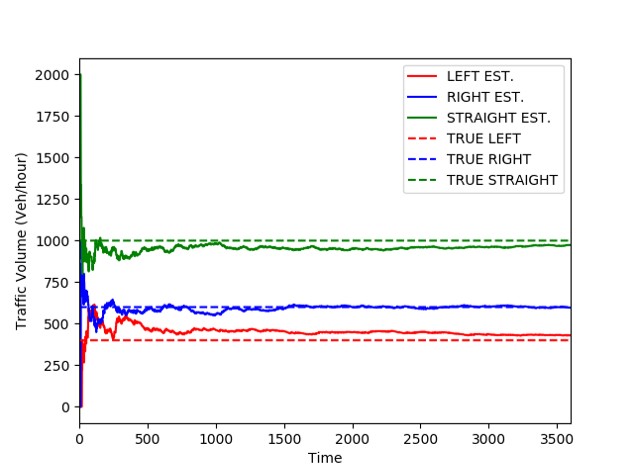}}}
\vfill
\subfloat[MPR = 90\%]{
{\includegraphics[width=0.6\textwidth]{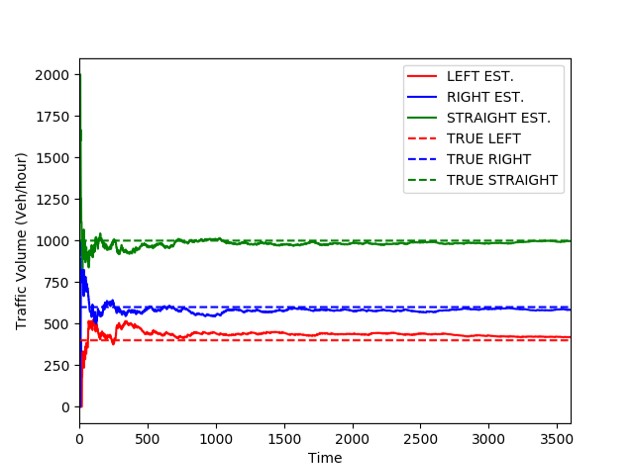}}}

\caption{Estimating Static Routing Decisions with CAV Data}
\end{figure}
In the above figures, solid lines represent estimated values, color-coded based on the routing decision type~(e.g. Left, Right and Straight). Actual values are similarly color-coded and represented by dashed lines.

\begin{figure}[H]
\centering
\includegraphics[width=0.75\textwidth]{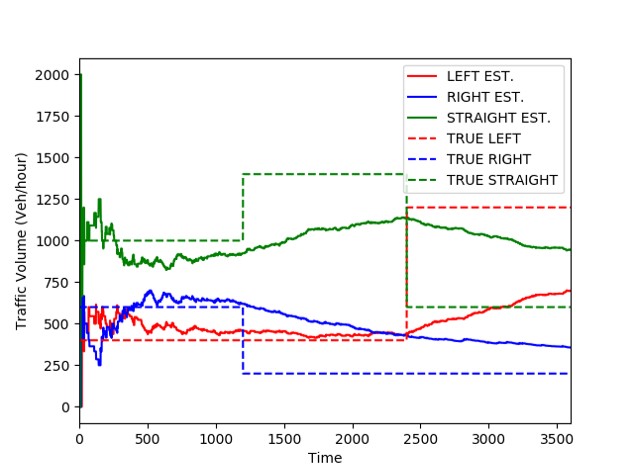}
\caption{Estimating Dynamic Routing Decisions with CAV Data}
\label{fig:tse_dynamic}
\end{figure}

\begin{figure}[H]
\centering
\includegraphics[width=0.75\textwidth]{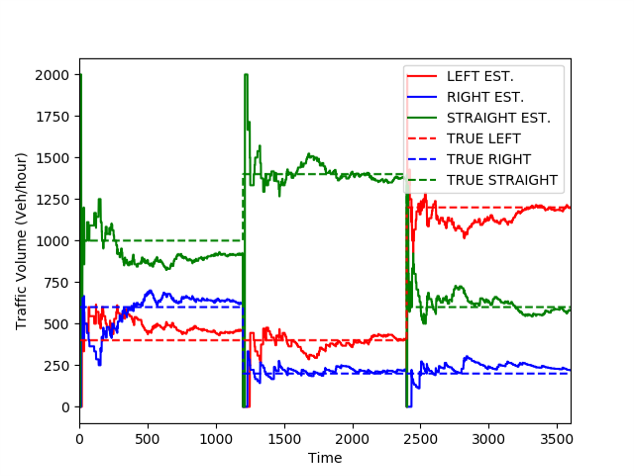}
\caption{Ideal reaction to changes in routing decisions}
\label{fig:tse_dynamic_ideal}
\end{figure}

The simulation results show that the estimations are fairly accurate even for low MPRs, however, one must note that in a realistic scenario these decisions are dynamic. Let's see how well the estimator responds to dynamic decisions. Fig.~\ref{fig:tse_dynamic} demonstrates the estimations along with the actual dynamic decisions for a market penetration of~$30\%$. In this simulation, routing decisions change at two times during the simulation; once at minute 20 and another time at minute 40 of the one hour simulation. As evident in the graphics, the estimations fail to track the changes in a timely manner. 
There are two factors contributing to this delay in response: First the algorithm keeps an infinite record of CAV information which delays the convergence to the new value as the new CAV decisions come in. This effect can be minimized by flushing the algorithm's memory when a significant change to the overall routing decisions is detected. Such ideal response to changes is depicted in~\ref{fig:tse_dynamic_ideal}.

However, detecting such changes is non-trivial and itself part of the problem that the algorithm is trying to solve. The second factor is that low market penetration rates negatively affect the accuracy of the estimations.

\begin{figure}[H]
\centering
\includegraphics[width=0.75\textwidth]{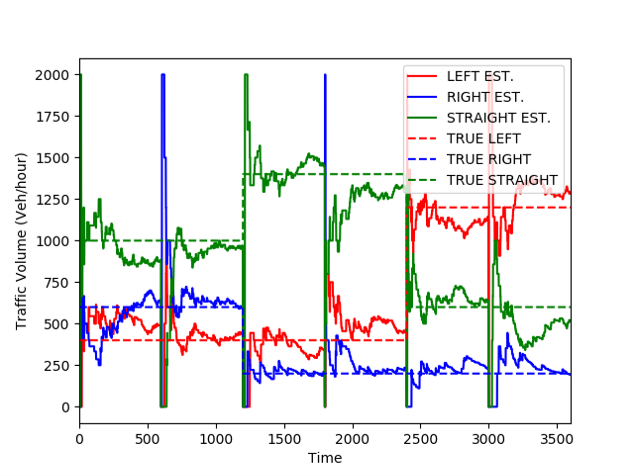}
\caption{Periodic Flushing of CAV Information Records}
\label{fig:tse_dynamic_reset}
\end{figure}

A heuristic method is proposed to address the first factor. The proposed heuristic flushed the algorithm's memory of CAV information periodically. The intervals between these operations is defined as a configuration property named~$p$. Fig.~\ref{fig:tse_dynamic_reset} demonstrates the algorithm's estimates for an identical simulation with the flushing interval set to 10 minutes.

As evident in the above figure, memory flushing helps decrease convergence time significantly; However, it also causes significant undershoots and overshoots at the time of flushing~(i.e. timestamps: 10, 20, 30, 40, 50 minute).

To reduce these effects, another configuration property is proposed to let the algorithm keep its estimation prior to the flushing for a certain period after the memory flush. This configuration property is named~$r$. This period of time will allow the algorithm begin to track the reference value given its limited number of samples while keeping the premature estimate from being published.

Algorithm~\ref{alg:tse_alg} shows the routing decision estimator algorithm in pseudo code.
\begin{algorithm}[H]
  \noindent\begin{minipage}{\linewidth}
   \caption{Routing Decision Estimator}
    \begin{algorithmic}[1]
    \Function{Estimator}{}
	\While{$True$}
		\State $F = getTraffic()$ \footnote{F is the total incoming traffic volume}
		\State $V_{cav} = getCAVInfo()$\footnote{$V_{cav}$ is an array containing all cav information}
		\State $\tilde{F} = updateEstimation(F,V_{cav})$
		\If{$t\%p == 0$}
			\State $resetCAVMemory()$\footnote{Reset CAV Memory at each interval defined by parameter $p$}
			\State $\tilde{F}_(t,t+r) = \tilde{F}_(t)$\footnote{Keep current estimations for $r$ minutes}
	    \EndIf
        \State $t++$
        \EndWhile
       \EndFunction
       \Function{updateEstimation}{$F,V_{cav}$}

       \State $\tilde{F}_{left} = \frac{|V_{cav,left}|}{|V_{cav}|} \times F$
       \State $\tilde{F}_{right} = \frac{|V_{cav,right}|}{|V_{cav}|} \times F$
       \State $\tilde{F}_{straight} = \frac{|V_{cav,straight}|}{|V_{cav}|} \times F$
  	\State $\tilde{F} = \{\tilde{F}_{left}, \tilde{F}_{right}, \tilde{F}_{straight}\}$
        \State $Return(\tilde{F})$
       \EndFunction

\end{algorithmic}
    \label{alg:tse_alg}
\end{minipage}
\end{algorithm}

\begin{figure}[H]
\centering
\includegraphics[width=0.75\textwidth]{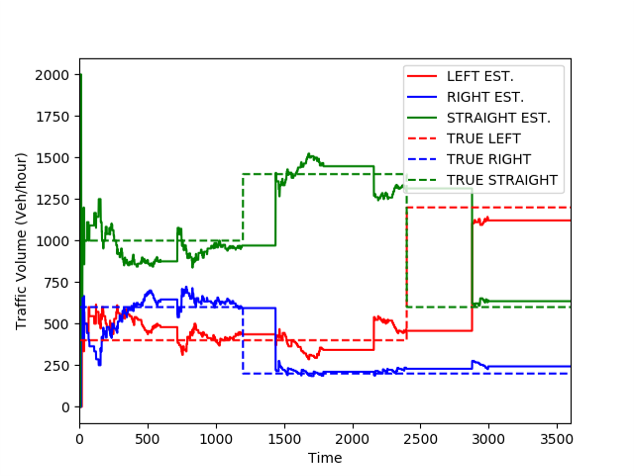}
\caption{Temporarily Rely on Previous Estimates $p = 10m, r = 2m$}
\label{fig:tse_dynamic_reset_remember}
\end{figure}

Fig.~\ref{fig:tse_dynamic_reset_remember} demonstrates an identical simulation with the following parameter settings:
\begin{itemize}
    \item $p = 10 minutes$
    \item $r = 2 minutes$
\end{itemize}

One should note that the step-like changes in actual routing decisions are an extreme case that does not happen in a real-world scenario. The choice of step function was made to better demonstrate the algorithm's ability to track the changes. The introduction of the new parameter~$r$ eliminates the negative effect of memory of flush. However, both parameters require tuning for new roads and known MPRs. The next section provides an example of parameter tuning for different MPRs on the same abstract road defined in this chapter.
\subsection{Parameter Tuning}
Multiple scenarios are designed and simulated in PTV Vissim software to find out the parameters that result in higher accuracy in estimating dynamic routing decisions.

Table~\ref{tbl:tse_tuning} lists the parameter and scenario configurations considered for parameter tuning simulations in Vissim. Root mean squared error~(RMSE, Eq. ~\ref{eq:rmse}), is used for comparisons. However, for easier visual comparison, Normalized root mean square error~(NRMSE) is selected as defined in equation~\ref{eq:nrmse}.
\begin{equation}
\label{eq:rmse}
     RMSE = \sqrt{\frac{1}{n}\Sigma_{i=1}^{n}{(y_i -\tilde{y})^2}}
\end{equation}
\begin{equation}
\label{eq:nrmse}
     NRMSE = \frac{RMSE}{y_{max}-y_{min}}
\end{equation}

\begin{table*}[ht]
\centering
\caption{Parameter Tuning Scenario Settings}
\label{tbl:tse_tuning}
\resizebox{\textwidth}{!}{%
\begin{tabular}{@{}ccccc@{}}
\toprule
\multicolumn{5}{c}{Algorithm  Parameters} \\ \midrule
 Name & Notation & Value(s) & Number of Values & Unit \\ \midrule
Memory Flush Interval & $p$ & (5,10) & 6 & minute \\ \midrule
Keep Estimation Interval & $r$ & (1,4) & 4 & minute \\ \midrule
\multicolumn{5}{c}{Simulation Settings} \\ \midrule
 Name & Notation & Value(s) & Number of Values & Unit \\ \midrule
Total traffic & $F$ & 2000  & 1 & veh/hour \\ \midrule
Market Penetration Rate & $mpr$ & $(0.1,0.9)$& 9 & - \\ \midrule
Decision Change Interval & $T_{route}$ & ${5,10,20}$& 3 & minute \\ \midrule
Simulation Time & $T_{s}$ & 60& 1 & minute \\\bottomrule\\
\multicolumn{1}{c}{\# of Simulations} &\multicolumn{3}{c}{$|p|\times|r|\times|mpr|\times|T_{route}| = 648$} & - \\
\bottomrule
\end{tabular}%
}
\end{table*}

Table~\ref{tbl:tse_tuning_results} lists the results of the parameter tuning simulations for each of the 27 simulation settings.

\begin{table}[!t]
\centering
\caption{Parameter Tuning: Chosen Parameter Values}
\label{tbl:tse_tuning_results}
\centering
        \small
        \setlength{\tabcolsep}{5pt}
\begin{tabular}{@{}cccccc@{}}
\toprule
\multicolumn{6}{c}{Parameter Tuning Best Results} \\ \midrule
sim \# & $mpr$ & $T_{route}$  & $p_{best}$ & $r_{best}$ & $NRMSE$\\ \midrule
1 & 0.1 & 5 & 5 & 2 & 0.31 \\ \midrule
2 & 0.1 & 10 & 7 & 3 & 0.25 \\ \midrule
3 & 0.1 & 20 & 10 & 4 & 0.18 \\ \midrule
4 & 0.2 & 5 & 5 & 3 & 0.22 \\ \midrule
5 & 0.2 & 10 & 7 & 3 & 0.18 \\ \midrule
6 & 0.2 & 20 & 10 & 3 & 0.14 \\ \midrule
7 & 0.3 & 5 & 5 & 2 & 0.19 \\ \midrule
8 & 0.3 & 10 & 7 & 2 & 0.15 \\ \midrule
9 & 0.3 & 20 & 10 & 2 & 0.12 \\ \midrule
10 & 0.4 & 5 & 5 & 2 & 0.15 \\ \midrule
11 & 0.4 & 10 & 7 & 2 & 0.12 \\ \midrule
12 & 0.4 & 20 & 10 & 2 & 0.10 \\ \midrule
13 & 0.5 & 5 & 5 & 2 & 0.13 \\ \midrule
14 & 0.5 & 10 & 8 & 2 & 0.10 \\ \midrule
15 & 0.5 & 20 & 10 & 2 & 0.06 \\ \midrule
16 & 0.6 & 5 & 5 & 2 & 0.11 \\ \midrule
17 & 0.6 & 10 & 8 & 1 & 0.09 \\ \midrule
18 & 0.6 & 20 & 10 & 1 & 0.06 \\ \midrule
19 & 0.7 & 5 & 5 & 1 & 0.11 \\ \midrule
20 & 0.7 & 10 & 9 & 1 & 0.08 \\ \midrule
21 & 0.7 & 20 & 10 & 1 & 0.05 \\ \midrule
22 & 0.8 & 5 & 5 & 1 & 0.09 \\ \midrule
23 & 0.8 & 10 & 10 & 1 & 0.06 \\ \midrule
24 & 0.8 & 20 & 10 & 1 & 0.03 \\ \midrule
25 & 0.9 & 5 & 5 & 1 & 0.08 \\ \midrule
26 & 0.9 & 10 & 10 & 1 & 0.05 \\ \midrule
27 & 0.9 & 20 & 10 & 1 & 0.03 \\ \bottomrule
\end{tabular}
\end{table}
\pagebreak

The MPR and frequency of dynamic decisions are the major factors affecting the algorithm's estimation accuracy. The following conclusions can be made about the configuration of the algorithm's parameters $p$ and $r$.
\begin{itemize}
    \item \textbf{$p$} should be configured according to the expected frequency of changes in routing decisions and the estimated market penetration rate. Although such estimations will not always be available, a default value for this parameter can be calculated through simulations using real-world or pure simulation data. A default value of $10m$ is suggested in this paper.
    \item \textbf{$r$} should be configured according to the above mentioned estimations and the value chosen for the $p$ parameter. Simulations showed that generally a value between 1 and 3 minutes works best for all considered scenarios.
\end{itemize}

\begin{figure}[H]
    \centering
    \includegraphics[width=0.9\textwidth]{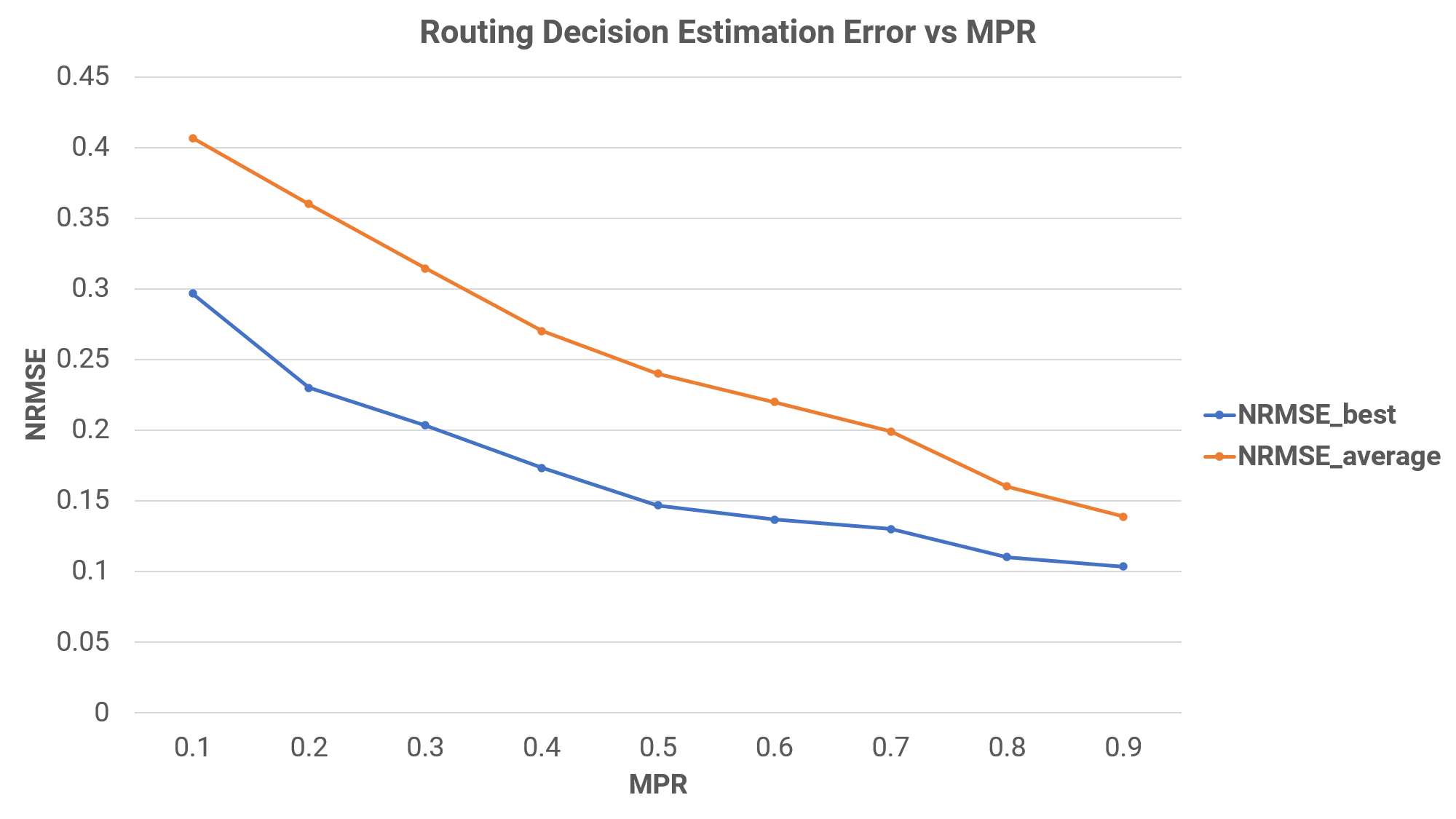}
    \caption{Routing Decisions Estimation Error}
    \label{fig:tse_sim_results}
\end{figure}

Fig.~\ref{fig:tse_sim_results} depicts average error for best and all parameter configurations against the range of values for market penetration rate.
Fig.~\ref{fig:tse_sim_results} shows that the algorithm, if properly tuned, can provide estimations with errors as low as (20\% - 30\%), for lower penetration rates; and errors as low as (5\% - 10\%) for higher penetration rates.

The above figure provides an insight into the effects of parameter tuning on the estimation error. The results show that the impact of parameter tuning has a negative relation with the MPR and its influence is more apparent for MPRs below 40\% where it can decrease the error by up to 35\%. However, the algorithm is less sensitive to parameter values for higher MPRs and can provide estimations with errors below 15\% even with default parameter configuration.

\section{Conclusions}\label{sec:conclusion_5}
This chapter introduced a novel heuristic method for estimation of dynamic routing decisions. The proposed method leverages communications from connected vehicles to estimate aggregated routing decisions based on the samples from CV information. This method provides the user with two configuration properties to periodically flush its memory and to maintain its estimation for a window of time immediately after the memory has been flushed. 

Simulated scenarios were designed and implemented that aim to identify optimal parameter values for the algorithm under each scenario settings. Simulation results show that the optimal parameter values depend on the estimated change frequency of routing decisions as well as the market penetration rate of CVs. Default values were proposed for cases where the estimations are not available or not dependable. 

The simulation results also provide insights into the effect of market penetration rate on the algorithm's accuracy in estimating routing decision. An important takeaway from the results is that the algorithm's performance is significantly boosted with MPR values above $30\%$ at which point average NRMSE for all scenarios (including non-tuned parameters) drops below 30\% and for ideally tuned parameters, NRMSE further dips below 20\%. Therefore, the proposed method can achieve error rates that are reasonably low and acceptable for the optimal signal timing model given the sensitivity analysis results in Section~\ref{sec:input_noise_sensitivity}.
\chapter{Real-Time Data-Driven Adaptive Signal Control}\label{chpt:method}
According to the U.S. Federal Highway Administration, Outdated traffic signal timing currently accounts for more than 10 percent of all traffic delays~\cite{fhwa}. On average, adaptive signal control technologies improve travel time by more than 10 percent. In areas with particularly outdated signal timing, improvements can be 50 percent or more.

Adaptive signal control technologies also react to unexpected events, such as crashes and special events. By adjusting traffic signal timing in real-time to reflect actual conditions on the road, travelers enjoy a more reliable trip. Studies indicate that crashes could be reduced by up to 15 percent through improved signal timing. Adaptive signal control technology can reduce the intersection congestion that causes many crashes.

By receiving and processing data from sensors to optimize and update signal timing settings, adaptive signal control technologies can determine when and how long lights should be green. Adaptive signal control technologies help improve the quality of service that travelers experience.

Adaptive signal control is known for its expensive set up which is mostly associated to the cost of the equipment and their maintenance. This chapter proposes an adaptive signal controller that requires minimal sensor installation and relies on a combination of low cost sensors and CAV data for traffic state estimation. The proposed method utilizes the method proposed in Chapter \ref{chpt:signal} for updating the signal timing settings at each iteration.

\section{The Proposed Method}
This chapter introduces an adaptive signal control system that relies on the signal timing method introduced in Chapter~\ref{chpt:signal} along with a traffic state estimator as introduced in Chapter~\ref{chpt:tse}. The proposed system does not require installation of any additional sensors to estimate the incoming traffic; instead it relies on information received from fixed sensors installed at each leg of the intersection (or information from the neighboring intersections) and the additional routing decision information from connected vehicles.

\begin{figure}[H]
    \centering
    \includegraphics[width=\textwidth]{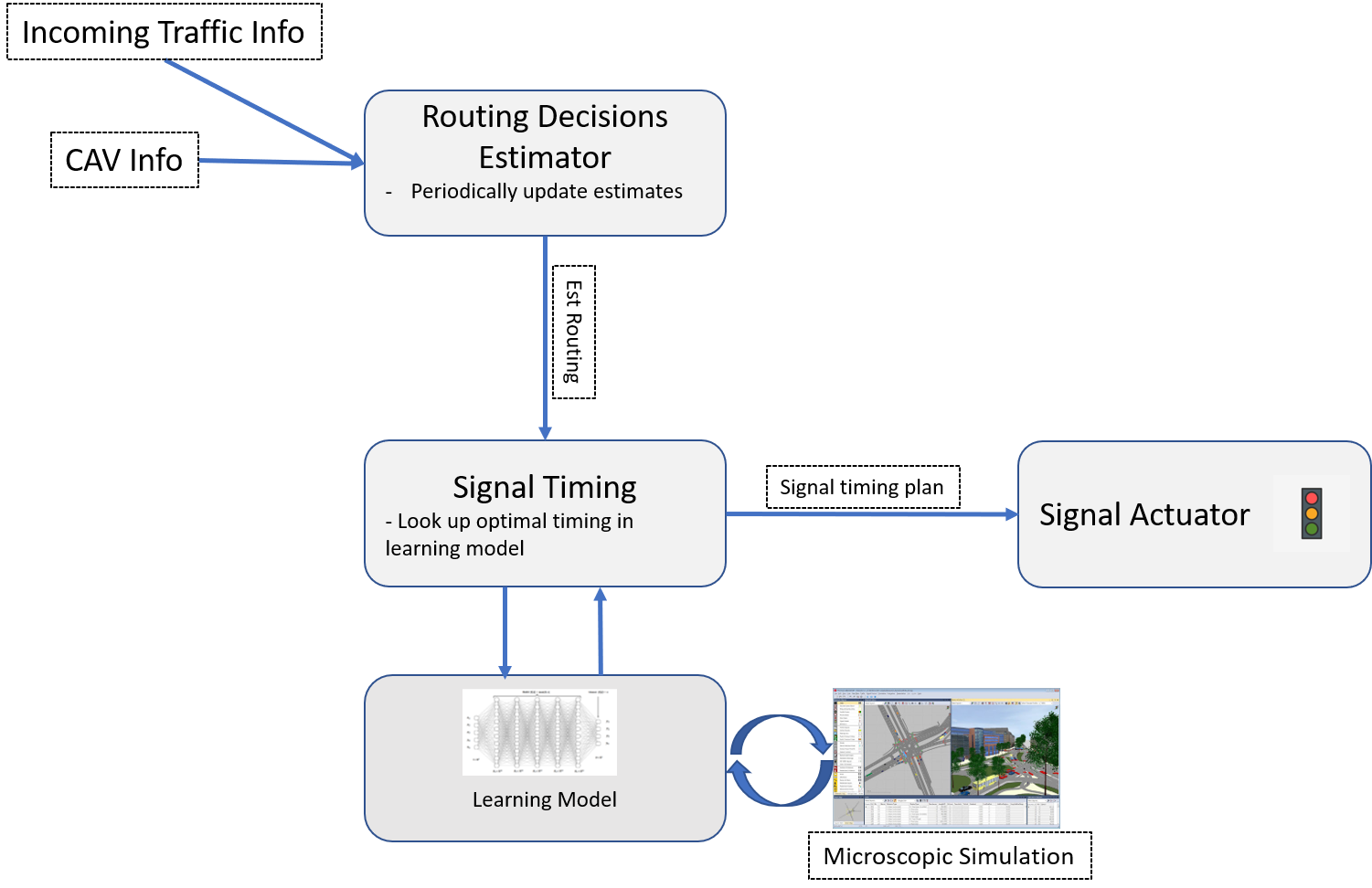}
    \caption{The proposed signal controller at a glance}
    \label{fig:controller_holistic}
\end{figure}

Fig.~\ref{fig:controller_holistic} shows a holistic view of the controller architecture.

The main component of the proposed method is depicted in Algorithm ~\ref{alg:main}. 
\begin{algorithm}[H]
  \noindent\begin{minipage}{\linewidth}
   \caption{Adaptive Signal Controller}
    \begin{algorithmic}[1]
    \Function{Signal Controller}{}
	\While{$True$}
	    \State $F = Lane group flows$
	    \State $T = updateTrafficFromSensors()$
	    \ForAll{$f_{i} \in F$}
            \State $f_{i} = getFlows(T)$ \footnote{A call to algorithm \ref{alg:tse_alg}}
        \EndFor
		\State $G = getGreenTimes(F)$ \footnote{A call to the trained model in chapter \ref{chpt:signal}}
		\State $updateSignals()$ \footnote{Update the signal timing settings for the next cycle}
        \EndWhile
       \EndFunction
\end{algorithmic}
    \label{alg:main}
\end{minipage}
\end{algorithm}
The algorithm runs an infinite loop that updates signal timing settings according to the state of the traffic and the recommended signal timings of the algorithm proposed in chapter \ref{chpt:signal}.

In each iteration, the incoming traffic information are read from the sensors (or neighboring intersections), routing decisions are estimated and converted into lane group flows. The lane group flows are passed as arguments to the signal timing algorithm and the calculated signal timing settings (including the new cycle time, green time, etc.) are passed on to the traffic light actuators for the next cycle.

Algorithm parameters, input and output are listed in Table~\ref{tbl:alg_params}.
\begin{table*}[ht]
\small
\centering
\caption{Algorithm Parameters, Inputs and Outputs}
\label{tbl:alg_params}
\resizebox{\textwidth}{!}{%

\begin{tabular}{@{}cccc@{}}
\toprule
\textbf{Name}& \textbf{Notation} & \textbf{Type} & \textbf{Description} \\
\bottomrule
\multicolumn{4}{c}{\textbf{Input}} \\ \bottomrule
Traffic Volume & T & array & Incoming traffic volume of each leg of the intersection\\ \midrule
Lost Time & L & number & Total lost time of the intersection\\ \midrule
 
\multicolumn{4}{c}{\textbf{Output}} \\ \bottomrule
Cycle Length & C & number & The length of the cycle in seconds\\ \midrule
Green Times & G & array & Green times of each phase\\ \midrule

\multicolumn{4}{c}{\textbf{Parameters}} \\ \bottomrule
Memory Flush Interval & $p$ & number & Intervals at which the cav info is flushed\\ \midrule
Remember Time & $r$ & number & Intervals at which the previous estimations will be used\\
\bottomrule
\end{tabular}%
}
\end{table*}

\begin{figure}[H]
    \centering
    \includegraphics[width=0.7\textwidth]{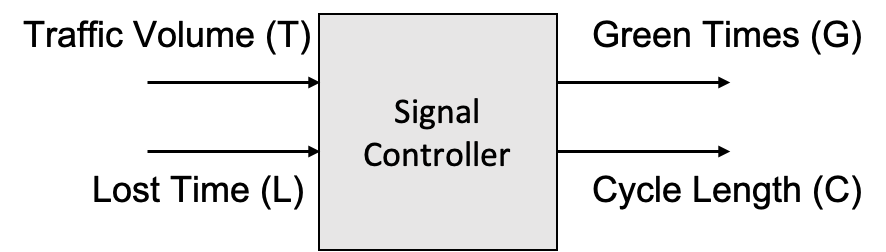}
    \caption{The proposed method as a black box model}
    \label{fig:alg_blackbox}
\end{figure}

Fig.~\ref{fig:alg_blackbox} depicts the system as a black box model.

The rest of this chapter focuses on validating the proposed method's performance under different traffic levels and different levels of dynamism in vehicles' aggregated routing decisions; and studies the effects of different market CAV market penetration rates on its performance.
\section{Simulated Experiments}
The simulations are performed on two intersections under a range of traffic levels, frequency of routing decision changes and market penetration rates. The proposed method is tested with default and pre-tuned parameters. The final results are compared to the case where the actual routing decisions are known. Average vehicular delay is chosen as the metric for comparisons as it is also the metric (target variable) used for the training of the signal timing model in Chapter~\ref{chpt:signal}.

\begin{figure}[H]
\hspace*{\fill}

    \centering
    \begin{minipage}{0.4\textwidth}
       \centering
    \includegraphics[width=3in]{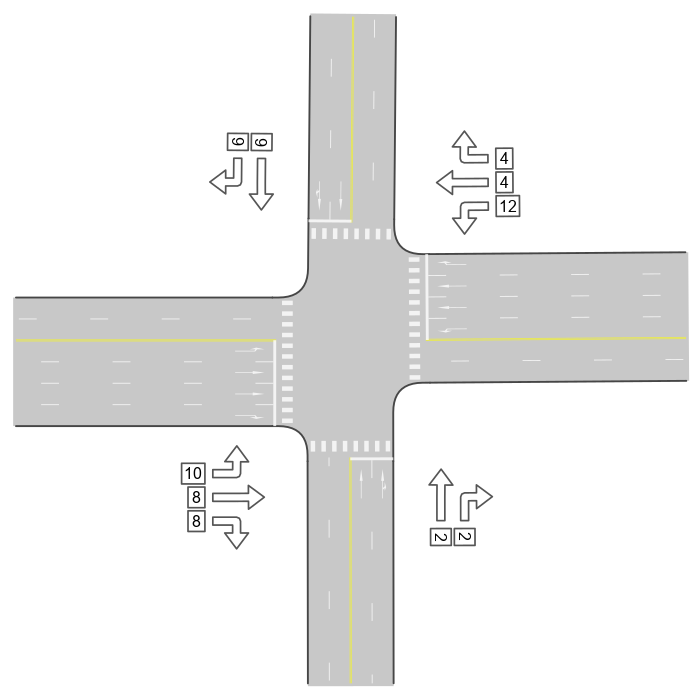}
    \caption{Intersection Model 1}
    \label{fig:int1}
    \end{minipage}\hfill
    \begin{minipage}{0.4\textwidth}
     \centering
             \vspace{0.8in}
    \includegraphics[width=3in]{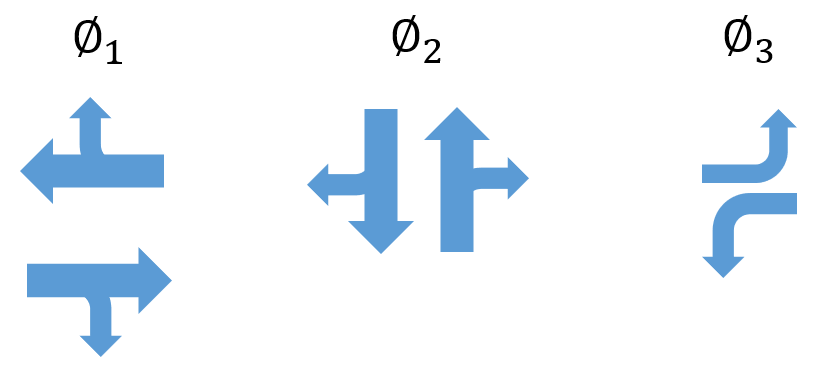}
        \vspace{0.7in}
    \caption{Intersection 1: Phase Diagram}
    \label{fig:int1_ph}
    \end{minipage}
    \hspace*{\fill}
\end{figure}

\begin{figure}[H]
\hspace*{\fill}
    \centering
    \begin{minipage}{0.4\textwidth}
     \centering
    \includegraphics[width=3in]{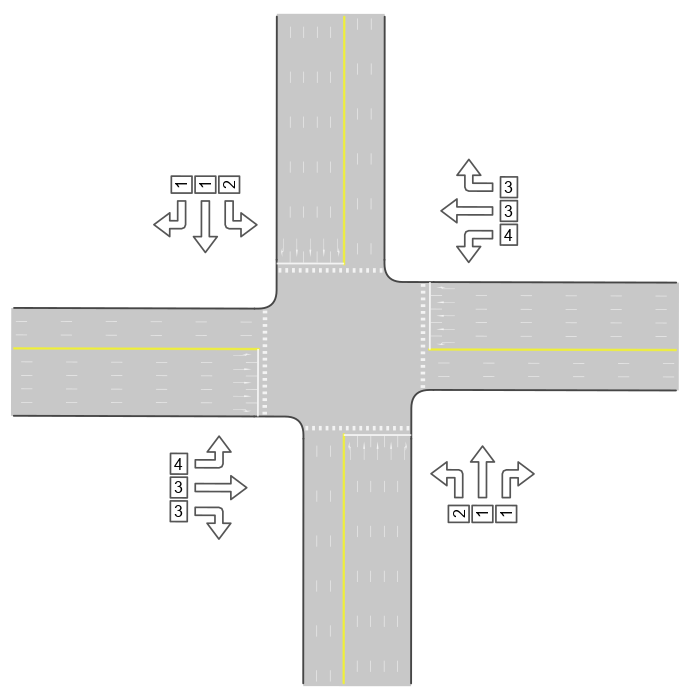}
    \caption{Intersection Model 2}
    \label{fig:int2}
    \end{minipage}\hfill
    \begin{minipage}{0.4\textwidth}
      \centering
              \vspace{0.7in}
    \includegraphics[width=3in]{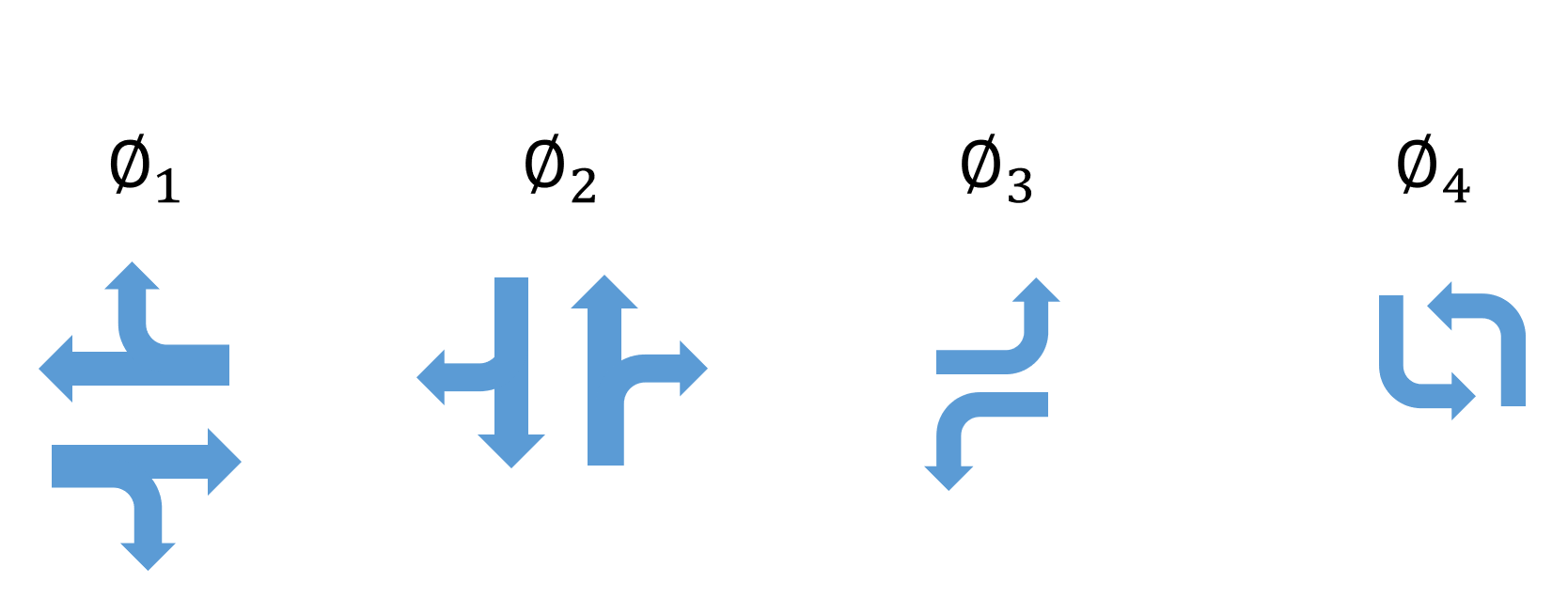}
            \vspace{0.9in}

    \caption{Intersection 2: Phase Diagram}
    \label{fig:int2_ph}
    \end{minipage}
        \hspace*{\fill}
\end{figure}

Figs.~\ref{fig:int1}, \ref{fig:int1_ph}, \ref{fig:int2} and \ref{fig:int2_ph} show the two intersection models (snapshot from Vistro software) and the phase diagram of their signal controllers.

Intersection 1 has four legs; East-West legs both have 5 lanes, with two lanes going straight, one dedicated lane turning left and one dedicated lane turning right. North-South legs both have one two lanes, one going straight and the other turning right. A 3-phase signal controller is considered for this model as depicted in Fig.~\ref{fig:int1_ph}.

Intersection 2 has four symmetrical legs. Each leg has 3 lanes going straight and 2 dedicated lanes for left and right turns respectively. A 4-phase signal control scheme is considered for this intersection as depicted in Fig.~\ref{fig:int2_ph}.

Table~\ref{tbl:comp_methods} list the configurations considered for comparisons.

\begin{table}[htbp]
\centering
\caption{Simulations: Method Configurations}
\resizebox{\textwidth}{!}{%
\begin{tabular}{@{}|l|l|@{}}
\toprule
\textbf{Notation} & \textbf{Description}                                                        \\ \midrule
$HCM$   & Routing decisions are known                                                 \\ \midrule
$DDTC_{known}$   & Routing decisions are known                                                 \\ \midrule
$DDTC_{default}$ & Routing decision unknown. Default parameters used for state estimation.     \\ \midrule
$DDTC_{tuned}$   & Routing decision unknown. Tuned parameter values used for state estimation. \\ \bottomrule
\end{tabular}%
}
\label{tbl:comp_methods}
\end{table}
The $HCM$ method calculates the intersection parameters according to the highway capacity manual's guidelines; PTV Vistro software is used to generate parameter values and the expected delay is verified through microscopic simulation in Vissim.

$DDTC$ refers to the proposed method named \textit{Data-Driven Traffic Control}, and the subscripts of \emph{known}, \emph{default} and \emph{tuned} refer to the three cases of the routing decisions being known to the algorithm, not known and estimated with default parameters or not known and estimated with pre-tuned parameter values, respectively.

$DDTC_{known}$ is expected to outperform the other candidates and will be used as a baseline to evaluate the cases where the routing decisions unknown.
\subsection{Simulation Scenarios}
The simulation scenarios were designed to cover a wide range of traffic levels, frequency of change in routing decisions and MPRs.

Tables~\ref{tbl:final_sim_int1} and \ref{tbl:final_sim_int2} list the scenarios for intersections 1 and 2, respectively. The scenarios are designed to evaluate the system holistically. Three traffic levels are considered to cover unsaturated, saturated and over-saturated scenarios for each of the intersections. Market penetration rates of 10\%, 50\% and 90\% are considered to evaluate the effects of MPR on the traffic state estimator and as a result on the entire system's performance as indicated by the chosen evaluation metric of average vehicular delay.

Decision change interval (DCI) is another scenario parameter chosen to represent the inverse of frequency at which the overall routing decisions change for all legs of the intersection. An array of pre-defined routing decisions are considered for each leg and at the beginning of each decision change interval, the new decision rates are applied for each turning movement. For example if $DCI = 10m$ for a given leg of the intersection, the decision rate for each turning movement is updated every 10 minutes.

\begin{table*}[!t]
\centering
\caption{Intersection 1: Comparison Scenarios}
\label{tbl:final_sim_int1}
\resizebox{\textwidth}{!}{%
\begin{tabular}{@{}ccccccc@{}}
\toprule
\textbf{Scenario \#} & \textbf{Total Traffic (veh/h)} & \textbf{MPR(\%)} & \textbf{DCI(minute/change)} & \textbf{Traffic Level} & \textbf{MPR Level} & \textbf{Routing Dynamism Level}\\ \midrule

1 & \multirow{9}{*}{4000} & 10\% & 20 & Low & Low & Low \\ \cmidrule(r){1-1} \cmidrule(l){3-7} 
2 &  & 10\% & 10 & Low & Low & Medium \\ \cmidrule(r){1-1} \cmidrule(l){3-7} 
3 &  & 10\% & 5 & Low & Low & High \\ 

\cmidrule(r){1-1} \cmidrule(l){3-7} 
4 &  & 50\% & 20 & Low & Medium & Low \\ 
\cmidrule(r){1-1} \cmidrule(l){3-7} 
5 &  & 50\% & 10 & Low & Medium & Medium \\ \cmidrule(r){1-1}
\cmidrule(l){3-7} 
6 &  & 50\% & 5 & Low & Medium & High \\ 

\cmidrule(r){1-1} \cmidrule(l){3-7} 
7 &  & 90\% & 20 & Low & High & Low \\ \cmidrule(r){1-1} \cmidrule(l){3-7} 
8 &  & 90\% & 10 & Low & High & Medium \\ \cmidrule(r){1-1} \cmidrule(l){3-7} 
9 &  & 90\% & 5 & Low & High & High \\ \midrule

10 & \multirow{9}{*}{6000} & 10\% & 20 & Medium & Low & Low \\ \cmidrule(r){1-1} \cmidrule(l){3-7} 
11 &  & 10\% & 10 & Medium & Low & Medium \\ \cmidrule(r){1-1} \cmidrule(l){3-7} 
12 &  & 10\% & 5 & Medium & Low & High \\ 

\cmidrule(r){1-1} \cmidrule(l){3-7} 
13 &  & 50\% & 20 & Medium & Medium & Low \\ 
\cmidrule(r){1-1} \cmidrule(l){3-7} 
14 &  & 50\% & 10 & Medium & Medium & Medium \\ \cmidrule(r){1-1}
\cmidrule(l){3-7} 
15 &  & 50\% & 5 & Medium & Medium & High \\ 

\cmidrule(r){1-1} \cmidrule(l){3-7} 
16 &  & 90\% & 20 & Medium & High & Low \\ \cmidrule(r){1-1} \cmidrule(l){3-7} 
17 &  & 90\% & 10 & Medium & High & Medium \\ \cmidrule(r){1-1}
\cmidrule(l){3-7} 
18 &  & 90\% & 5 & Medium & High & High \\ \midrule

19 & \multirow{9}{*}{8000} & 10\% & 20 & High & Low & Low \\ \cmidrule(r){1-1} \cmidrule(l){3-7} 
20 &  & 10\% & 10 & High & Low & Medium \\ \cmidrule(r){1-1} \cmidrule(l){3-7} 
21 &  & 10\% & 5 & High & Low & High \\ 

\cmidrule(r){1-1} \cmidrule(l){3-7} 
22 &  & 50\% & 20 & High & Medium & Low \\ 
\cmidrule(r){1-1} \cmidrule(l){3-7} 
23 &  & 50\% & 10 & High & Medium & Medium \\ \cmidrule(r){1-1}
\cmidrule(l){3-7} 
24 &  & 50\% & 5 & High & Medium & High \\ 

\cmidrule(r){1-1} \cmidrule(l){3-7} 
25 &  & 90\% & 20 & High & High & Low \\ \cmidrule(r){1-1} \cmidrule(l){3-7} 
26 &  & 90\% & 10 & High & High & Medium \\ \cmidrule(r){1-1}
\cmidrule(l){3-7} 
27 &  & 90\% & 5 & High & High & High \\ \bottomrule

\end{tabular}%
}
\end{table*}

\begin{table*}[!t]
\centering
\caption{Intersection 2: Comparison Scenarios}
\label{tbl:final_sim_int2}
\resizebox{\textwidth}{!}{%
\begin{tabular}{@{}ccccccc@{}}
\toprule
\textbf{Scenario \#} & \textbf{Total Traffic (veh/h)} & \textbf{MPR(\%)} & \textbf{DCI(minute/change)} & \textbf{Traffic Level} & \textbf{MPR Level} & \textbf{Routing Dynamism Level}\\ \midrule

1 & \multirow{9}{*}{6000} & 10\% & 20 & Low & Low & Low \\ \cmidrule(r){1-1} \cmidrule(l){3-7} 
2 &  & 10\% & 10 & Low & Low & Medium \\ \cmidrule(r){1-1} \cmidrule(l){3-7} 
3 &  & 10\% & 5 & Low & Low & High \\ 

\cmidrule(r){1-1} \cmidrule(l){3-7} 
4 &  & 50\% & 20 & Low & Medium & Low \\ 
\cmidrule(r){1-1} \cmidrule(l){3-7} 
5 &  & 50\% & 10 & Low & Medium & Medium \\ \cmidrule(r){1-1}
\cmidrule(l){3-7} 
6 &  & 50\% & 5 & Low & Medium & High \\ 

\cmidrule(r){1-1} \cmidrule(l){3-7} 
7 &  & 90\% & 20 & Low & High & Low \\ \cmidrule(r){1-1} \cmidrule(l){3-7} 
8 &  & 90\% & 10 & Low & High & Medium \\ \cmidrule(r){1-1} \cmidrule(l){3-7} 
9 &  & 90\% & 5 & Low & High & High \\ \midrule

10 & \multirow{9}{*}{9000} & 10\% & 20 & Medium & Low & Low \\ \cmidrule(r){1-1} \cmidrule(l){3-7} 
11 &  & 10\% & 10 & Medium & Low & Medium \\ \cmidrule(r){1-1} \cmidrule(l){3-7} 
12 &  & 10\% & 5 & Medium & Low & High \\ 

\cmidrule(r){1-1} \cmidrule(l){3-7} 
13 &  & 50\% & 20 & Medium & Medium & Low \\ 
\cmidrule(r){1-1} \cmidrule(l){3-7} 
14 &  & 50\% & 10 & Medium & Medium & Medium \\ \cmidrule(r){1-1}
\cmidrule(l){3-7} 
15 &  & 50\% & 5 & Medium & Medium & High \\ 

\cmidrule(r){1-1} \cmidrule(l){3-7} 
16 &  & 90\% & 20 & Medium & High & Low \\ \cmidrule(r){1-1} \cmidrule(l){3-7} 
17 &  & 90\% & 10 & Medium & High & Medium \\ \cmidrule(r){1-1}
\cmidrule(l){3-7} 
18 &  & 90\% & 5 & Medium & High & High \\ \midrule

19 & \multirow{9}{*}{12000} & 10\% & 20 & High & Low & Low \\ \cmidrule(r){1-1} \cmidrule(l){3-7} 
20 &  & 10\% & 10 & High & Low & Medium \\ \cmidrule(r){1-1} \cmidrule(l){3-7} 
21 &  & 10\% & 5 & High & Low & High \\ 

\cmidrule(r){1-1} \cmidrule(l){3-7} 
22 &  & 50\% & 20 & High & Medium & Low \\ 
\cmidrule(r){1-1} \cmidrule(l){3-7} 
23 &  & 50\% & 10 & High & Medium & Medium \\ \cmidrule(r){1-1}
\cmidrule(l){3-7} 
24 &  & 50\% & 5 & High & Medium & High \\ 

\cmidrule(r){1-1} \cmidrule(l){3-7} 
25 &  & 90\% & 20 & High & High & Low \\ \cmidrule(r){1-1} \cmidrule(l){3-7} 
26 &  & 90\% & 10 & High & High & Medium \\ \cmidrule(r){1-1}
\cmidrule(l){3-7} 
27 &  & 90\% & 5 & High & High & High \\ \bottomrule

\end{tabular}%
}
\end{table*}

\subsection{Simulation Results}
\begin{figure}[H]
    \centering
    \includegraphics[width=\textwidth]{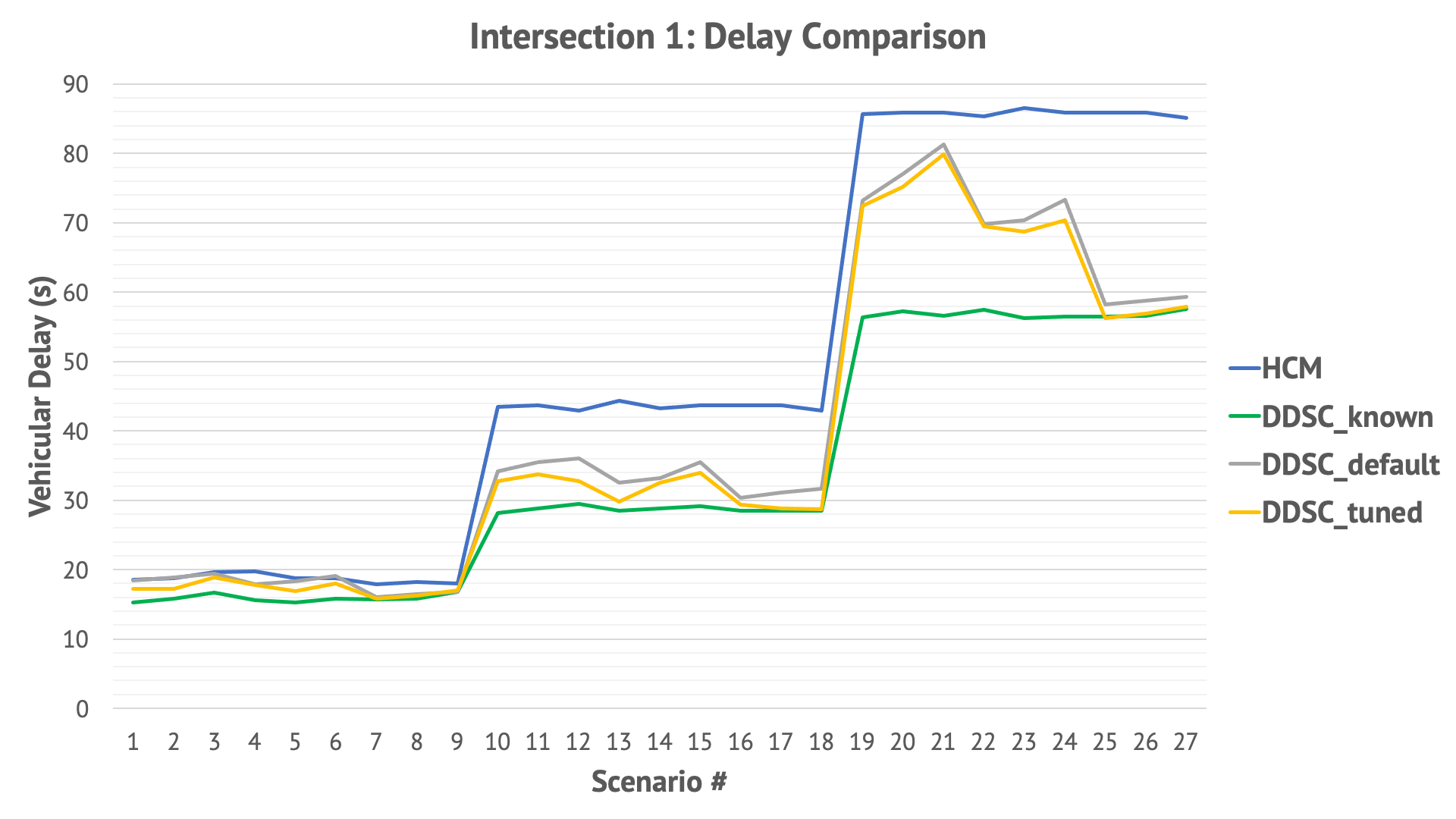}
    \caption{Intersection 1: Average Vehicular Delays}
    \label{fig:int1_results}
\end{figure}

\begin{figure}[H]
    \centering
    \includegraphics[width=\textwidth]{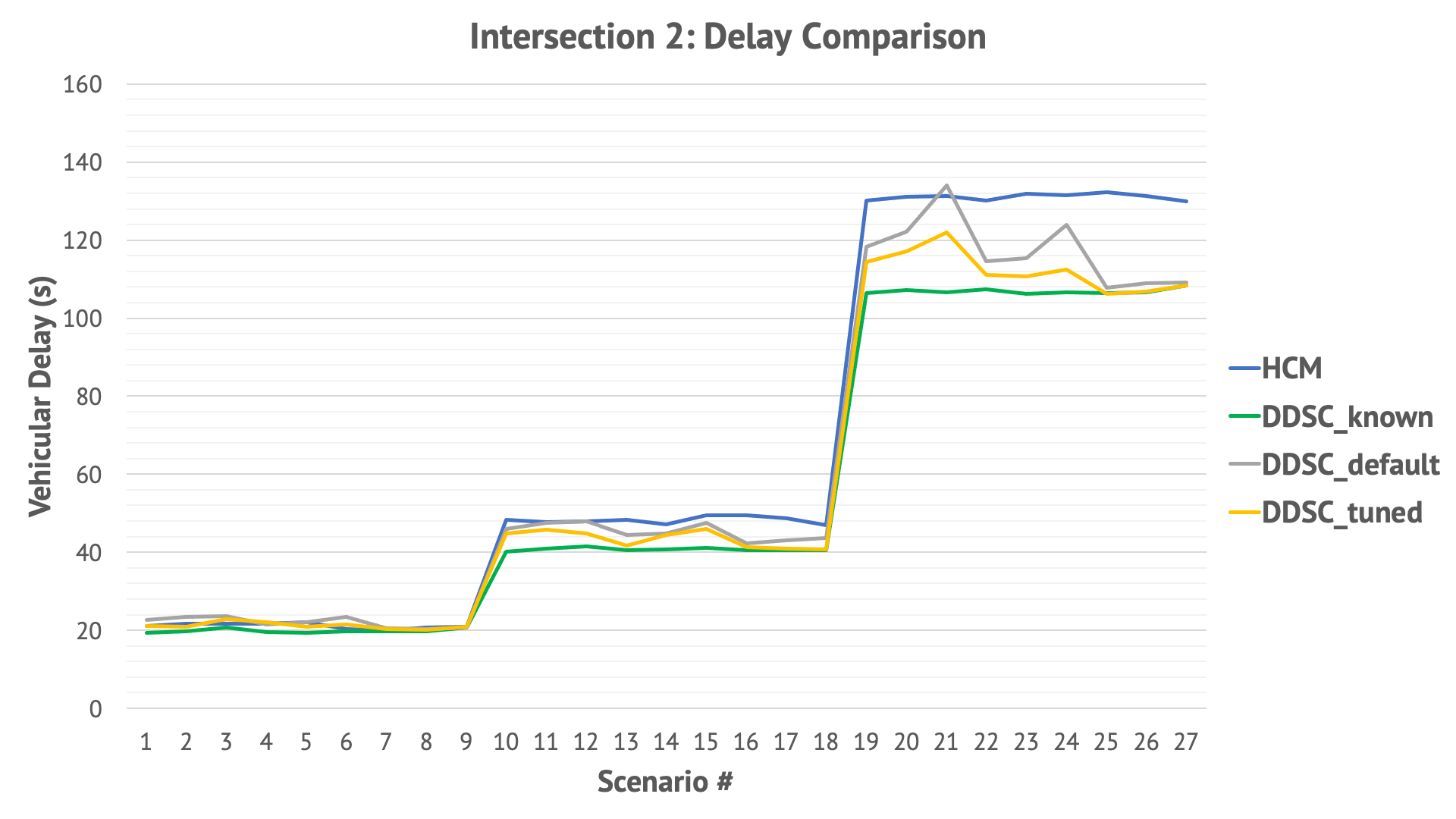}
    \caption{Intersection 2: Average Vehicular Delays}
    \label{fig:int2_results}
\end{figure}

Figs. \ref{fig:int1_results} and \ref{fig:int2_results} demonstrate the results for the two intersection models.

Method $DDTC_{known}$ yields the lowest vehicular delays. This was expected since the optimal timing system outperforms HCM as detailed in chapter \ref{chpt:signal} and the routing decisions are known to the system. 

$DDTC_{tuned}$ outperforms $DDTC_{default}$ as expected given its pre-tuned parameters; however, the results become almost identical for $mpr = 90\%$ (simulations 7-9, 16-18 and 25-27) as the routing estimation errors drop below 10\% even with un-tuned parameters.
\subsubsection{Influence of Traffic Level}
The scenarios are divided into three categories in terms of traffic level: Low (sims 1-9), Medium (sims 10-18) and High (sims 19-27). These categories can be identified with a glance at the figures above and noticing two step-like increases to the delays. 

The algorithm's performance in outperforming HCM and the importance of parameter tuning become more important with higher traffic levels. The proposed algorithm decreased HCM's delay by 15\%, 18.5\% and 22\% for low, medium and high traffic levels, respectively.

It should be noted that the $MPR$ parameter has no influence on the $HCM$ and $DDTC_{known}$ methods in this particular case, as both of these systems are assumed to be aware of the state of the traffic, hence the traffic state estimator is ignored.

\subsubsection{Influence of MPR}
Market penetration rate is perhaps the most important parameter affecting the proposed algorithm's performance. According to the results, on average, vehicular delays decrease by 15\% when mpr increases from 10\% to 50\% for both $DDTC_{tuned}$ and $DDTC_{default}$. This is explained by the traffic state estimator's reliance on CAV information.

The results showed that for an $MPR > 40\%$ the proposed method's performance (average delay) falls within $5\%$ of the ideal case where the routing decisions are known; and it becomes almost identical to the ideal case for $MPR >= 90\%$.
\subsubsection{Influence of Decision Change Interval}
Decision change interval is a major factor affecting the traffic state estimator's performance. Extremely short intervals (scenario numbers divisible by 3) greatly affect the proposed method's performance by increasing the estimator's error. This is dealt with to some extent by performing parameter tuning offline. 

It should also be noted that in real-world scenarios these intervals are usually longer than 5 minutes.

\section{Conclusions}\label{sec:conclusion_6}
This chapter proposed a data-driven approach to adaptive signal control. The proposed system relies on a data-driven method for optimal signal timing introduced in chapter \ref{chpt:signal} and a data-driven heuristic method for estimating routing decisions as described in chapter \ref{chpt:tse}.

This novel system requires no additional sensors to be installed at the intersection, reducing the installation costs compared to typical settings of state-of-the-practice adaptive signal controllers.

Several simulated scenarios were designed and implemented in a microscopic traffic simulator to cover a range of settings from low to high traffic levels, low to high market penetration rate and low to high levels of dynamism in routing decisions.

Simulations showed that the proposed algorithm outperforms HCM and when its parameters are tuned offline, it can decrease average vehicular delay by up to 22\%.

\chapter{Summary of Contributions \& Future Work}\label{chpt:conclusions}

\section{Contributions}
This research offers several contributions to cooperative intersection management, signal timing and traffic state estimation as listed below.
\subsection{Platoon-Based Autonomous Intersection Management}
Chapter~\ref{chpt:cim} introduced a reservation-based policy that utilizes two cost functions designed to minimize total vehicular delay and its variance to derive optimal schedules for platoons of vehicles. The proposed policy guarantees safety by not allowing vehicles with conflicting turning movement to be in the conflict zone at the same time. Moreover, a greedy algorithm was designed to search through all possible schedules to pick the best that minimizes a cost function based on a trade-off between total delay and variance in delay.

A simulator software was designed in Matlab to compare the results of the proposed policy in terms of average delay per vehicle and variance in delay with that of a 4-phase traffic light. A simple communication protocol was designed for V2I communication and two policies were introduced for the controller to minimize total delay and delay variance according to the cost functions tailored for platoons of vehicles.

According to the simulation results, the proposed controller minimizes travel delay and variance while increasing intersection throughput and reducing fuel consumption, when compared to traffic light policies. The simulations also verify the positive effect of platoon size on fuel consumption and intersection throughput. The simulations show that platooning can significantly decrease average vehicular delay as well as decrease the communication overhead by a factor of up to the average size of the platoons.

\subsection{A Data-Driven Approach to Optimal Signal Timing}
Chapter~\ref{chpt:signal} introduced a novel data-driven approach to the optimal signal timing problem. The proposed method relies on recorded data from various traffic scenarios through microscopic simulations. Multiple models can be built with different objective(s) such as minimizing delay, queue length or fuel consumption. The proposed model can be customized to each network's unique geometrical and traffic conditions, outperforms existing state-of-the-practice pre-timed signal control model and provides new insights into the the traffic control problem.

The following applications could be considered for the proposed method.
\begin{itemize}
    \item \textbf{Signal Timing Model} The proposed model can be utilized as a replacement to existing state-of-the-practice pre-timed signal timing methods such as the HCM model.
    \item \textbf{Signal Timing Helper} A trained model can be utilized as a helper method to any signal timing planner. For example, the model can provide customized cycle length and green time ranges to the signal timing planner.
    \item \textbf{Adaptive Signalization Method} With proper modifications, the proposed model can be applied in the context of real-time traffic responsive signal control. These modifications include readjusting the training scenarios as well as the structure of the learning model. A traffic state estimator would be required to provide the input to the model. The traffic state estimator module coupled with the proposed traffic signal model, form a real-time traffic responsive controller.
\end{itemize}
\subsection{Leveraging CAV Communications for Traffic State Estimation}
Chapter~\ref{chpt:tse} introduced a novel heuristic method for the estimation of dynamic routing decisions. The proposed method leverages communications from CAV(or CVs) to estimate aggregated routing decisions based on the samples from CAV information. This method provides the user with two configuration properties to periodically flush its memory and to maintain its estimation for a window of time immediately after the memory has been flushed.

The proposed algorithm, if properly tuned, can provide estimations with errors as low as (20\% - 30\%), for market penetration rates lower than 30\% and it offers an error as low as (5\% - 10\%) for higher penetration rates (mpr > 70\%).
\subsection{Real-Time Data-Driven Adaptive Signal Control}
Chapter \ref{chpt:method} proposed a data-driven approach to adaptive signal control. The proposed system relies on a data-driven method for optimal signal timing and a data-driven heuristic method for estimating routing decisions.

The proposed system requires no additional sensors to be installed at the intersection, reducing the installation costs compared to typical settings of state-of-the-practice adaptive signal controllers.

Simulations showed that the proposed algorithm outperforms HCM and when its parameters are tuned offline, it can decrease average vehicular delay by up to 22\%. 

\section{Application}
The proposed approach in Chapter 6 can be implemented and applied to any given intersection with minimal requirements for additional infrastructure. The intersection's geometrical information and designed signal structure can be imported into a traffic simulator.

The training set for the signal timing model can be generated using the simulator or recorded through real-world observations. The latter would require additional infrastructure to enable sensing of vehicle states including location and delay. If such infrastructure is available, online learning can also be applied where the learning model updates its weights according to the newly observed data.

Assuming a sufficiently large CV market penetration rate, the proposed approach is expected to improve traffic.
\section{Future Work}
An important area of future work for any type of urban traffic controller is design and implementation of real-world settings for any proposed controller to be tested in a safe yet realistic manner. Honda's smart intersection concept is one of the few examples of such efforts in the past few years.

As a side note, one of the shortcomings of the research in this area (ITS) is the lack of data and code sharing. In the era of open-source movement, it is time for researchers to openly share their methodology, data and code with other researchers to speed up the progress. 

\subsection{Cooperative Intersection Management}
Assuming all-autonomous traffic is one the major shortcomings of the previous work on CIM. Most of the research has focused on theoretical solutions for all-autonomous intersections, ignoring less \emph{controllable} aspects of the roads (e.g. human-driven vehicles, cyclists, pedestrians). Future work has to consider the trade-offs of such realistic scenarios and propose solutions that can be gradually tested in real-world settings and introduced into the public roads.

Another important area is to provide generally acceptable benchmarks for new methods to be compared by. This is a non-trivial task and requires great knowledge of the problem to design a benchmark that provides a holistic view of the many aspects of the intersection control problem.
\subsection{Adaptive Signal Control}
An important area that was not addressed in this dissertation is taking into account less controllable road users such as pedestrians and cyclists. Adaptive signal control should also be adaptive to the pedestrian traffic and unexpected events and road conditions. 

The signal timing model presented in this work can be further developed into an online learning model that provides better estimations using real-world observations of the same intersection. This could result in significant improvements in performance and provide a fully customized traffic controller for any given traffic intersection.

\appendix
\addcontentsline{toc}{section}{Appendices}
\renewcommand{\thesubsection}{A}
\section*{Appendices}
\subsection{Evaluation of HCM's Delay Estimation}
  \label{first_appendix}
Intersection delay can be estimated by measurement in the field, microscopic simulation and analytical models. Among the three methods, analytical models are known for their lower precision when compared to simulation and field measurement. We take advantage of the data collected for the comparisons section to validate and quantify the inaccuracy of HCM's estimation of control delay.
  
\begin{figure}[htbp]
\captionsetup[subfloat]{captionskip=5pt}
    \centering
    \caption{Estimated Delay Comparison: Analytical Model vs. Microscopic Simulation}
        \subfloat[First Intersection]{\includegraphics[width=0.75\textwidth]{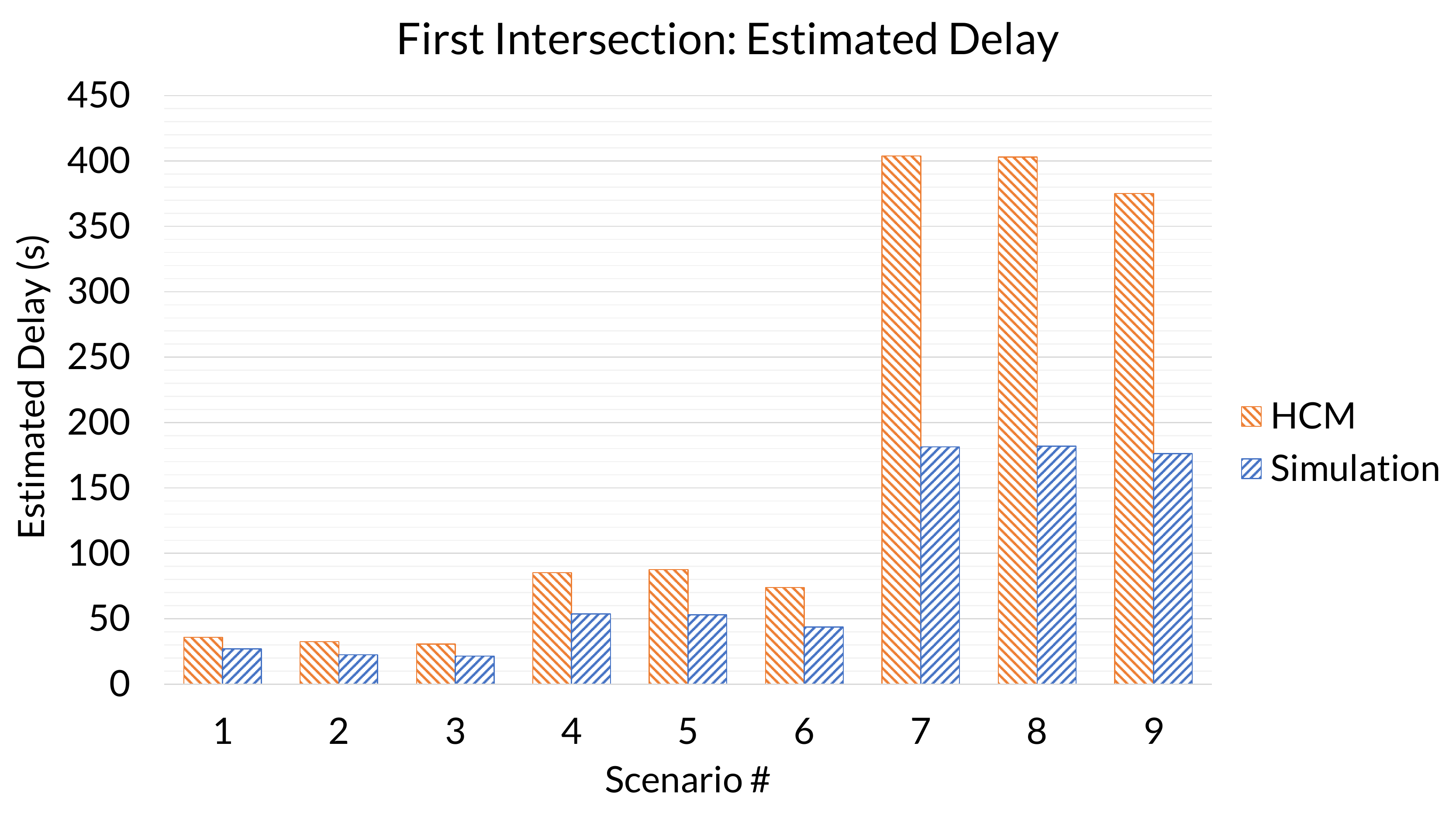}}
        \qquad
        \subfloat[Second Intersection]{\includegraphics[width=0.75\textwidth]{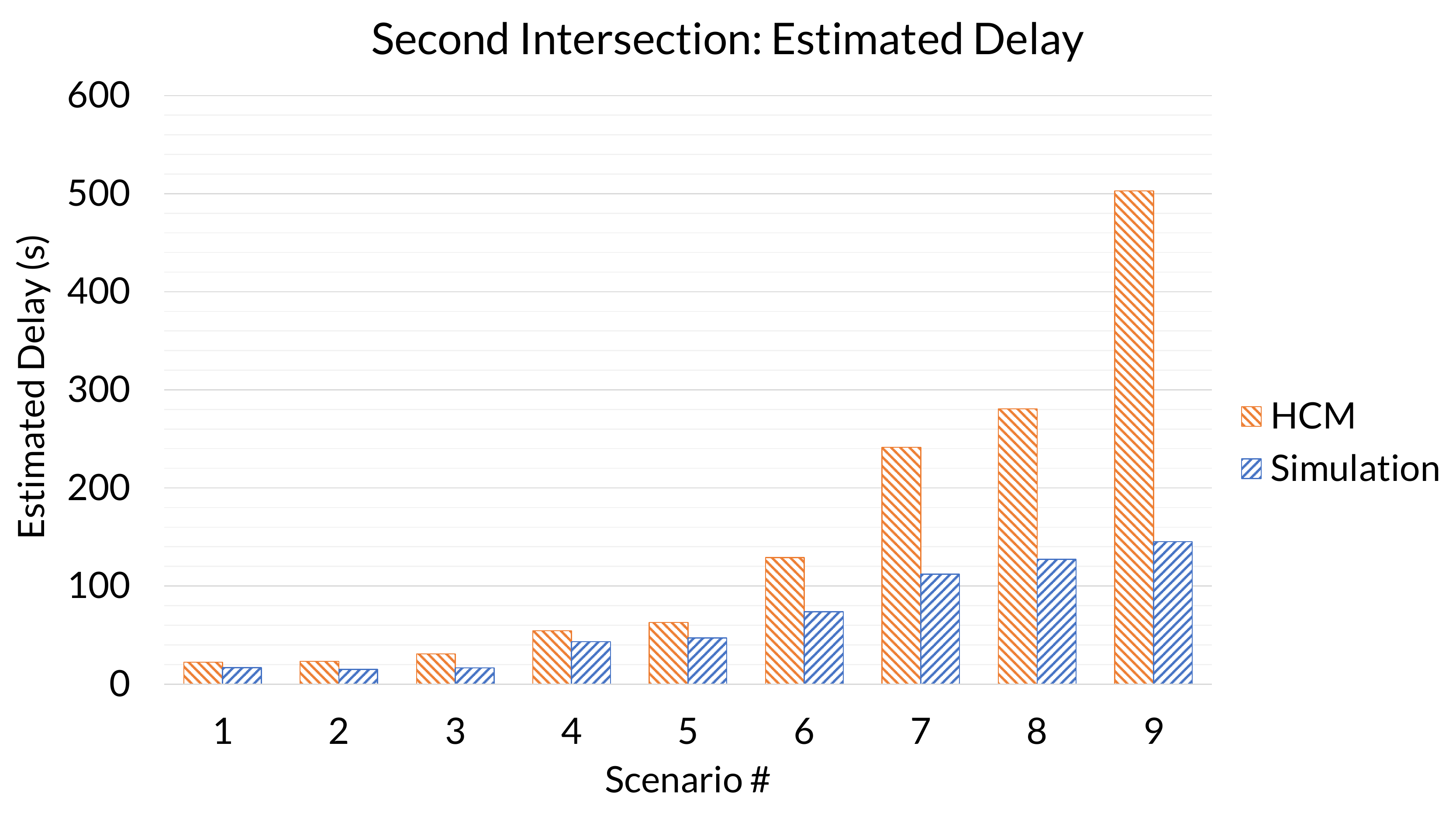}}
        \qquad
        \subfloat[Third Intersection]{\includegraphics[width=0.75\textwidth]{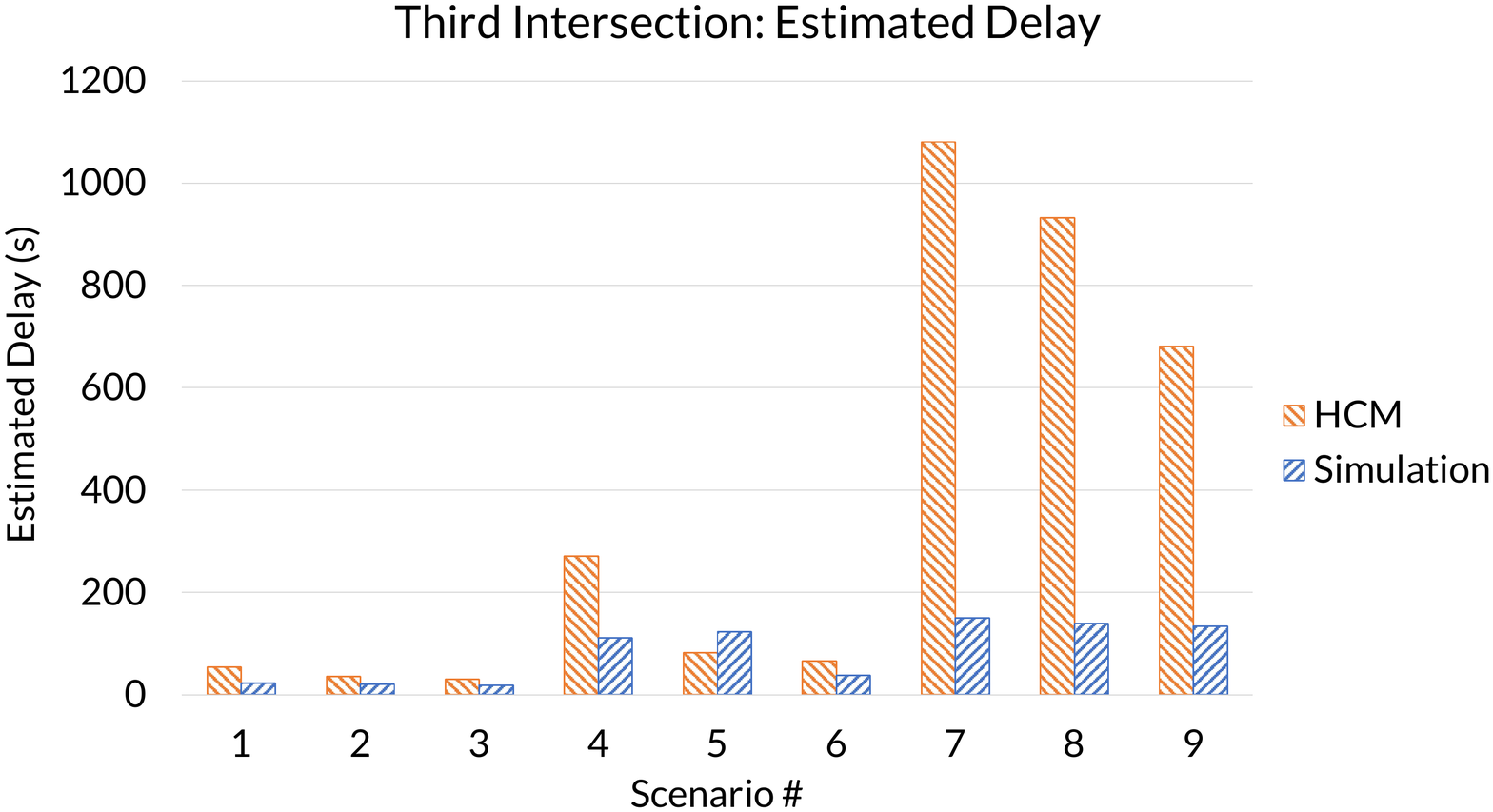}}
    \label{fig:est_delays}
\end{figure}

Fig.~\ref{fig:est_delays} demonstrates a comparison between HCM's estimated control delay and the measurements from a microscopic traffic simulator given identical settings. As can be seen in these graphics, HCM's estimations were always higher than those from the microscopic simulations. A more detailed numerical comparison showed that, on average HCM's estimations are~$100\%$ larger than the estimations from microscopic simulation. HCM's overestimation is even more significant for high traffic scenarios that results in highly inaccurate estimations of the optimal cycle length. 
  
\cleardoublepage
\addcontentsline{toc}{chapter}{References}

\end{document}